\documentclass[final,1p,times,twocolumn,review,authoryear]{elsarticle}

\usepackage{amssymb}
\usepackage{algorithm}
\usepackage{algpseudocodex}

\usepackage{float}
\usepackage{amsmath}
\usepackage{subcaption}
\usepackage{graphicx}
\usepackage{booktabs}
\usepackage{tabularx}
\usepackage{geometry}

\usepackage{caption} 
\usepackage{subcaption} 

\usepackage{amsthm}

\makeatletter
\def\ps@pprintTitle{%
  \let\@oddhead\@empty
  \let\@evenhead\@empty
  \def\@oddfoot{\hfill \@date}%
  \let\@evenfoot\@oddfoot}
\makeatother

\begin{document}

\begin{frontmatter}

\title{A representational framework for learning and encoding structurally enriched trajectories in complex agent environments}

\author[label1]{Corina Cătărău-Cotuțiu}
\ead{Corina.Catarau-Cotutiu@citystgeorges.ac.uk}

\author[label1]{Esther Mondragón}
\ead{e.mondragon@citystgeorges.ac.uk}
\author[label1]{Eduardo Alonso}

\affiliation[label1]{
    organization={Artificial Intelligence Research Centre (CitAI), Department of Computer Science},  
    addressline={City, University of London},  
    city={London},  
    postcode={EC1V 0HB},  
    country={United Kingdom}
}

\begin{abstract}
The ability of artificial intelligence agents to make optimal decisions and generalise them to different domains and tasks is compromised in complex scenarios. One way to address this issue has focused on learning efficient representations of the world and on how the actions of agents affect them in state-action transitions. Whereas such representations are procedurally efficient, they lack structural richness. To address this problem, we propose to enhance the agent’s ontology and extend the traditional conceptualisation of trajectories to provide a more nuanced view of task execution. Structurally Enriched Trajectories (SETs) extend the encoding of sequences of states and their transitions by incorporating hierarchical relations between objects, interactions, and affordances. SETs are built as multi-level graphs, providing a detailed representation of the agent dynamics and a transferable functional abstraction of the task. SETs are integrated into an architecture, Structurally Enriched Trajectory Learning and Encoding (SETLE), that employs a heterogeneous graph-based memory structure of multi-level relational dependencies essential for generalisation. We demonstrate that SETLE can support downstream tasks, enabling agents to recognise task relevant structural patterns across CREATE and MiniGrid environments. Finally, we integrate SETLE with reinforcement learning and show measurable improvements in downstream performance, including breakthrough success rates in complex, sparse-reward tasks.

\end{abstract}

\begin{keyword}
Representation learning \sep Trajectories \sep Reinforcement Learning \sep Heterogeneous graphs \sep Graph Neural networks

Code available here: https://github.com/CatarauCorina/setle\_code

\end{keyword}

\end{frontmatter}


\section{Introduction}

Autonomous systems perceive their environment, take actions, and adapt their behaviour through experience. One key paradigm in this field is Reinforcement learning (RL) \citep{sutton1998reinforcement}, which enables agents to learn from interaction with their environment by using reward signals to guide decision-making. Unlike supervised learning, which relies on labelled data, or unsupervised learning, which identifies patterns in data, RL allows agents to develop policies that maximise cumulative rewards over time. This approach has led to significant advancements in Artificial Intelligence (AI), for instance in game-playing AI \citep{silver2017mastering}, robotics \citep{levine2018learning}, and autonomous systems \citep{kiran2021deep}. Despite these successes, RL agents face fundamental challenges when transferring knowledge from simpler environments to complex, dynamic ones requiring strong generalisation and knowledge transfer.

Representation learning plays a crucial role in addressing these challenges by developing structured, transferable encodings of an agent’s environment \citep{bengio2013representation, lesort2018state}. Various approaches have been proposed, for example, hierarchical RL \citep{pateria2021hierarchical} introduces structured abstractions for multi-step decision-making but remains constrained by predefined reward structures \citep{sutton1999between, bacon2017option}. Goal-conditioned RL enhances planning by leveraging future states as learning targets, yet still depends on explicitly defined objectives \citep{kaelbling1993learning, schaul2015universal}. Disentanglement techniques refine state representations by separating controllable and uncontrollable factors, improving sample efficiency \citep{higgins2018towards}.

Although these proposals have been instrumental in enhancing policy learning, improving sample efficiency, and facilitating task-specific generalisation, moving beyond compression methods towards enriching structural dependencies between states and actions is a critical step for more robust and general-purpose intelligence \citep{nie2017}. 

Traditional representations provide a functional foundation, however, they lack the flexibility to support generalisation and knowledge transfer effectively, which may require enriching RL’s ontology. Rather than treating states as undifferentiated wholes, decomposing them into their constituent objects and integrating interaction semantics yields a more hierarchically structured and transferable representation of task dynamics. This approach would enhance generalisation across tasks, allowing agents to recognise shared structural patterns even in novel environments. 

Encoding isolated interactions between objects, however, is not enough; agents need to also learn how these interactions influence future decisions and outcomes. To model this effectively, it is essential to incorporate affordances, which describe the potential interactions available to an agent based on both its capabilities and the environment \citep{gibson1977theory}. An affordance is a property of an object in relation to others and from the perspective of an agent that prompts a specific use or interaction. For example, a chair may afford sitting to a human but not to a large animal, such as an elephant. Structured representations of affordances enable an agent to predict the effects of its interactions, providing a foundation for adaptive motor planning and effective task generalisation \citep{schack2013representation}, also enable agents to anticipate how changes in context impact success or failure, reducing the risk of misgeneralisation \citep{khazatsky2021affordances}. 

Despite all the advantages that representing affordances would introduce, they alone would only apply to a given state. That is, affordances do not track how interactions unfold sequentially.
Tracking how interactions
unfold sequentially requires maintaining a structured history of transitions, a trajectory. By retaining this structured trajectory rather than relying solely on the Markov assumption that states encapsulate all relevant history, disregarding transitions \citep{niv2019learning}, agents will no longer be compelled to make decisions based only on their current state and will be able to learn the dependencies that shape long-term task execution, enabling more
adaptive and informed decision-making.
Encoding such sequences of interactions and their distant effects has been shown to be essential for AI systems to adapt to novel scenarios \citep{eysenbach2022contrastive}, enabling the agent to capture long-term dependencies and avoid inefficient learning or goal misgeneralisation \citep{di2022goal}.

We define a Structurally Enriched Trajectory (SET) as an extension of the concept of trajectory. A SET goes beyond the conventional flat state-action structure by providing a structured representation that encompasses not only sequences of state-action transitions, but also encodes critical components that influence task execution. Unlike raw trajectories, which merely record the steps taken, SETs incorporate shared low-level components (objects, interactions) and the affordances that emerge from them, encapsulating information about not just what happened but how different elements contributed to varying results.

\hfill \break
To implement such a conceptual framework that expands the RL ontology to capture hierarchical dependencies and shared trajectory knowledge, we propose a computational model that uses hierarchical subgraphs to model interactions between states, actions, and affordances, ensuring that learnt representations retain relational dependencies crucial for task execution and generalisation, which we call Structurally Enriched Trajectory Learning and Encoding (SETLE). We construct hierarchical graphs that capture interactions between states, actions, and affordances, ensuring that learnt representations retain relational dependencies essential for task execution. These structured representations empower agents to learn compact trajectory encodings, enabling them to recognise task-relevant similarities and differences across diverse environments.
Graph-based learning models provide a powerful framework for capturing hierarchical causal relationships, which we pose as essential for understanding the sequence of interactions and how actions shape long-term outcomes. 

These developments present an opportunity to expand the representation of trajectories with hierarchical structures. Unlike traditional representations that treat states and actions as independent transitions, graph structures naturally encode dependencies across multiple levels of abstraction, preserving how interactions evolve over time. As the need for modelling complex dependencies increases, Graph Neural Networks (GNNs) have become a key tool for learning from structured data, enabling AI systems to capture rich relational information. Heterogeneous Graph Neural Networks (HGNNs) \citep{wang2021selfsupervised} extend this capability by incorporating multiple node and edge types, allowing a more detailed representation of diverse interactions in dynamic environments.
Still, despite their merits, most existing graph-based approaches focus on encoding full graphs
rather than leveraging subgraphs with shared components. This limits their ability to model localised structures while capturing commonalities across different subgraphs. By treating entire graphs as independent entities, these methods often fail to exploit the reusability of nodes, making generalisation more challenging. A subgraph-based approach, where fundamental components are shared, offers a more effective way to represent task dynamics and facilitate knowledge
transfer.

Summarising, the contributions of this research are as follows:

\begin{enumerate}
    \item A conceptual framework for RL representations that extends RL's ontology to incorporate affordances and hierarchical trajectory structures, enabling agents to capture sequences of different states and their relational dependencies.
    \item A formal conceptualisation of SET to expand the notion trajectories to include not only state-action sequences but also objects, interactions, affordances, and task-relevant components.
    \item A novel architecture SETLE to extract and encode SETs, including:
    \begin{enumerate}
        \item A novel Hierarchical Memory Structure: A multi-level memory system that encodes task elements at varying levels of abstraction, capturing both fine-grained interactions and high-level trajectory patterns, supporting robust generalisation across tasks.
        \item Heterogeneous Sub-Graph Encoder: A specialised encoder designed to learn trajectory representations by leveraging subgraphs with shared components rather than treating full graphs as independent entities.
       {\item A comprehensive evaluation demonstrating that integrating SETLE into a reinforcement learning agent leads to substantial improvements in policy learning, sample efficiency, and generalisation.}
    \end{enumerate}
   
    \item Experimental evidence suggesting (in CREATE and MiniGrid) that SETLE effectively captures both task-specific patterns and cross-task similarities, supporting structured trajectory learning in dynamic, episodic environments.

\end{enumerate}

The remainder of this paper is structured as follows: first, we review related work on trajectory representation learning and heterogeneous graph models, contextualising the limitations of current approaches. Next, we introduce the SETLE framework, detailing its hierarchical graph construction and encoding. {We then present in Sections 4 and 5 experimental results that first validate the representational power of SETLE's encoder in CREATE and MiniGrid environments, and second, demonstrate its effectiveness when integrated into a full reinforcement learning loop.} Finally, we discuss the broader implications of our findings and outline future research directions.

\section{Related work}
 {
A core challenge in reinforcement learning (RL) is generalisation. While agents can master specific tasks, they often struggle to transfer knowledge to new situations. This limitation stems largely from how they represent their experiences. Early approaches focused on learning compressed state representations \citep{wang2021selfsupervised}, aiming to distil high-dimensional observations into compact vectors. However, while efficient, these static snapshots often fail to generalize well, as not all aspects of a state are transferable to new tasks .

A more powerful approach is to learn from trajectories, the full sequence of states and actions leading to a result. Yet, the first and most basic information gleaned from a trajectory is its outcome \citep{williams1992reinforce}. Knowing whether a sequence of actions led to success or failure is a crucial learning signal. But relying solely on outcomes is insufficient for transferring knowledge. For instance, the outcome of travelling to a destination can be achieved by either riding a bike or driving a car. While the outcome is the same, the skills required are entirely different \citep{collins2021}. Knowing only that both methods were 'successful' provides no transferable knowledge about how to perform either task. This emphasises that understanding the trajectory, the way in which a task is completed, is crucial for adapting and transferring skills.

Furthermore, even learning from trajectories may not suffice to capture the nuances of information composition and organisation. Classically, they are defined as "flat" sequences of state-action pairs, a format that lacks deep structural information. This overlooks the rich internal structure that humans intuitively grasp. A key strength of human problem-solving is the ability to develop hierarchical and causal models of our interactions, allowing us to recognize structural similarities and adapt our strategies dynamically \citep{schack2013representation,morris2022neural}. The key challenge for AI, therefore, is not just to record trajectories, but to learn to represent their internal structure. This section reviews the evolution of attempts to solve this problem, from early temporal abstractions to modern, heterogeneous graph-based methods.

\subsection{Early Trajectory Representations: From Flat Sequences to Temporal Abstractions}
Methods like the Outcome-Driven Actor-Critic (ODAC) trajectory-based prediction models, and memory-based RL frameworks such as Sequential Episodic Control (SEC) all adopt simple, shallow sequences of state-action pairs \citep{rudner2021outcome, haines2018outcome, freire2024sequential}, focusing on linear dependencies rather than hierarchical structures \citep{sutton1998reinforcement}.

While these approaches improve learning by preserving sequential patterns, their flat memory structure prevents them from capturing the deeper hierarchical dependencies essential for structured learning and skill reuse.

Early efforts to add structure focused on temporal abstraction. For example, Precup and collaborators' options framework (Precup et al., 2000) temporally extended actions, allowing agents to execute multi-step behaviours. Similarly,  Efroni et al., (2021) work on RL with trajectory feedback, shifted evaluation from single steps to entire trajectories.
However, both approaches treated behaviours as indivisible units, lacking the granularity needed for structured generalisation. Fine-grained execution details, including the object properties and interaction dynamics, were disregarded in these approaches, making it difficult to identify reusable behavioural patterns.

\subsection{The Shift to Structured Representations: Skills and Latent Spaces}
To better capture a trajectory's internal structure, subsequent research focused on learning representations of entire trajectories. The Self-Consistent Trajectory Autoencoder (SeCTAR) model framed the problem as one of representation learning, building continuous latent spaces of trajectories to represent structured behaviours \citep{co-reyes2018sectar}. While a significant step, SeCTAR's latent space does not explicitly segment trajectories into transferable skills, limiting its use for compositional learning.

In contrast, algorithms like Constructing Skill Trees (CST) automatically segment demonstration trajectories into reusable skills \citep{konidaris2011cst}. CST enabled the decomposition of long tasks into meaningful sub-policies. However, CST's reliance on human-provided demonstrations means it cannot autonomously discover skills through its own exploration.

\subsection{Relational Learning: The Emergence of Graph-Based Approaches} 
While the aforementioned methods advanced temporal and hierarchical modelling of trajectories, they did not inherently enable the encoding of the relational dependencies between entities within a state. Understanding how different components interact over time requires a representation that can capture both temporal and relational dependencies.

Graph-based approaches provide a natural way to represent this structured knowledge. Heterogeneous Graph Neural Networks (HGNNs) are particularly well-suited, as they can encode multi-level interactions between different types of nodes (e.g., objects, agents, affordances) using mechanisms like meta-paths \citep{wang2021selfsupervised}. Some research has explored causal graphs to model structured transitions rather than mere correlations, though these often rely on semi-supervised learning, making them unsuitable for pure RL settings \citep{jin2024causal}.

\subsection{The State of the Art: Heterogeneous and Self-Supervised Graph Models}
Recent advances in HGNNs have focused on self-supervised learning. Heterogeneous Co-contrastive Learning (HeCo) uses two complementary views (network schema and meta-path) to learn rich node embeddings for static graphs \citep{wang2021selfsupervised}. Graph Transformer Co-contrastive Learning (GTC) extends this with a Transformer-based architecture to better capture both local and global dependencies \citep{sun2025gtc}.

Despite their power, these state-of-the-art methods share a critical limitation: they are designed for static, node-centric tasks on a single, large graph. Their focus on node-level representations and fixed graph structures limits their applicability in the dynamic, episodic environments of RL, where representations must evolve to reflect the cumulative nature of actions over a trajectory.

\subsection{Identifying the Gap and Positioning SETLE}
The evolution of trajectory representations highlights a clear gap in the literature. While trajectory-based models often lack relational detail, advanced graph-based models are not designed for the dynamic, episodic nature of RL, revealing the need for a framework that combines the strengths of both. Existing methods often fail to account for shared entities reoccurring across trajectories, which prevents them from recognising common patterns and limits generalisation.

To overcome these limitations, it is crucial to expand the RL ontology to include structured representations of task elements and cross-episode relationships. SETLE is designed to provide a solution. By modelling individual trajectories as distinct, structured subgraphs with shared components, our framework directly addresses the limitations of prior static, node-centric, or flat-sequence models.

}
\section{Methodology}
This section outlines the conceptual framework and its implementation (which can be found here https://github.com/CatarauCorina/setle\_code). SETLE was designed to represent the dynamics of task execution by structuring trajectories as hierarchical subgraphs. Unlike traditional methods that aggregate information across entire graphs, we introduce a subgraph-centric approach tailored for episodic data. Each episode is treated as an independent subgraph, with its representation depending solely on its local neighbourhood.

A key distinction of our framework is the ability to handle shared nodes that may appear in multiple trajectories or episodes. These shared nodes serve as bridges between episodes, allowing the emergence of cross-episode patterns. For example an object, such as a ball, might be used across different episodes but in varying contexts (e.g., being pushed to a goal in one task and placed in a bucket in another). Similarly, an action like \texttt{push} could have different effects depending on the affordances and states in multiple episodes.

By encoding these shared nodes within their respective trajectory subgraphs, the relationships learnt from one trajectory or episode can inform the understanding of similar interactions in other episodes, enhancing generalisation across tasks.

We first define the ontology of our framework in Subsection 3.1,. then describe the process of constructing Structurally Enriched Trajectories (SETs) as heterogeneous graphs Subsection 3.2, ensuring they encode multi-level relationships between objects, interactions, states, and affordances. We next detail the methodology for learning SET embeddings in Subsection 3.3, which can be used for downstream tasks, including integration into reinforcement learning pipelines as seen in Section 5.

\subsection{SETLE Ontology}  
This section defines key components of our framework, including interactions, Structurally Enriched Trajectories (SETs), affordances, and hierarchical memory, which together provide a structured representation of task execution.  

\subsubsection{Interactions and Objects} 
Objects and interactions are structured embeddings extracted from the environment that encode task-relevant features that influence decision-making. An interaction $i_t$ represents the interplay between an agent-object or object-object relationship observed within the environment. However, due to the constraints of the environment used in our experiments, interactions are limited to agent-object relations, where the agent is always a ball object, interacting with various other objects in the scene.

\subsubsection{Affordances}
Following \citep{catarau2023aigenc}, an affordance is conceptualised as the learnt relationships between an (effect, reward) pair and a (concept-object, interaction) tuple. When an agent applies an action, it produces an interaction, which in turn generates an effect (a state transition) and an associated reward, both of which are stored as an affordance. This structured representation is designed to improve generalise across tasks by predicting how different interactions influence trajectories across varying contexts.

Formally, an affordance \( A \) encapsulates an agent’s action, the current state, and the resulting trajectory \citep{catarau2023aigenc}:

\begin{equation}
A = \{s_t, a_t, s_{t+1}, r_t\}
\end{equation}
where \( s_t \) is the current state, \( a_t \) is the agent's action, \( s_{t+1} \) is the resulting state, and \( r_t \) is the reward at time \( t \). 

\subsubsection{Structurally Enriched Trajectories (SETs)}\label{sec:set} 
In reinforcement learning (RL), a trajectory is traditionally defined as a sequence of states and actions:

\begin{equation}
    \tau = \{(s_0, a_0), (s_1, a_1), \dots, (s_t, a_t),(s_{t+1}, a_{t+1}), \dots, (s_T, a_T)\}
\end{equation}

where $s_t \in S$ represents the agent's state at time step $t$, $a_t \in A$ is the action taken, and $T$ denotes the length of the episode.  

We define a \textbf{Structurally Enriched Trajectory (SET)} as an extension of this classical trajectory that incorporates additional structural components, capturing relational dependencies and affordance-based interactions. A SET is represented as a hierarchical graph:  

\begin{equation}
    \mathcal{G}_{\tau} = (V, E, \Phi)
\end{equation}

Where:  

\begin{description}
    \item[$V = \bigcup_{t=0}^{T} \{O_t, I_t, \mathfrak{\hat{f}}_t, s_t\}$ is the set of nodes, where:] \begin{description}
    \item 
    \item $O_t$ represents the set of all objects present at time step $t$.
    \item $I_t$ represents the set of all interactions occurring at time step $t$. 
    \item $\mathfrak{\hat{f}}_t$ represents the affordance at time step $t$.
    \item $s_t$ is the state of the agent at time step $t$
\end{description}
\end{description}

The union across all $t$ ensures that $V$ contains the full trajectory of objects, interactions, affordances, and states throughout the episode.

\begin{description}
    \item[$E \subseteq V \times V$ ] is the set of edges, encoding temporal transitions $(s_t, \mathfrak{\hat{f}}_t, s_{t+1})$ and structural rrelationships, such as object interactions $(o_i, o_j)$.  
\end{description}

\begin{description}
    \item[$\Phi: V \to \mathbb{R}^d$] is a feature extraction function that assigns high-dimensional embeddings to objects, interactions, and states.
\end{description}

This enriched representation allows us to model task execution beyond stepwise transitions, capturing how task elements evolve over time and interact within different contexts.  

\subsubsection{ Ontology of the Hierarchical Memory} \label{sect:hier_mem} 
We organise multiple SETs into a hierarchical memory structure, represented as a large heterogeneous graph, that captures shared trajectory components across different episodes. The hierarchical memory is formally represented as:  

\begin{equation}
    \mathcal{M} = (\mathcal{G}_1, \mathcal{G}_2, \dots, \mathcal{G}_N)
\end{equation}
Where:  

- $\mathcal{G}_i$ represents an individual SET subgraph, modelling a structured trajectory of an episode. 

- Each subgraph \(\mathcal{G}_i\) can be further decomposed into a sequence of lower-level graphs \(\mathcal{G}'_i\), where each \(\mathcal{G}'_i\) corresponds to a state in the episode, decomposed into its fundamental components (objects and interactions).

- SETs share low-level components, such as objects and interactions, creating relational dependencies across multiple trajectories.

The hierarchical structure of the memory captures different levels of abstraction. At the lowest level, each state is represented as a graph \(\mathcal{G}'_i\), encoding the relationships between objects and their interactions.  At a higher level, SETs encapsulate structured trajectories, linking sequences of state-graphs through affordances into meaningful task representations.
Unlike traditional memory structures that store isolated trajectories, our hierarchical memory structure connects trajectories through shared elements, preserving both task-specific dependencies and cross-task relationships. This structured representation allows agents to retrieve relevant experiences based on shared trajectory components, improving adaptability and transfer in reinforcement learning environments.

To formally represent the hierarchical graph for a SET, we model the task environment as a Heterogeneous Information Network (HIN), defined as a graph \( G = (V, E, A, R, \Phi )\).

The network schema of \( G \), denoted as \( \Gamma = (A, R) \), is used to describe the structure of interactions between nodes of different types, capturing the local structure. The schema is used to define direct connections among various node types (such as states, interactions, affordances and trajectories), forming the foundation for encoding relationships in heterogeneous networks.

Additionally, we define *meta-paths* to capture the composite relationships within task episodes. A meta-path \( P \) is defined as a sequence \( A_1 \xrightarrow{R_1} A_2 \xrightarrow{R_2} \dots \xrightarrow{R_i} A_{i+1} \), where each \( A_i \) is a node type and each \( R_i \) is a relation. Meta-paths describe the indirect, higher-order connections between node types, such as sequences connecting states to the enriched trajectories, which are essential for capturing complex dependencies across task hierarchies.

This formal definition of the memory as a hierarchical, heterogeneous graph provides the structural foundation for our learning framework. The specific architecture used to process this structure and the mechanism by which knowledge is propagated across its shared nodes are detailed in Section \ref{sect:set_enc}.

\subsubsection{Structured Data Sources: CREATE and MiniGrid}
The constraints of our framework required environments that support structured, multi-step interactions and diverse tasks. To test our approach across different types of complexity, we selected two distinct platforms.

Our primary environment is Construction, Reuse, and Extension of Actions through Transfer and Exploration (CREATE) \citep{jain2020generalization}. CREATE is a diverse and challenging platform designed to study action generalisation, providing tasks that involve dynamic interactions with various objects and tools in a continuous physics-based simulation. It allows us to explore generalisation by varying task setups, forcing agents to adapt strategies and reuse knowledge from past experiences.

To complement CREATE's physical complexity, we also employ the MiniGrid benchmark suite. MiniGrid offers a simplified, discrete setting with minimalist 2D grid worlds composed of symbolic entities like keys and doors. It features a limited, discrete action space and agent-centric partial observability. This environment serves to test SETLE's capacity for abstraction and multi-step planning in a domain that emphasizes logical and planning complexity rather than perceptual richness.

\subsection{SETLE: Hierarchical Graph Construction} \label{sect: hier_graph_constr}

Hierarchical graphs were built to capture structured relationships between objects, their interactions, affordances and state transitions as SETs. We construct these representations by processing raw interactions from the environment and encoding them into heterogeneous graph nodes and edges that preserve both local and global dependencies. In this section, we cover the different operations required to obtain these graphs.

\subsubsection{SET Definition Through Random Exploration}

In RL, an episode consists of a complete sequence of interactions from an initial state to termination, encapsulating the trajectory an agent follows within a given task. To extract and formalise SETs (for a detailed definition, see Section \ref{sec:set}) we employ an exploration strategy within the CREATE environment \citep{jain2020generalization}. In this setup, agents engage with tasks such as pushing a ball to a goal or placing it inside a bucket, generating diverse trajectories. To maximise variability in the trajectory space and better capture adaptive patterns and task-relevant structures, RL agents operate under a random policy rather than optimised for specific rewards.

Each episode is represented as a SET graph (see Fig.\ref{fig:all_lvls}), encapsulating the task setup, actions taken, and the result of the task, which is stored for validation and further analysis in the hierarchical memory (see \ref{sect:hier_mem}). 

An episode \( E \) is defined as a sequence of state-action pairs:
\begin{equation}
        E = \{(s_0, a_0), (s_1, a_1), \dots, (s_t,a_t), \dots (s_T, a_T)\}
\end{equation}
 where \( s_t \in S \) is the state at time \( t \), and \( a_t \in A \) is the action taken at time \( t \).
   
A SET is labelled as \textbf{success} if the agent reaches the goal; otherwise, it is labelled as \textbf{failure}. The SET of an episode \( SET(E) \) is labelled as follows:
\begin{equation}
        SET(E) = 
    \begin{cases} 
    \text{success}, & \text{if } s_T \in G \\
    \text{failure}, & \text{otherwise}
    \end{cases}
\end{equation}
    where \( G \) is the set of goal states.

SETs are represented hierarchically (see Section \ref{hier_sec}) in the memory structure, including the episode’s inventory and action sequence. For each episode, the corresponding SET graph \( G_{SET} \) is constructed, consisting of nodes representing states and edges representing state transitions (i.e., as seen in Fig.\ref{fig:all_lvls}). 

\subsubsection{Building the Hierarchical Memory Graph}\label{hier_sec}

\begin{figure}[h!]
    \centering
    \includegraphics[width=1\linewidth]{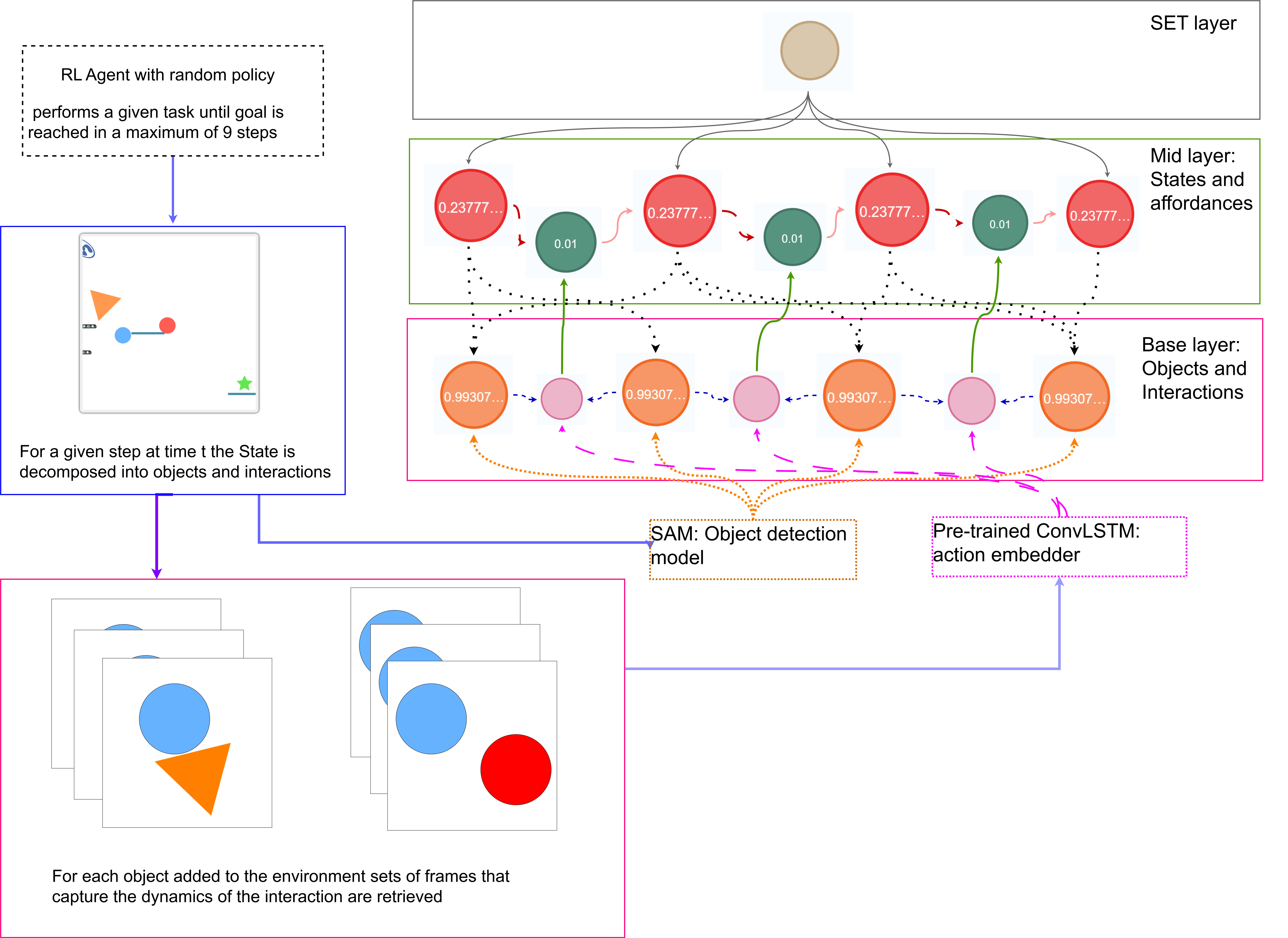}
    \caption{Hierarchical Memory building: The image shows the models used to build the hierarchical memory. The base layer extracts objects and interactions using the SAM object detection model and a pre-trained ConvLSTM, capturing fundamental task dynamics. The mid-layer represents states and affordances, modelling how interactions influence future states. At the highest level, the SET layer encodes trajectory-level abstractions, capturing long-term dependencies across sequential states. This hierarchical organisation enables efficient reasoning over task structures and adaptive decision-making.}
    \label{fig:hislore_data}
\end{figure}

To effectively model SETs, we structured episodic data into hierarchical heterogeneous graphs, where the feature extraction functions $\Phi: V \to \mathbb{R}^d$ (see Section \ref{sec:set}) assigned high-dimensional embeddings to nodes representing objects, interactions, and states. The nodes and their different relations are organised into hierarchical heterogeneous graphs with three distinct levels of abstraction: (1) a base layer for objects and interactions, where objects are extracted using the SAM model \citep{kirillov2023segment} and interaction dynamics are encoded via a ConvLSTM; (2) a mid-layer for states and affordances, capturing the transition dynamics and relationships between interactions and their effects; and (3) a trajectory layer for task-level abstractions, representing the cumulative impact of interactions and sequences (see  Fig. \ref{fig:hislore_data}). The selection of these specific models was based on a comparative analysis to ensure robustness and generalisability, as explained in our component analysis in Section \ref{sec:component_analysis}. This hierarchical organisation is stored in a graph database and is not simply a stratification of layers but a structured representation where lower levels influence higher-level abstractions. Objects and actions influence affordances, which in turn shape state transitions and contribute to the formation of SETs. This dependency across levels enables agents to reason over task structure, capturing both local dynamics and high-level task generalisation in complex environments. Formally, the hierarchical memory graph \textbf{G} representing a SET consists of:

\begin{figure}[h!]
    \centering
    \includegraphics[width=1\linewidth]{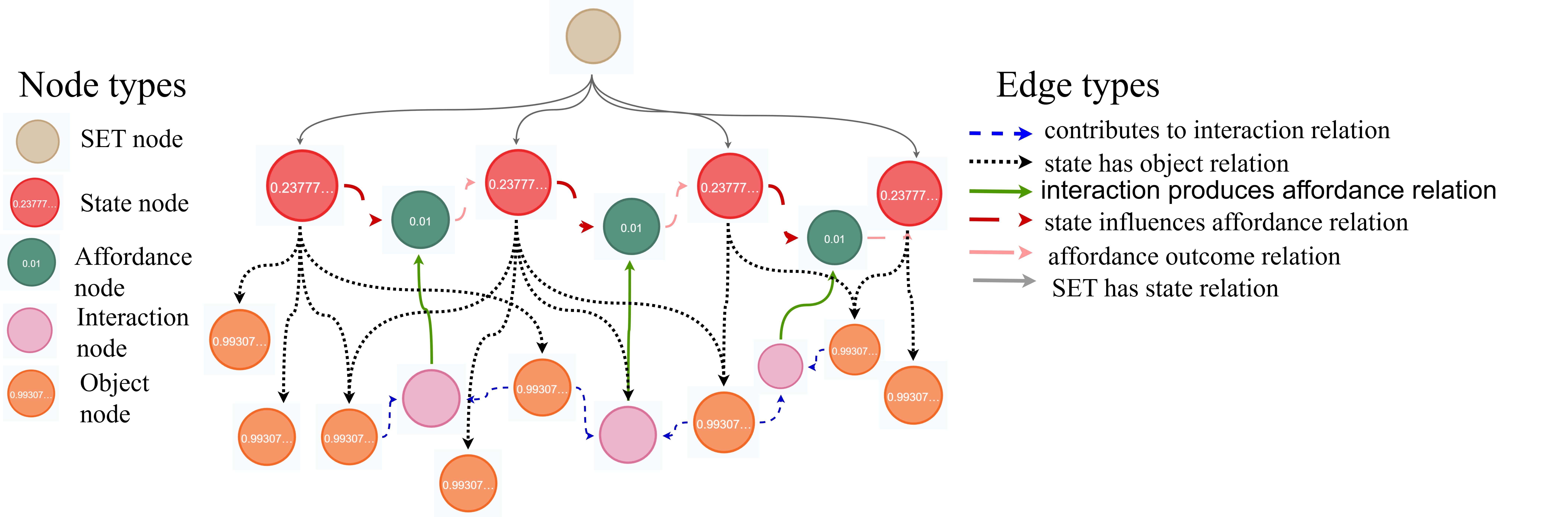}
    \caption{{ An example of the Hierarchical Structure of a SET Graph, illustrating Multi-Level Relational Dependencies.} At the lowest level, interactions (light-filled circular nodes) depend on objects (darker circular nodes). Moving upward, affordances (smaller darker nodes) emerge from interactions, which then influence states (highlighted in red with distinct borders) which in turn influence the next affordance. At the highest level, the SET node connects to the structure through temporal state dependencies. The edges indicate relationships such as object dependencies, contributions, and effects. Varying line styles help differentiate these relationships. This hierarchical structure showcases how task execution is represented by linking objects, interactions, affordances, and states.}
    \label{fig:all_lvls}
\end{figure}

\begin{figure}[h]
    \centering
    \includegraphics[width=1\linewidth]{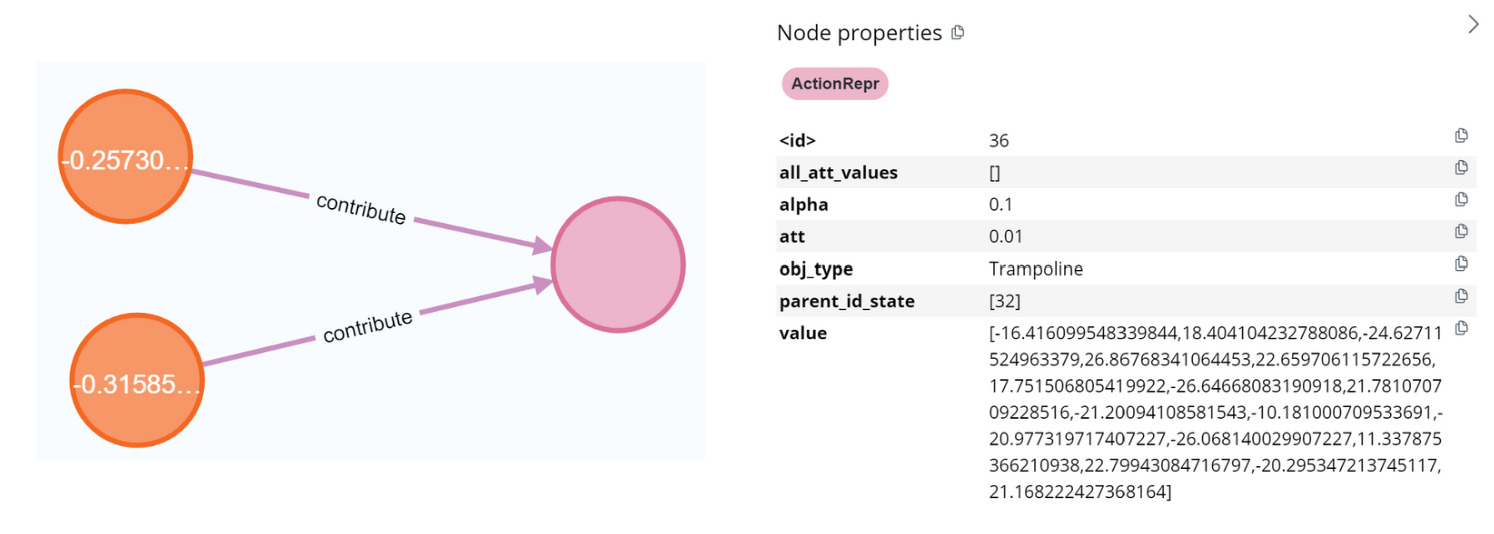}
    \caption{Representation of two objects interacting. Interactions are characterized not only by their execution , but also by the relationships they share with the objects involved. On the right, the different parameters of an Interaction are shown such as the type, and the vector representation. Representation obtained from the Neo4j \citep{lal2015neo4j} graph database interface.}
    \label{fig:obj_act}
\end{figure}

\paragraph{\textbf{Level 1: low-level abstractions (objects and interactions)}}
The lowest level of the hierarchy captures individual \textbf{objects} (Fig.~\ref{fig:obj_act} nodes in orange) and \textbf{interactions representations} (Fig.~\ref{fig:obj_act} node in pink), which form the fundamental units in the environment (Fig.~\ref{fig:obj_act}). 

The relationship between objects and interactions is fundamental to model structured learning. Fig.~\ref{fig:obj_act} illustrates how objects interact with each other, emphasising that an interaction is defined not only by its motion dynamics but also by the entities involved in its execution.

We have employed a combination of object detection and interaction representation learning techniques to construct these different representations of objects and interactions. 

\begin{enumerate}
    \item {Object Detection for extraction of object nodes}

Objects within each episode are extracted using the SAM (Segment Anything Model) object detection model \cite{kirillov2023segment}, which identifies and classifies entities in the environment. A process of checking whether the object exists already is carried out using a cosine similarity function.
\begin{equation}
    \text{Similarity}(o_{\text{new}}, o_i) = \frac{\mathbf{v}_{\text{new}} \cdot \mathbf{v}_i}{\|\mathbf{v}_{\text{new}}\| \|\mathbf{v}_i\|}, \quad \forall i \in \text{Memory} 
\end{equation}

\begin{equation}
    \text{If } \max_i \, \text{Similarity}(o_{\text{new}}, o_i) < \tau, \text{ add } o_{\text{new}} \text{ as a new node.}
\end{equation}

\item {Learning Interactions Representations using ConvLSTM for the extraction of interaction nodes}

An encoder is used to model object interactions effectively, capturing their temporal and spatial dynamics within the environment. While traditional approaches like the Neural Statistician have been utilized to learn latent representations for unordered datasets by generating a posterior \( q(c|D) \) over the context \( c \) of each dataset \( D \), their application to environments with high task similarity, such as CREATE, reveals notable limitations. Specifically, the shared context \( c \) across episodes often results in poor separability between distinct tasks, hindering the generalisation of representations.

We use a ConvLSTM-based encoder to address the challenges of differentiating between similar objects and their interactions across tasks. This encoder is designed to capture the temporal dependencies inherent in sequences of object interactions. By encoding temporal dynamics, the ConvLSTM model provides a richer and more structured representation of the task environment. Furthermore, to ensure that the embeddings effectively distinguish between similar and distinct interactions, we couple the ConvLSTM with a triplet loss function. The triplet loss enforces separation in the embedding space by bringing embeddings of similar interactions closer together while pushing embeddings of distinct ones further apart. As such we were able to produce robust and discriminative interaction representations, forming a foundational component for subsequent hierarchical encoding.

 Let \(\mathcal{F}_a = \{f_1, f_2, \dots, f_n \mid \text{all frames associated with interaction } a\}\) represent the set of frames associated with a specific interaction \(a\). For a given frame \(x_a \in \mathcal{F}_a\), we denote \(x_p \in \mathcal{F}_a\) as a positive sample (a frame associated with the same interaction \(a\)) and \(x_n \notin \mathcal{F}_a\) as a negative sample (a frame associated with a different interaction). 

The triplet loss \(L_{\text{triplet}}\) is defined as:
\begin{equation}
    L_{\text{triplet}} = \max(0, d(f(x_a), f(x_p)) - d(f(x_a), f(x_n)) + \alpha)
\end{equation}

Where \(f(x)\) is the embedding function that maps a frame into the embedding space, \(d(\cdot)\) is a distance function (e.g., Euclidean distance), and \(\alpha\) is the margin. In our experiments, we set \(\alpha = 0.6\) after a hyper-parameter search of values ranging from 0.1-1.0. The final value provided optimal separation between embeddings of similar and distinct interactions. 

This margin value helps the network distinguish subtle differences between interactions while maintaining adequate separation between positive and negative pairs.

Using this approach, we observed a clear separability in the embedding space. We applied K-means clustering to the learned embeddings and revealed the distinct clusters corresponding to different types of interactions. In the environment used (CREATE), all interactions involve a ball object interacting with various other objects. Thus, our analysis focuses on the second object, as it determines the dynamics of the interaction.

While clustering separability is not perfect, the learned representations exhibit an emergent understanding of interaction similarity. For instance, interactions involving swings or trampolines may occasionally overlap in certain clusters, yet the model consistently groups bouncy objects together, distinguishing them from interactions involving fixed objects (Fig. \ref{fig:bouncy_cluster}). This behaviour highlights the network's ability to capture underlying object properties that are essential for solving tasks in the CREATE environment.

\begin{figure}[h]
    \centering
    \includegraphics[width=\linewidth]{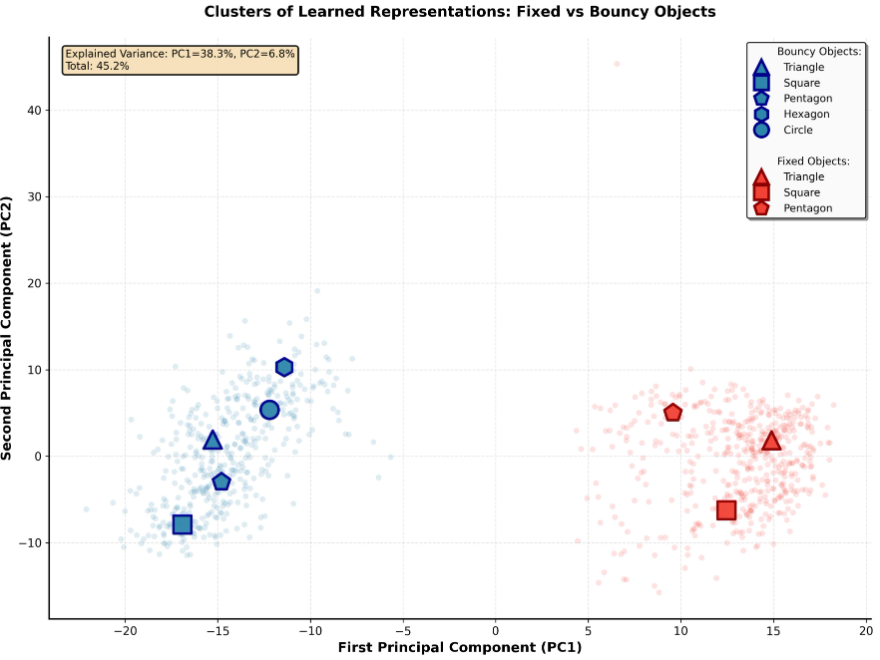}
    \caption{Visualisation of the interaction encoder’s ability to learn semantically meaningful representations. PCA projection of interaction embeddings with overlaid K-means cluster centroids. Light points represent individual samples, while larger markers denote centroids, with color indicating physical property (Bouncy vs. Fixed) and marker shape representing object type. A clear separation emerges along PC1 (38.3\% variance), showing that physical interaction type is the dominant latent factor learned by the encoder. The encoder has successfully
learned to distinguish between these two functionally critical classes of interactions from visual data alone, a key prerequisite for building the SET hierarchy.
    }
    \label{fig:bouncy_cluster}
\end{figure}

\end{enumerate}

\paragraph{\textbf{Level 2: Affordances and State Transitions}}

\begin{figure}[!ht]
    \centering
    \includegraphics[width=\linewidth]{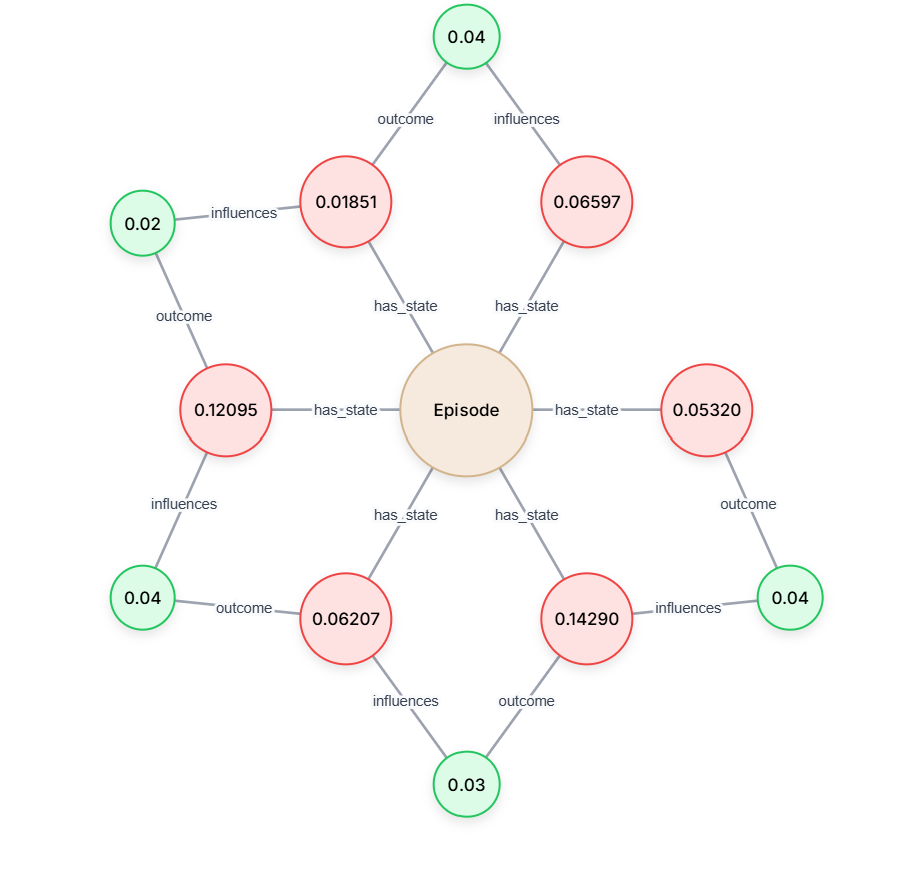}
    \caption{Illustration of the hierarchical structure in SETLE representing the second level of the hierarchy that consists of states (red, from the middle node the first set of connections) connected by affordances (green, the outer-most nodes), illustrating the transitions between states via affordances.}
    \label{fig:second_lvl}
\end{figure}

At the intermediate level, we represent states as heterogeneous graphs, denoted as \( G_{it}' \) (See Section \ref{sect:hier_mem}) at time \( t \). Each state comprises multiple interacting objects, connected through interaction representations that shape the decision-making process (Fig.~\ref{fig:all_lvls}). \textbf{Affordances} serve as critical links between states, interactions, and trajectories by capturing the transition dynamics (Fig.~\ref{fig:second_lvl}).

Affordances encode both immediate effects and long-term consequences, enabling the model to prioritise interactions that contribute to the successful completion of the task.

By explicitly modelling affordance relationships,
we improve the system’s ability to reason about sequential dependencies and facilitate adaptive decision-making in complex environments that move beyond traditional action-reward, capturing richer contextual information.

\paragraph{\textbf{Level 3: High-Level Abstraction of Structurally Enriched Trajectories (SETs)}}
The topmost layer in the hierarchical structure corresponds to the overall SET representation. In its raw form, a SET is defined by the underlying levels, comprising states, affordances, interactions, and objects (see Section \ref{sec:set}). However, utilising full graph structures for downstream tasks is computationally impractical. To address this, we learn a \textbf{latent representations} of SETs (see Section \ref{sect:set_enc}), effectively compressing the rich hierarchical information into a compact embedding. This latent representation forms the core of our proposed framework, enabling efficient and scalable utilisation in downstream tasks without losing critical task-relevant information.

The hierarchical graph structure obtained, illustrated in Figure \ref{fig:all_lvls}, captures the relationships between different levels of abstraction within an episode. The visualisation shows a diverse set of nodes, each representing a unique component of the episode, including objects, states, interactions, and affordances, linked through various relationships.

In the graph, nodes are colour-coded according to their roles:
\begin{enumerate}
    \item Orange nodes represent individual objects detected in the environment, each connected by edges that denote has\_object relationships to higher-level states or actions.
    \item Pink nodes represent interaction representations
    \item Red nodes represent states, encapsulating the conditions of the environment at specific time steps. Each state node can be further decomposed into low-level objects and interactions, creating a detailed snapshot of dynamics within each time frame.
    \item Green nodes denote affordances, which are essential links that describe how specific actions lead to transitions from one state to another. Affordances, as connections between actions and resulting states, also encapsulate reward information, guiding the agent towards interactions that are likely to yield positive outcomes.
\end{enumerate}

\subsection{SETLE: SET encoder} \label{sect:set_enc}

This section details the methodology for encoding SET subgraphs, including contrastive learning techniques and triplet loss, which structure and differentiate trajectory representations to support robust learning for downstream tasks such as reinforcement learning.
Building upon the structured representations introduced in Section \ref{sect: hier_graph_constr}, we now focus on encoding episodic graphs into compact latent vectors, ensuring that trajectory representations capture essential structural dependencies that facilitate effective generalisation and decision-making.

\begin{figure}
    \centering
    \includegraphics[width=1\linewidth]{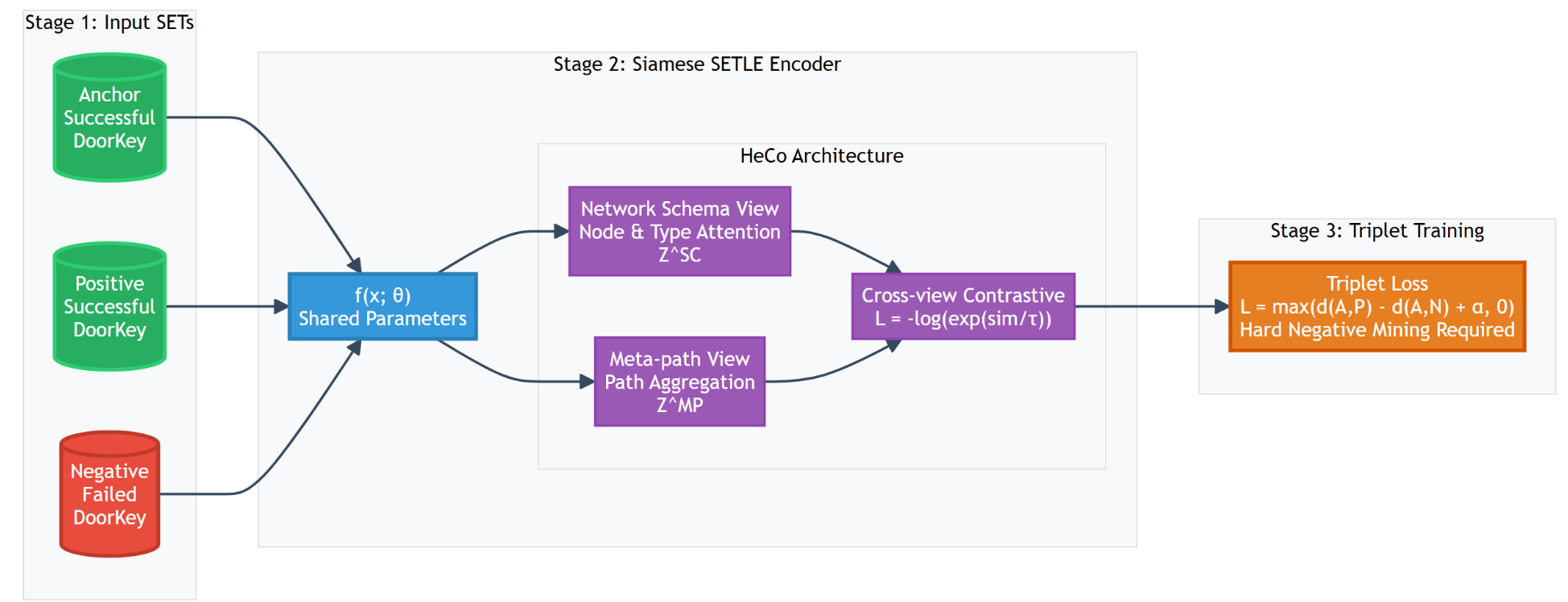}
    \caption{Overview of the SETLE encoder training pipeline. \textbf{Stage 1:} A triplet of SETs (an anchor, a positive, and a negative sample) is used as input. \textbf{Stage 2:} All three subgraphs are processed by an encoder with shared parameters (f(x; $\theta$)), which uses the HeCo architecture to generate embeddings from both the Network Schema and Meta-path views. These views are regularized by an internal cross-view contrastive loss. \textbf{Stage 3:} The final embeddings are used to compute the Triplet Loss, which pulls the anchor and positive samples together in the latent space while pushing the negative sample away.}
    \label{fig:encoder_pipeline}
\end{figure}

A graph encoder is trained with triplet loss to ensure that episodes with similar trajectories cluster together, distinguishing successful task executions from unsuccessful ones (Fig.\ref{fig:encoder_pipeline}).

\subsubsection{Graph Encoder Architecture Based on Heterogeneous Co-contrastive Learning (HeCo)}
To operate on the heterogeneous graphs and leverage the meta-path relationships formally defined in our Hierarchical Memory Structure (Section \ref{sect:hier_mem}), we employ a specialized graph encoder. Specifically, we adapt the Heterogeneous Co-contrastive Learning (HeCo) architecture that transforms graph-structured data into meaningful embeddings, preserving the relationships and dependencies between entities. Unlike standard feature extractors, graph encoders operate over relational data, using the graph topology to learn structured representations.

Graph encoders rely on message passing, a core operation in Graph Neural Networks (GNNs) \citep{scarselli2009gnn}, where each node iteratively aggregates information from its neighbours to refine its representation \citep{gilmer2017neural}. In its general form, message passing is defined as:

\begin{equation}
    H^{(l+1)} = \sigma \left( \tilde{D}^{-1/2} \tilde{A} \tilde{D}^{-1/2} H^{(l)} W^{(l)} \right)
\end{equation}

Where:
\begin{itemize}
    \item $H^{(l)} \in \mathbb{R}^{n \times d^{(l)}}$: Node representations at the $l$-th layer.
    \item $d^{(l)}$: Hidden dimension of the $l$-th layer.
    \item $W^{(l)} \in \mathbb{R}^{d^{(l)} \times d^{(l+1)}}$: Learnable weight matrix for the $l$-th layer.
    \item $n$: Number of nodes in the graph.
    \item $\tilde{A} = A + I$: Adjacency matrix $A$ with added self-loops (identity matrix $I$).
    \item $\tilde{D}$: Degree matrix of $\tilde{A}$.
    \item $\sigma$: Nonlinear activation function.
\end{itemize}

This process allows nodes to incorporate structural context from their neighbours, making graph encoders particularly effective for tasks that require relational reasoning.

Most existing approaches either encode entire heterogeneous graphs or focus on subgraph encoding for homogeneous data. Full-graph encoding enables global feature propagation, but it does not preserve localised dependencies. We model the hierarchical memory in \textbf{SETLE} as a collection of SET subgraphs, unlike standard GNNs that operate on a global graph, SETLE learns representations for independent trajectory (SET) subgraphs, where each SET node aggregates information solely from its connected states, interactions, objects, and affordances. By structuring trajectories (SETs) as subgraphs within the larger task graph (see Section \ref{sect:hier_mem}), SETLE captures common elements, such as objects and interactions, at lower levels, allowing \textbf{localised learning while maintaining global coherence across tasks}.

Building on the principles of HeCo (introduced in the Related Work section), we adapt its contrastive learning approach for episodic graphs. We treat each SET as a subgraph, with the \textit{SET node} acting as the central node. The central node refers to the primary node within a graph that aggregates information from surrounding nodes through both meta-path and schema-based message passing, ensuring that its embedding captures local structural dependencies as well as high-order semantic relationships within the heterogeneous graph. 

The encoding operation can be formally represented as:
\begin{equation}
    \text{SETLE-encoder}(G_{SET}) \rightarrow z_{SET}
\end{equation}

where \( z_{SET} \) is the latent representation of a Structurally Enriched Trajectory \( G_{SET} \).

To train the graph encoder, we use a triplet loss and a hybrid loss function to organise SET embeddings. Successful SETs from the same task are drawn together, while unsuccessful SETs or SETs from different tasks are pushed apart. This allows us to differentiate between successful and unsuccessful strategies, enhancing generalisation.
To ensure that episodes with similar trajectories are encoded closely in the latent space, we employ a triplet loss as such:
\begin{equation}
    L_{triplet}^{SET} = \max(0, d(g(y_a), g(y_p)) - d(g(y_a), g(y_n)) + \alpha)
\end{equation}

Here:
\begin{itemize}
    \item \( g(y) \) is the SET embedding function.
    \item \( y_a \), \( y_p \), and \( y_n \) are the anchor, positive, and negative SET samples.
    \item \( d \) is a distance metric.
    \item \( \alpha \) is a margin parameter.
\end{itemize}

By minimising this loss, the model learnt to align trajectory-specific embeddings, ensuring that episodes with similar trajectories (i.e., trajectories belonging to the same tasks) remained close while distinguishing those with different results.

To further improve the quality of SET representations, a modified \textbf{hybrid loss function} that combines the triplet loss with a cross-view contrastive loss is also used. This approach not only pushes similar SET embeddings together but also aligns two graph views: meta-path and schema, ensuring consistency across different perspectives of the episodic graph.

The two key views from HeCo are adapted to SETs as follows:

\begin{enumerate}
    \item Network Schema View

    In the network schema view, we encode the structural dependencies of each episode by considering relationships between node types (e.g., state, interaction, affordance) around the central SET node. This view allows us to capture local patterns directly connected to the episode.

    \item Meta-path View

    The meta-path view captures high-order semantic relationships by encoding dependencies across sequences of connected nodes, allowing the model to learn relationships that span multiple levels of the episode's hierarchy. The following steps outline the process of computing node embeddings under the meta-path view:

\begin{enumerate}
    \item For a given node "i" and a set of \( M \) meta-paths \( \{ P_1, P_2, \dots, P_M \} \), which represent high-order semantic relationships (e.g., \( \texttt{SET-St-Obj-Inter-Aff} \) or \( \texttt{SET-St-Aff-St} \)):
    \begin{enumerate}
        \item For each meta-path \( P_n \), we apply a meta-path specific Graph Convolutional Network (GCN) to encode its characteristics:
        \begin{itemize}
            \item Calculate the projected features \( h_i \) and \( h_j \) for nodes "i" and "j" in the meta-path.
            \item Use degree information \( d_i \) and \( d_j \) for nodes "i" and "j" to compute the updated embedding \( h^{P_n}_i \) based on the meta-path's semantic similarity.
        \end{itemize}
    \end{enumerate}
    \item For all \( M \) meta-paths, we obtain a set of embeddings \( \{ h^{P_1}_i, \dots, h^{P_M}_i \} \) for node "i."
    \item Use \textbf{semantic-level attention} to combine these embeddings into the final embedding \( z^{mp}_i \) under the meta-path view:
    \begin{itemize}
        \item Calculate the weight of each meta-path \( P_n \) for node "i."
        \item Determine the importance of each meta-path in the embedding fusion.
        \item Perform a weighted sum of meta-path embeddings to produce the final node embedding.
    \end{itemize}
\end{enumerate}

The selected meta-paths, \( \texttt{SET-St-Obj-Inter-Aff} \) and \( \texttt{SET-St-Aff-St} \), are designed to capture high-order semantic relationships within each episode in the RL environment. 

\begin{itemize}
    \item \textbf{\( \texttt{SET-St-Obj-Inter-Aff} \) (SET - State - Object - Interaction - Affordance):} 
    \begin{itemize}
        \item This meta-path reflects the hierarchical progression from the episode as a whole, through individual states, down to specific objects and actions, ultimately linking to affordances.
        \item By following this sequence, SETLE can capture the causal chain of events where each \textit{state} within a \textit{SET} is defined by its constituent \textit{objects} and the \textit{interactions} involving those objects. The inclusion of \textit{affordances} in the meta-path allows the model to understand how specific object interactions influence the likelihood of transitioning to desirable or goal-oriented states.
        \item This path is crucial for learning how particular combinations of objects and their interactions contribute to successful results, enabling the model to capture the semantic importance of specific object relationships in each episode. It supports the agent in recognising patterns of interaction that are effective across similar tasks, improving object selection and result prediction in new scenarios.
    \end{itemize}

    \item \textbf{\( \texttt{SET-St-Aff-St} \) (SET - State - Affordance - State):}
    \begin{itemize}
        \item This meta-path emphasises the transitions between states within a SET, specifically focussing on how each \textit{affordance} (a state-action-state link with associated rewards) facilitates state changes.
        \item By encoding this meta-path, SETLE captures the dynamics of state transitions within the trajectory. The affordance links provide a representation of the agent's actions and the effects, highlighting which transitions are likely to lead to success or failure.
        \item This meta-path is valuable for modelling temporal dependencies in SETs, as it reveals the sequences of states the agent encounters. Understanding these transitions is critical for building a representation that reflects the path to the goal, rather than just the individual states, thus helping the agent generalise across different task setups that share similar transition dynamics.
    \end{itemize}
\end{itemize}

Together, these meta-paths allow us to capture both \textbf{static relationships} (e.g., object interactions within a state) and \textbf{dynamic relationships} (e.g., state-to-state transitions driven by affordances) in the SET graph. By incorporating both aspects, we can construct a more comprehensive and semantically meaningful representation of each trajectory, allowing it to effectively generalise across tasks and improve decision-making in complex reinforcement learning environments.

\end{enumerate}

Considering these two views we define the hybrid loss as:
\begin{equation}
  L_{\text{hybrid}} = \lambda L_{\text{triplet}}^{\text{SET}} + (1 - \lambda) L_{\text{contrastive}}^{\text{view}}  
\end{equation}

Here:
\begin{itemize}
    \item \( L_{\text{triplet}}^{\text{SET}} \) organises SET embeddings in the latent space, as previously defined:
    \begin{equation}
            L_{\text{triplet}}^{\text{SET}} = \max(0, d(g(y_a), g(y_p)) - d(g(y_a), g(y_n)) + \alpha)
    \end{equation}
    where \( g(y) \) is the SET embedding function, \( y_a \), \( y_p \), and \( y_n \) are the anchor, positive, and negative SET samples, \( d \) is a distance metric, and \( \alpha \) is a margin parameter.

    \item \( L_{\text{contrastive}}^{\text{view}} \) aligns the meta-path and schema view embeddings to ensure consistency across views:
    \begin{equation}
         L_{\text{contrastive}}^{\text{view}} = -\log \frac{\exp(\text{sim}(z_{\text{mp}}, z_{\text{sc}}) / \tau)}{\sum_{z' \in \mathcal{Z}} \exp(\text{sim}(z_{\text{mp}}, z') / \tau)}
    \end{equation}

    Where:
    \begin{itemize}
        \item \( z_{\text{mp}} \) and \( z_{\text{sc}} \) are the meta-path and schema view embeddings for the same SET.
        \item \( \text{sim}(\cdot, \cdot) \) is a similarity function, such as cosine similarity.
        \item \( \tau \) is the temperature scaling parameter.
        \item \( \mathcal{Z} \) represents the set of embeddings, including positive and negative samples.
    \end{itemize}
    \item \( \lambda \) is a hyperparameter that controls the balance between the triplet loss and the contrastive loss, with \( 0 \leq \lambda \leq 1 \).
\end{itemize}

By combining these objectives, the hybrid loss ensures that SET embeddings are both task-specific (via the triplet loss) and consistent across graph views (via the contrastive loss). 

{
\subsubsection{Implicit Knowledge Propagation via Shared Embeddings}
\label{sec:implicit_propagation}
As established in Section \ref{sect:hier_mem}, shared nodes serve as the "bridges" for knowledge transfer. This section details the implicit learning mechanism that makes this transfer possible during the encoder's training. A key feature of SETLE's design is the implicit propagation of knowledge across episodic subgraphs during the training of the encoder. While there is no direct message-passing operating between different subgraphs during a single forward pass, knowledge is transferred through the shared, updatable embeddings of common nodes during the iterative training process.

The process, is implemented as follows:
\begin{itemize}
    \item \textbf{Independent Subgraph Encoding:} At each training step, a triplet of self-contained episode subgraphs (an anchor, a positive, and a negative sample) is loaded. The features and graph structure for each are prepared independently before being passed to the model, ensuring message passing is confined within the boundaries of each individual trajectory.
    \item \textbf{Shared Node Embeddings:} A single, globally-stored feature embedding table is maintained within the HeCo model for each node type. This updatable table is referenced via unique node IDs every time a concept (e.g., the `red ball` object) appears, ensuring the same vector representation is used across all episodes.
    \item \textbf{Update via Backpropagation with Gradient Accumulation:} During training, gradients from the hybrid loss function, calculated on the output of the episode triplet, flow back via the \texttt{loss.backward()} call. These gradients are accumulated over several steps (e.g., 20 steps in our experiments). The model's parameters, including the shared embedding layers, are then updated synchronously. This use of gradient accumulation allows the model to aggregate signals from multiple triplets before updating the shared embeddings, leading to more stable learning.
    \item \textbf{Implicit Knowledge Transfer:} The model's parameters are persistent across the entire training loop. Therefore, an update to a shared node's embedding from a triplet processed at step \(i\) is immediately available when that node appears in another triplet at a subsequent step \(i+k\). This iterative refinement is how learning from one experience implicitly informs the processing of another. To use an analogy, the shared nodes function like recurring characters in a narrative; their state (embedding) is shaped by the events of one chapter (an episode triplet) and they carry that updated state with them into subsequent chapters.
\end{itemize}
}
\subsubsection{Collecting data for training}
To populate the Hierarchical Memory Structure ($\mathcal{M}$) with the individual SET subgraphs ($\mathcal{G}_i$) needed for the training process described above, and specifically to generate the anchor, positive, and negative samples required by our hybrid triplet loss, we first generate a corpus of SETs from agent interactions. These trajectories are then labelled by their task outcome (success or failure) to create a dataset for training, following these steps:

\begin{enumerate}
    \item Random exploration: We begin with an exploration phase where the agent performs multiple tasks within the CREATE and MiniGrid environment. During this phase, a random policy agent operates in the environment until it reaches the goal state. For tractability, we have limited the number of steps the agent can take in an environment to 9, if it has not reached a goal state by then, the SET is labelled as unsuccessful. We repeat the process until sufficient successful trajectories have been collected for training.
    \item SET Labelling: After each episode, we assess whether the agent successfully completed the task. If the episode ends with the agent achieving the goal state, the episode SET is labelled as: success; otherwise, it is labelled as: failure. Alongside the label, we store relevant metadata about the setup, actions taken, and objects encountered, allowing for later validation and analysis. Successful SETs are stored in the agent's memory (see Section \ref{sect:hier_mem}), completed with the inventory of objects, interactions, and state transitions that led to the successful completion.
\end{enumerate}

Once a sufficient number of trajectories have been collected, we train the SETLE encoder using a triplet loss function.

\section{Multi-Environment Experiments}
{
This section evaluates the effectiveness of SETLE’s hierarchical graph encoder in encoding task-specific information across environments with different complexity profiles. The evaluation is structured around two key experimental domains: the CREATE environment, which tests perceptual and physical complexity, and the MiniGrid environment, which tests logical and planning complexity.

While the CREATE environment may appear visually simple, it provides a unique advantage as a testbed. It maintains a low computational overhead, yet incorporates complex, continuous physics dynamics —such as a ball being propelled by a fan or bouncing off different surfaces—that are often absent in grid-world settings. This unique combination makes it an ideal environment for studying foundational principles like tool use and affordance learning, without requiring the extensive processing power of high-fidelity simulators. In contrast, MiniGrid offers a simplified, discrete setting that tests SETLE’s capacity for abstraction and multi-step planning under partial observability.

The evaluation is structured into two key sets of experiments: margin sensitivity analysis and ablation studies.
In the first set of experiments (S1), we assessed how different margin values in the triplet loss function affect the quality of learnt embeddings. By varying the margin parameter, we analysed its influence on the clustering of success and failure SETs in the embedding space. This experiment provided insights into the optimal margin settings that enhanced separability while maintaining robust intra-cluster cohesion.

The second set of experiments (S2) consisted of ablation studies to assess the impact of hierarchical structures and relational dependencies in SETLE’s SET design. We systematically modified or removed key components, such as hierarchical organisation, sequentiality, and affordance-based state transitions, to understand their contributions to SET representation. By comparing the clustering performance of modified versions of SET's structure against its complete hierarchical design, we demonstrated the necessity of structured graph-based encoding for effective trajectory learning.

In the following subsections, we first present the detailed results from the CREATE environment, which has a richer hierarchical structure including interaction dynamics. We then present the results from MiniGrid, where the graph structure is simplified, as interaction-level dynamics cannot be extracted meaningfully. This allows us to test SETLE’s capacity for abstraction under reduced representational complexity.
}
{
\subsection{Experimental Setup and Implementation Details}
This section provides a comprehensive overview of the model architecture, training configuration, and hyperparameters used in our experiments.

Our model architecture is based on HeCo (Heterogeneous Co-contrastive Learning) with a GraphSAGE encoder. The encoder uses a 3-layer architecture with 64 hidden units per layer, producing node embeddings with 64 hidden dimensions. The architecture is configured to process 2 meta-paths, as defined in Section 3.3.1, and the attention mechanism employs 8 heads with 64-dimensional key/value vectors. To ensure reproducibility, all random seeds for PyTorch, NumPy, and Python's random were fixed at the beginning of each run. We used the PyTorch default initialization methods: Xavier/Glorot for linear layers and He for convolutional layers.

For training, we used the Adam optimizer with a learning rate of 0.001 and a weight decay (L2 regularization) coefficient of 1e-4. To simulate a larger batch size and stabilize training, gradients were accumulated over 20 steps before each parameter update. The encoder was trained for up to 200 epochs, with early stopping based on validation loss. The training objective was a hybrid loss function that combines a triplet objective with a cross-view contrastive loss.

To ensure the robustness of our findings for the encoder, performance was assessed over \textbf{8 different runs}. For the downstream reinforcement learning experiments presented in Section 5, each agent configuration was trained and evaluated across \textbf{3 different random seeds}. The data for the encoder training was sourced from agent interactions and divided using a standard \textbf{80-20 train/validation split}. The model was trained on the training set with \textbf{early stopping} based on performance on the validation set to prevent overfitting.

\subsection{Component Analysis and Design Justification}
\label{sec:component_analysis}
Beyond evaluating the complete framework, it is important to justify the design choices for its critical prerequisite components. Rather than performing traditional ablations on these foundational parts, this section analyses their contribution by comparing alternative approaches and highlighting the framework's modularity across different environments.

\paragraph{ConvLSTM Encoder for Interaction Dynamics}
As introduced in Section \ref{hier_sec}, the ConvLSTM encoder is designed to capture rich, temporal interaction dynamics from visual input, which is essential in physics-based environments like CREATE. Our multi-environment evaluation serves as a natural experiment for this component's role. In the symbolic MiniGrid environment, where there are no complex visual dynamics to model, the ConvLSTM and its corresponding `ActionRepr` nodes are deliberately omitted from the graph structure. The framework's success in both domains demonstrates that the ConvLSTM is a powerful but \textbf{modular component}: its contribution is to enable the processing of complex perceptual dynamics when they exist, but the core SETLE framework is robust enough to function without it in logically-driven domains.

\paragraph{Object Discovery: Foundational Models for Robust Perception}
A prerequisite for our framework is a robust object discovery module that can generalize across different visual domains without environment-specific retraining. In our initial research, we evaluated traditional object-centric models, specifically \textbf{Slot Attention}. While effective on its training data, we found that it failed to generalize to the visual styles of our RL environments, producing unstable segmentations even after fine-tuning. In contrast, a foundational model—the Segment Anything Model (SAM), provided excellent, zero-shot object segmentation across all tested domains without any modification. This design choice was critical for ensuring the robustness and generalizability of the entire SETLE pipeline, which would otherwise be bottlenecked by a brittle, domain-specific perception module.
}
\subsection{S1: SET Embedding Analysis: Margin and Loss Function Impact}

The first set of experiments aimed to evaluate the ability of SETLE’s hierarchical graph encoder to represent enhanced trajectory information effectively. The primary focus was on assessing whether the embeddings generated by SETLE can distinguish between successful and unsuccessful episodes within and across tasks. Using episodic data collected from the CREATE environment, each episode was encoded as a heterogeneous graph (SET) that comprises objects, actions, states, and affordances.

We investigated how well SETLE's SET embeddings captured the underlying task dynamics by analysing clustering quality using various configurations of hyper-parameters and different loss functions.

The evaluation focused on specific pparameters, including:

\begin{itemize} \item \textbf{Margin Values:} We experimented with margin values for the triplet loss, specifically testing values of [0.1, 0.2, 0.5, 1.2, 1.5]. This parameter regulates the separability of embeddings in the latent space by controlling how far apart unsuccessful SETs are pushed from successful ones. \item \textbf{Loss Functions:} Two types of loss functions were employed: \begin{enumerate} \item \textbf{Classical Triplet Loss:} Ensures SET embeddings for similar episodes are close, while those for dissimilar SETs are further apart. \item \textbf{Hybrid Triplet Loss:} Combines ta triplet loss with a cross-view contrastive loss. The contrastive component aligns embeddings from the schema and meta-path views, while the triplet component enhances task-based separability. \end{enumerate} \end{itemize}


The following metrics were used to evaluate the separability of success and failure clusters:

\begin{itemize}
    \item \textbf{Silhouette Score}: Measures the cohesion and separation of clusters, the values can range between -1 and 1, with higher values indicating more distinct clusters. Values closer to 1 suggest that episodes are well-clustered within their respective success or failure groups.
    
    \item \textbf{Davies-Bouldin Index (DBI)}: Quantifies the average similarity ratio of each cluster with its most similar cluster, with lower values indicating better clustering performance and less overlap between clusters.
    
    \item \textbf{Dunn Index}: Assesses the ratio between the minimum inter-cluster distance and the maximum intra-cluster distance, with higher values representing greater separability and well-defined clusters.
\end{itemize}

\subsubsection{Margin Impact and Results}

\paragraph{\textbf{Effect of Margin Values}}

We evaluated the impact of varying margin values on the clustering quality of SET embeddings. The results provided insights into the cohesion and separation of clusters for both successful and unsuccessful episodes. High cohesion within success or failure clusters indicated that the embeddings effectively captured the underlying patterns of similar SETs, while high separation between these clusters reflected the model's ability to distinguish the dynamics of successful and unsuccessful episodes. These insights were critical for evaluating the suitability of different margin values in shaping the latent space for SET-specific encoding.

The clustering results were analysed using three standard metrics: \textbf{Silhouette Score}, \textbf{Davies-Bouldin Index (DBI)}, and \textbf{Dunn Index}. 

Table \ref{table:clustering_metrics} summarises the clustering performance across margin values. Higher margins (1.2 and 1.5) generally yielded better separability, reflecting clearer distinctions between successful and unsuccessful episodes. 

\begin{table*}[h!]
\centering
\caption{Clustering Results for Success and Failure for SET encodings at Different Margin Values}
\label{table:clustering_metrics}
\begin{tabular}{p{1.0cm} p{1.4cm} p{1.4cm} p{1.4cm} p{1.4cm} p{1.9cm} p{1.9cm}}
\toprule
\textbf{Margin} & \textbf{Silhouette Score (Success)} & \textbf{Silhouette Score (Failure)} & \textbf{DBI (Success)} & \textbf{DBI (Failure)} & \textbf{Dunn Index (Success)} & \textbf{Dunn Index (Failure)} \\
\midrule
0.1 & 0.7632 & 0.8531 & 0.4514 & 0.2319 & 2.2497 & 3.4662 \\
0.2 & 0.7959 & 0.8659 & 0.3205 & 0.2233 & 2.9952 & 3.3552 \\
0.5 & 0.7943 & 0.8798 & 0.3378 & 0.1771 & 2.5417 & 5.3031 \\
1.2 & 0.7588 & 0.8535 & 0.3603 & 0.2305 & 2.3577 & 3.5070 \\
1.5 & 0.8137 & 0.8990 & 0.2974 & 0.1548 & 2.7778 & 5.5415 \\
\bottomrule
\end{tabular}
\end{table*}

From the results in Table 1, it was shown that higher margin values generally improved clustering quality for both success and failure SETs. Notably:

\begin{itemize}
    \item The \textbf{Silhouette Score} showed a consistent improvement with increasing margin, peaking at a margin of 1.5, where the success and failure clusters were well-separated.
    \item The \textbf{Davies-Bouldin Index (DBI)} decreased as the margin increased, reaching its lowest value at a margin of 1.5. This suggested that higher margins reduced the overlap between clusters, leading to more distinct success and failure groupings.
    \item We observed an appreciable improvement in the \textbf{Dunn Index} with higher margins, especially for failure SETs, which achieved their highest separability at a margin of 1.5. This indicated that increasing the margin value results in more well-defined clusters, with greater distances between the success and failure groups.
\end{itemize}

Higher margin values proved necessary for effective clustering in this context because the CREATE environment is relatively simple, with low-resolution frames and limited visual complexity. In such environments, embeddings based on episode encodings can appear similar due to the simplicity of the scenes and fewer unique features across frames. Using higher margins (1.2 and 1.5), we ensured that embeddings for episodes with different outcomes or tasks are separated more distinctly, overcoming the limitations posed by the low frame quality and simplifying environment. 

\begin{figure}[!ht] \centering \includegraphics[width=\linewidth]{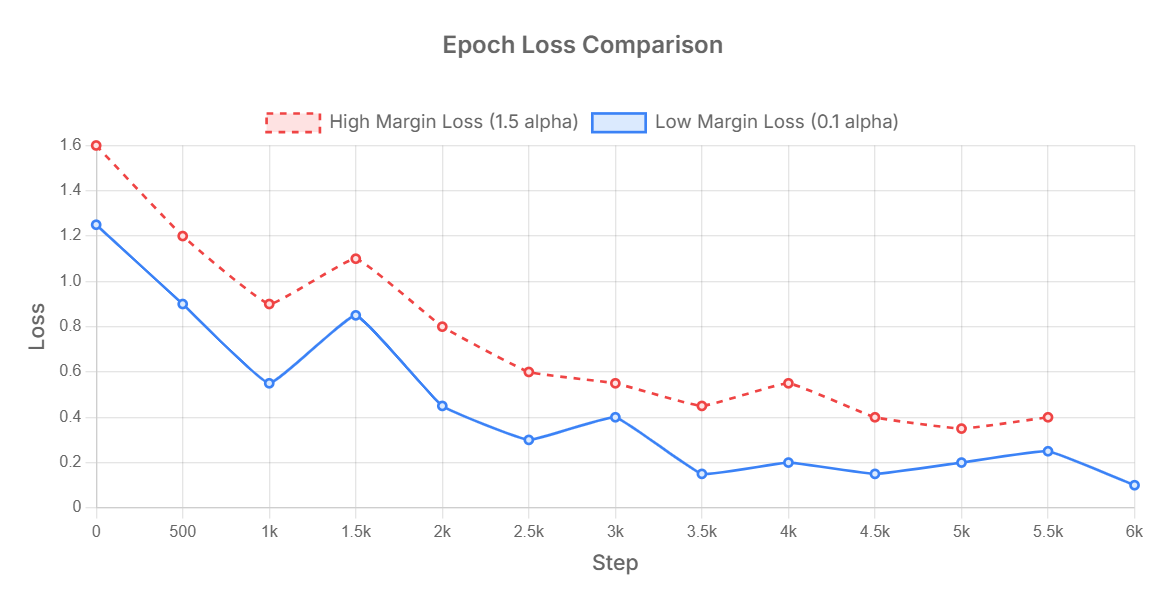} \caption{Comparison of training loss curves for high-margin ($\alpha$=1.5, red dashed line) and low-margin ($\alpha$=0.1, blue solid line) configurations of the loss. Both models demonstrate stable convergence, indicating that while the margin value is critical for the final separability of embeddings (see Table 1), it does not significantly hinder the model's ability to learn (Section 4.6).} \label{fig:loss} \end{figure}

In terms of epoch loss, we did not observe significant changes (see Section 4.6) with varying margin values Fig.\ref{fig:loss}. The overall epoch loss progression remained relatively stable, suggesting that while higher margins did not heavily impact convergence or loss reduction during training, they did have a positive effect on the quality of clustering in the latent space. In the cluster analysis below, we examined how these higher margin values enhanced episode separability, supporting generalisation and task-based decision-making. { It is important to note the trade-off between clustering quality and training stability. While our statistical analysis (Section 4.6) confirmed that lower margin values lead to a significantly lower final training loss (t(5)= -2.83, p=0.037), the results in Table 1 clearly show that higher margin values are superior for achieving the primary goal of this encoder: creating a well-separated latent space. This demonstrates a classic hyperparameter trade-off where a stronger constraint (a higher margin) improves the desired structural outcome at the cost of a slightly more difficult optimization process.}

\subsubsection{Visualization of Outcome Embeddings}

To complement the quantitative analysis of clustering metrics such as Dunn Index, DBI, and Silhouette Score, we employed K-means clustering and visualisation techniques to provide a qualitative assessment of the learnt SET embeddings. By reducing the dimensionality of the embeddings using PCA, we examined the spatial distribution of clusters in two dimensions, revealing the distinctiveness of task and result representations. These visualisations offered additional insights into how well SETLE's graph encoder captured the structural and outcome-specific features of the tasks, highlighting its ability to group similar episodes while maintaining separation between dissimilar ones. The tight grouping of points within each task cluster, along with minimal overlap between clusters of different tasks, indicated that the graph encoder has learnt meaningful representations that distinguish both the task type and the outcome.

\begin{figure}[!ht]
    \centering
    \begin{subfigure}[b]{1\textwidth}
        \includegraphics[width=\textwidth]{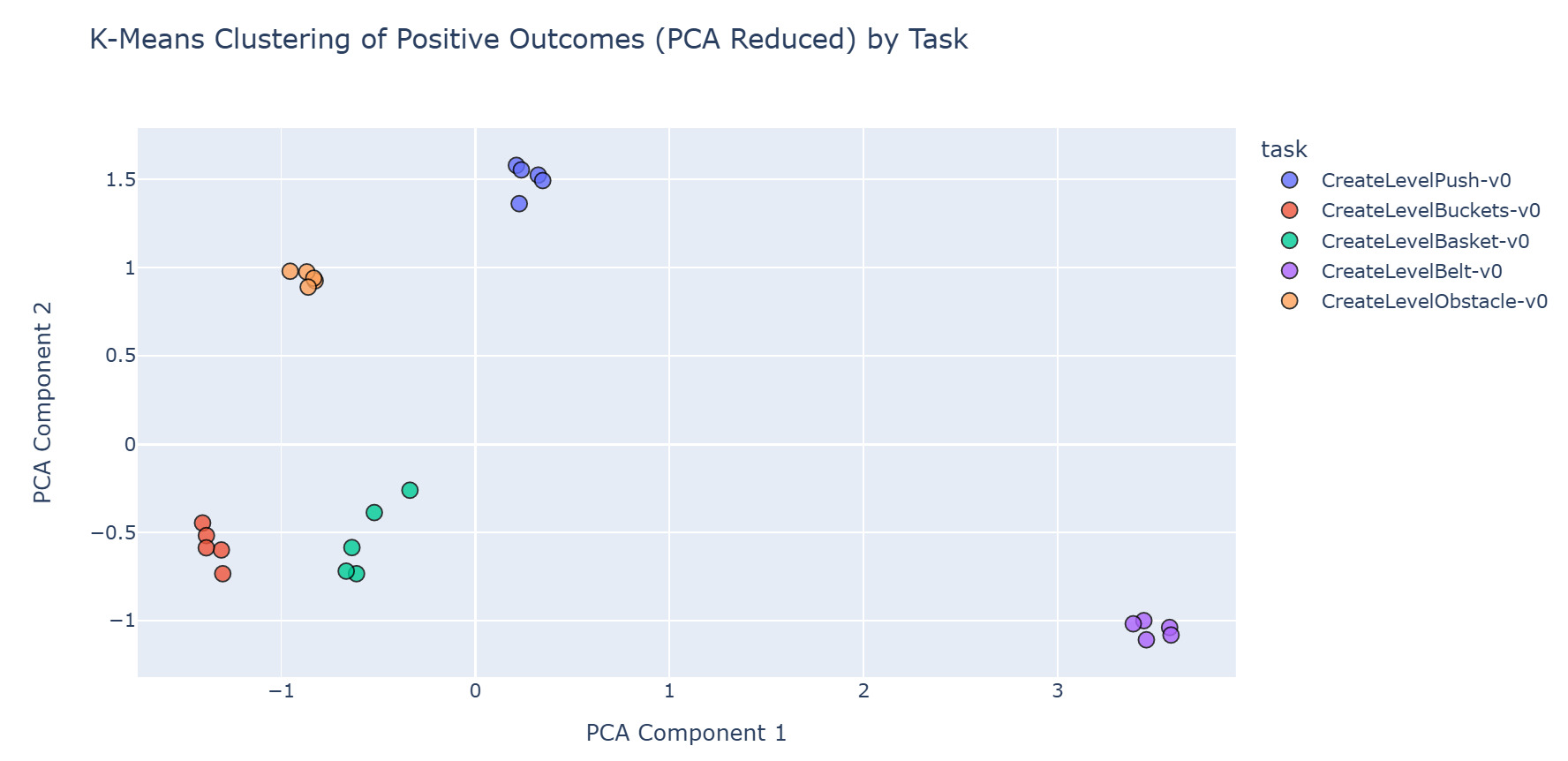}
        \caption{Task-level clusters: K-means Clustering of SET Encodings (PCA Reduced). Each point represents a SET in the CREATE environment, colour-coded by task.}
        \label{fig:cluster_res}
    \end{subfigure}
    \hfill
    \begin{subfigure}[b]{1\textwidth}
        \includegraphics[width=\textwidth]{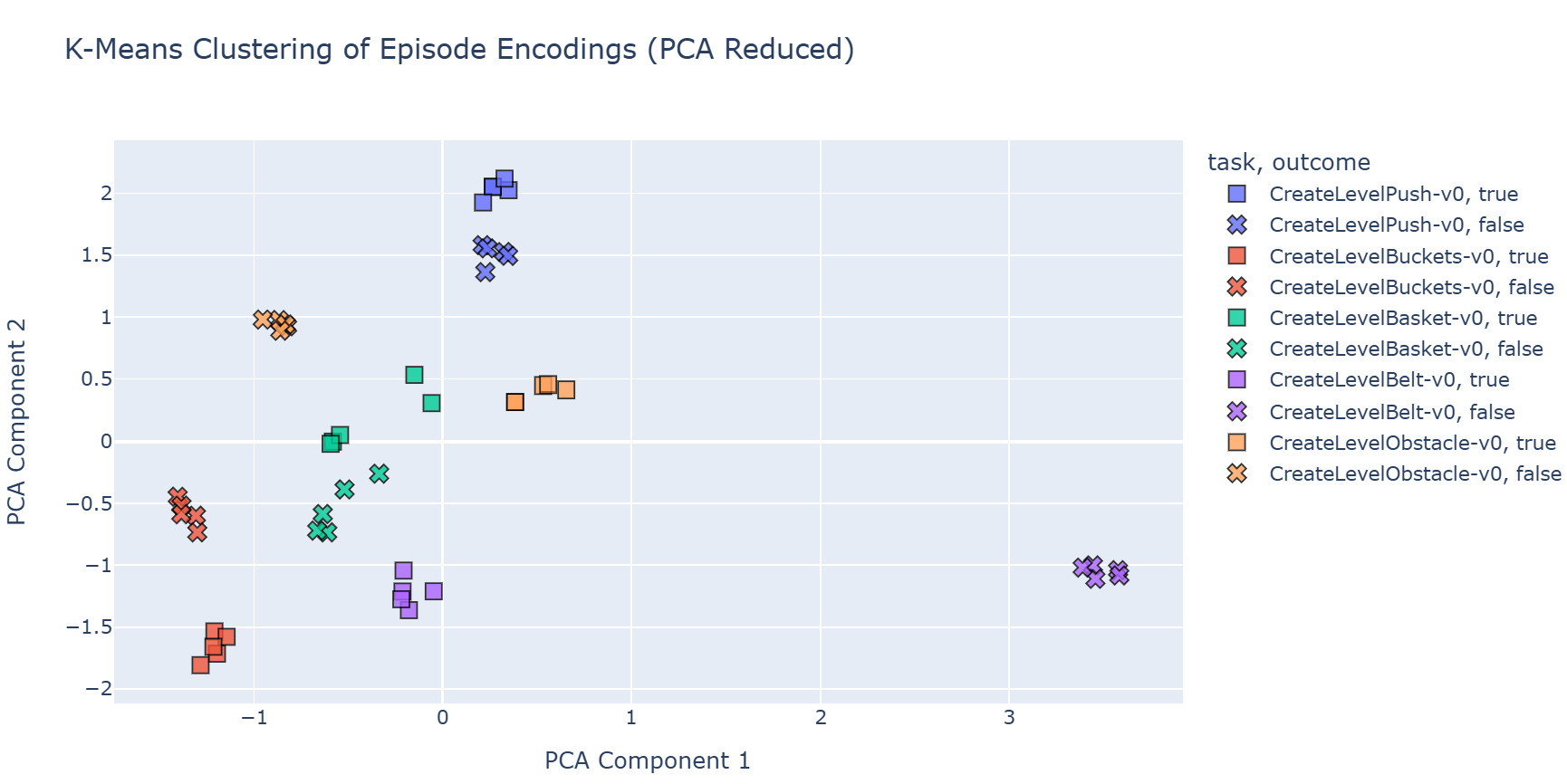}
        \caption{Outcome and task-specific clusters: Each point represents a SET in the CREATE environment, colour-coded by task, where circles mark successful SETs and crosses mark failed ones.}
        \label{fig:cluster_res2}
    \end{subfigure}
    \caption{K-means clustering of SET embeddings. (a) Clustering by task only. (b) Clustering by task and outcome (success vs. failure). SETLE effectively captures both task-level distinctions and outcome-specific information.}
    \label{fig:task_outcome_clusters}
\end{figure}

Figure \ref{fig:cluster_res} shows the K-means clustering of SET embeddings, colour-coded by task. The results highlight SETLE’s ability to capture task-level distinctions and result-specific information.

Additionally, the separation of successful and unsuccessful SETs within the same task, indicated by distinct markers in the plot, further highlighted SETLE’s effectiveness in encoding task-specific information. For example, episodes from CreateLevelPush that successfully reached the goal state were grouped closely, while failed attempts were separated, providing evidence that SETLE captures not only the task identity but also the quality of task execution Figure \ref{fig:cluster_res2}.

\subsubsection{Loss Functions Impact and Results}

To further analyse the impact of different optimisation strategies on SET representation, we compared the performance of two loss functions: the simple triplet loss and the hybrid loss. While the triplet loss focusses on enforcing distance constraints between similar and dissimilar SET embeddings, the hybrid loss integrates contrastive learning by aligning embeddings across schema and meta-path views. This combination aims to enhance both structural consistency and task separability.

\begin{figure}[h!]
    \centering
    \begin{subfigure}[b]{1\textwidth}
        \includegraphics[width=\textwidth]{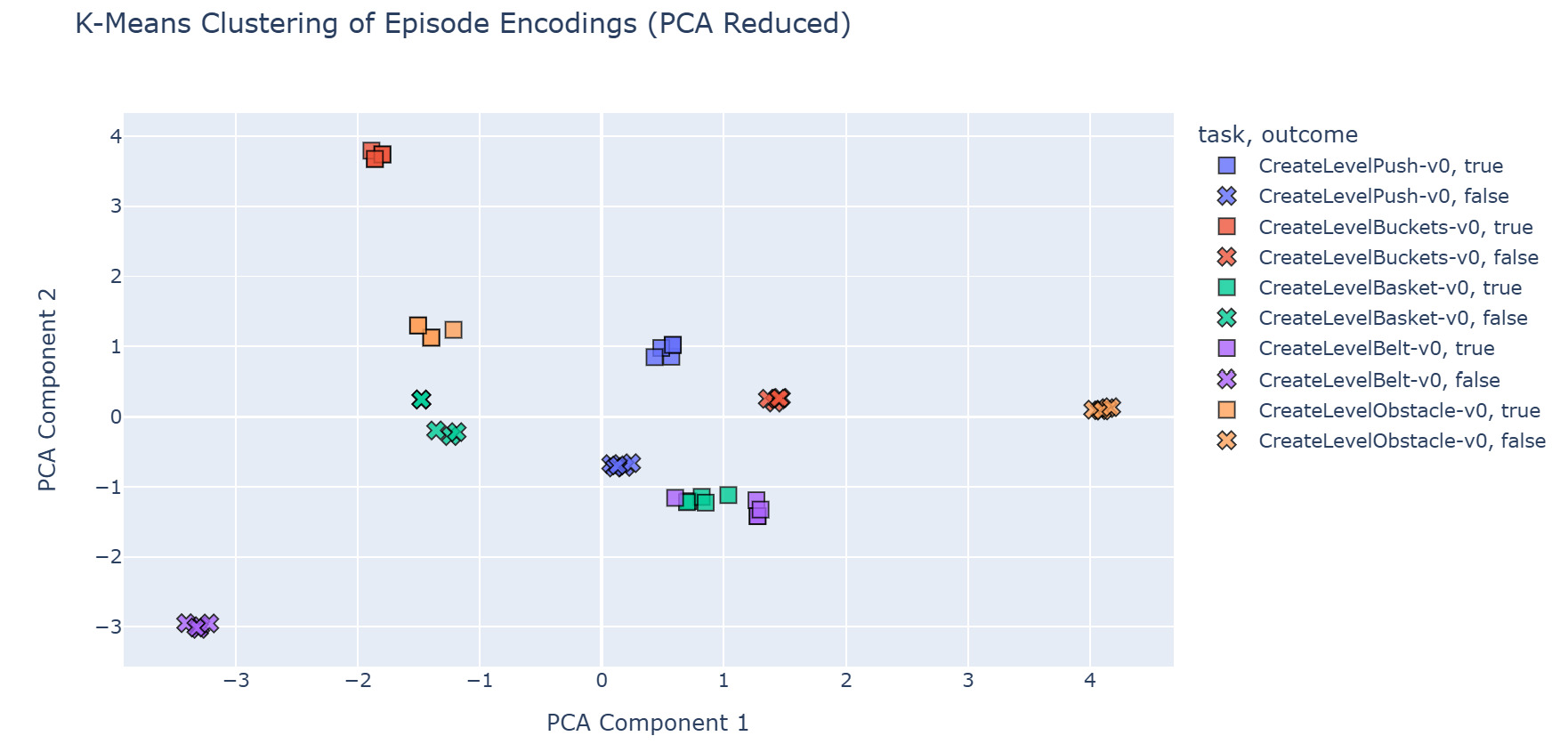}
        \caption{Triplet Loss trained encoder clustering results.}
        \label{fig:triplet_loss}
    \end{subfigure}
    \hfill
    \begin{subfigure}[b]{1\textwidth}
        \includegraphics[width=\textwidth]{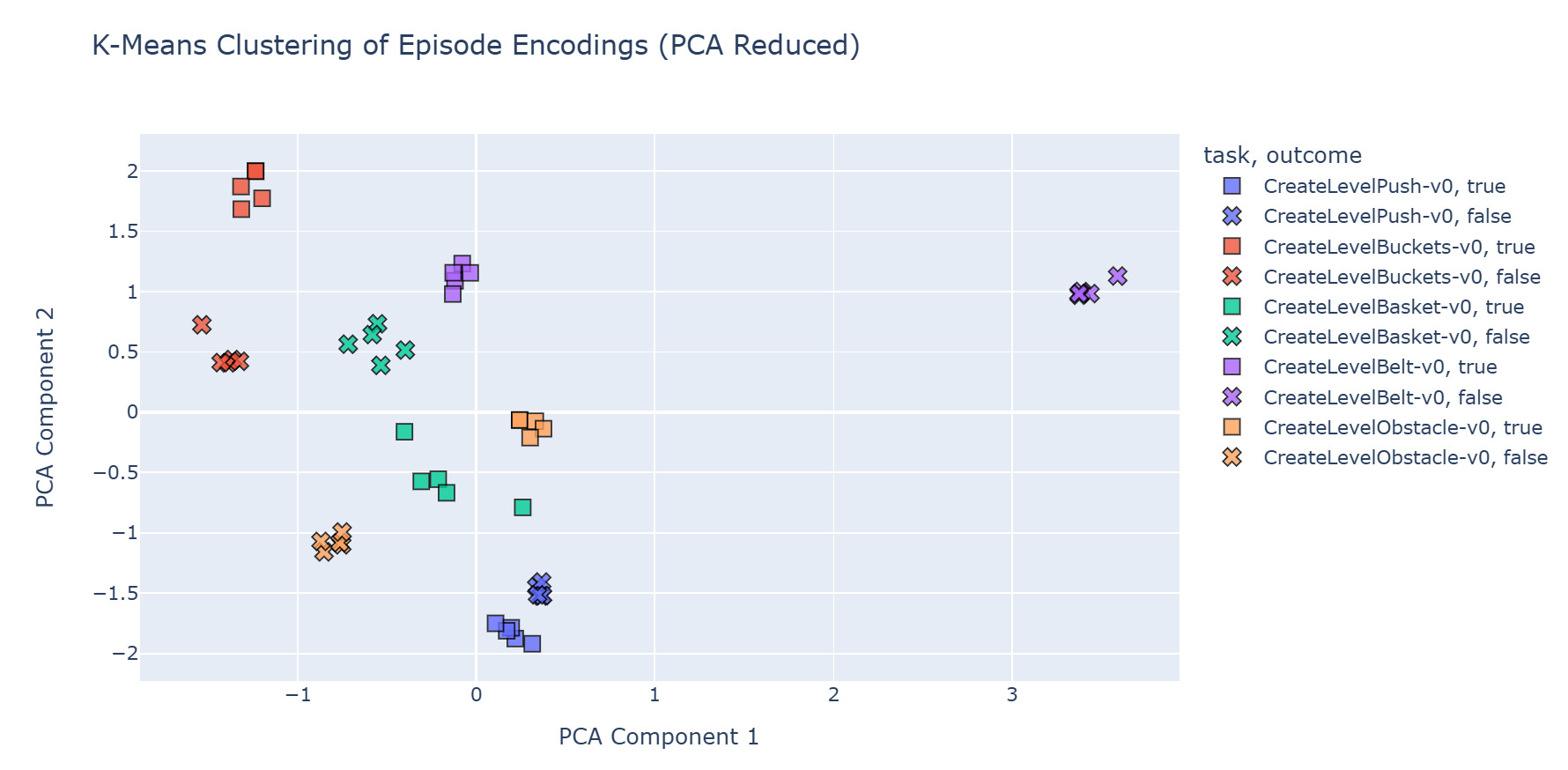}
        \caption{Hybrid Loss trained encoder clustering results.}
        \label{fig:hybrid_loss}
    \end{subfigure}
    \caption{
        Comparison of K-means clustering of SET embeddings under different loss functions. 
        (a) Triplet loss: Provides reasonable task separation but limited inter-task alignment. 
        (b) Hybrid loss: Enhances task separation and aligns result-specific embeddings more effectively by leveraging cross-view contrastive learning. Successful SETs (\(\circ\)) are closer within the same task, while failed ones (\(\times\)) remain distinct, reflecting improved generalization and result differentiation.
    }
    \label{fig:loss_comparison}
\end{figure}

To assess the effectiveness of the hybrid loss in improving representation compared to the standard triplet loss, we visualised the SET embeddings generated by SETLE under both loss functions. Each point represents a SET, colour-coded by task, with successful episodes marked as \(\square\) and unsuccessful episodes marked as \( \times \). The visualisations are provided in Figures \ref{fig:triplet_loss} and \ref{fig:hybrid_loss}, where subfigure (a) shows the results with triplet loss and subfigure (b) shows the results with hybrid loss.

\paragraph{Observations on Inter-task Improvements:}
The hybrid loss demonstrated inter-task improvements compared to the triplet loss:
\begin{itemize}
    \item \textbf{Task Separation:} The hybrid loss resulted in better separation between tasks compared to the triplet loss. In Figure \ref{fig:hybrid_loss}, tasks with shared structural dynamics (e.g., tasks using similar objects like buckets or baskets) were positioned closer in the embedding space, while maintaining distinct clusters for unrelated tasks. The hybrid loss leveraged cross-view contrastive signals (e.g., schema and meta-path views) to capture shared dynamics between tasks. This led to embeddings that better encode structural similarities and inter-task relationships, facilitating generalisation.
    
  \item \textbf{Outcome-specific Encoding:} While both loss functions distinguished successful and unsuccessful SETs within the same task, the hybrid loss brought positive and negative outcomes from the same task closer together compared to the triplet loss. This behaviour arose from the cross-view contrastive learning mechanism in the hybrid loss, which emphasised shared structural and task-specific features across both positive and negative outcomes. By leveraging complementary views (e.g., schema and meta-path), the hybrid loss encouraged the embeddings of episodes from the same task, regardless of their outcome, to cluster more closely while maintaining sufficient separation between success and failure states. This reflected a more nuanced encoding of the shared task dynamics and structural similarities inherent in both successful and unsuccessful trajectories.

\end{itemize}

The hybrid loss considerably enhanced the representation of inter-task relationships while maintaining strong intra-task and outcome-level differentiation, by using cross-view contrastive learning in addition to triplet-based alignment.

\paragraph{Comparison with State-of-the-Art Heterogeneous Graph Encoders} 

While state-of-the-art heterogeneous graph encoders such as HeCo and GTC excel in static graph settings and node-level tasks like classification and clustering, they are not directly comparable to SETLE due to differences in problem setting, data structure, and application focus. SETLE addresses the unique challenge of encoding episodic, dynamic graphs, where each episode forms a structured subgraph (SET) representing trajectories. Unlike the static, node-centric graphs typically used by HeCo and GTC, the data in SETLE comes from episodic environments like CREATE, where the graphs evolve dynamically over time and must encapsulate hierarchical abstractions across objects, interactions, states, affordances, and trajectories. 

 This fundamental difference in problem formulation and the nature of the data makes direct comparisons with existing heterogeneous graph encoders unsuitable and underscores the distinct contribution of SETLE to learning hierarchical SET representations.

 \subsection{Comparison on CREATE environment}

In \cite{jain2020generalization}, action representations were evaluated by encoding agent actions across episodes and analysing their separability within the embedding space. Their results highlighted effective clustering of actions, particularly for distinguishing tasks requiring distinct agent behaviours.

In contrast, SETLE shifted the focus to trajectory representations, encoding not only the actions but also the intermediate states, transitions, and affordances that cumulatively define the task’s success or failure. The comparison reveals that action representations, as emphasised by \cite{jain2020generalization}, excel in precision for short-term predictions, such as determining the next best action in a sequence. SETLE, however, takes a broader approach by also learning interaction representations using a ConvLSTM architecture and then integrating them with state, affordance, and object information to enable comprehensive SET representation learning. By combining detailed interaction-level encoding with higher-level abstractions of task trajectories, SETLE provides a unified framework that supports both immediate decision-making and generalisation across tasks. This integration highlights the complementary nature of interaction and trajectory representations, with SETLE excelling in task-level analysis while maintaining robust interaction representation capabilities.

\subsubsection{S2: Ablation Studies Emphasising the Model's Design}

To further investigate the importance of SETLE’s hierarchical structure, we conducted three additional experiments: random state removal, sequentiality disruption, and flattened representations. These experiments highlighted the robustness of the model and its reliance on structured representations for accurate outcome embeddings.

1. Random State Removal

\textbf{Objective:} Demonstrate the importance of sequential states in accurately representing SETs.

\textbf{Design:} Randomly removed a fixed proportion (e.g., 10\%, 20\%, 40\%) of the states in the graph. Encode the modified SET using SETLE and compare clustering quality. Below we presented the results of removing 40\% of the states, having observed that for lower values the results were not statistically reliable.

\textbf{Remarks:} State removal disrupted the sequential flow and relational dependencies, leading to poorer cluster cohesion and separability (see Table \ref{table:state_removal}).

\textbf{Results:}

\begin{table*}[ht!]
\centering
\caption{Comparison of Clustering Results for Random State Removal vs. Complete Graphs}
\label{table:state_removal}
\begin{tabular}{p{2.0cm} p{2.2cm} p{1.8cm} p{6.0cm}} 
\toprule
\textbf{Metric} & \textbf{State Removal} & \textbf{Complete} & \textbf{Remarks} \\
\midrule
Silhouette Success & 0.756 & 0.797 & Reduced cohesion and separation of success clusters. \\
Silhouette Failure & 0.827 & 0.820 & Less cohesive clustering. \\
DBI Success & 0.472 & 0.345 & Higher overlap and less distinct success clusters. \\ 
DBI Failure & 0.298 & 0.314 & Marginally better cluster definition. \\ 
Dunn Index Success & 1.838 & 2.561 & Reduced separation and compactness of success clusters. \\ 
Dunn Index Failure & 2.904 & 2.618 & Less compact and separated failure clusters. \\ 
\bottomrule
\end{tabular}
\end{table*}

2. Sequentiality Disruption

\textbf{Objective:} Test how breaking the sequential order of states impacts SET representation.

\textbf{Design:} Randomly shuffled the sequence of states in each episode while keeping all nodes intact.

\textbf{Remarks:} Sequential disruption reduced the ability to encode cumulative effects, leading to embeddings that are less meaningful and cohesive.

\textbf{Results:}
\begin{itemize}
    \item \textbf{Silhouette Scores:} Shuffled sequences exhibit lower clustering quality (success: 0.785 compared to 0.748).
    \item \textbf{Dunn Index:} Noticeable degradation in success cluster compactness (2.285 compared to 1.613).
    \item \textbf{DBI:} Increased overlap between shuffled task clusters (success: 0.378 compared to 0.494).
\end{itemize}

3. Flattened Representations

\textbf{Objective:} Tested the importance of hierarchical elements in trajectories by replacing the hierarchical graph with a flat graph structure.

\textbf{Remarks:} Flattening the graph drastically degraded the quality of the learnt task representations, highlighting the importance of relational data for meaningful task dynamics (see Table \ref{table:flattening_set}).

\begin{table*}[ht!]
\centering
\caption{Impact of Flattening SET Hierarchical Structure on Clustering Results}
\label{table:flattening_set}
\begin{tabular}{p{1.8cm} p{1.8cm} p{1.8cm} p{6.5cm}} 
\toprule
\textbf{Metric} & \textbf{Flattened} & \textbf{Complete} & \textbf{Impact of Flattening} \\
\midrule
Silhouette Success & 0.705 & 0.770 & Decreased cohesion and separation of success clusters. \\
Silhouette Failure & 0.782 & 0.823 & Reduced distinct boundaries for failure clusters. \\
DBI Success & 0.595 & 0.473 & Increased overlap and less distinct success clusters. \\
DBI Failure & 0.425 & 0.305 & Higher overlap between failure clusters. \\
Dunn Index Success & 1.326 & 1.652 & Reduced separation and compactness of success clusters. \\
Dunn Index Failure & 1.864 & 2.888 & Poorer separation and compactness of failure clusters. \\
\bottomrule
\end{tabular}
\end{table*}

{
\subsection{Performance Evaluation in Reduced-Complexity Environments: MiniGrid}

While the CREATE environment \citep{jain2020generalization} showcased SETLE’s ability to learn from rich, dynamic, and continuous visual environments, a central hypothesis of our work is that the same underlying mechanism can generalise across varying representational regimes—including low-resolution, symbolic environments. To test this, we extended SETLE to the MiniGrid benchmark suite, a grid-based environment that offers highly abstract, discrete observations and minimal visual complexity.

Unlike CREATE, MiniGrid lacks continuous interactions or a persistent inventory system and features a limited, discrete action space. Accordingly, the graph construction process was modified: "Action Representation" nodes, previously used to encode rich agent-tool-object dynamics, were omitted. The resulting MiniGrid-specific SETs contain only Object, State, and Affordance nodes, linked via relational edges that trace the agent’s symbolic trajectory.

This adaptation serves to isolate and evaluate the contribution of SETLE’s graph-based inductive bias in a minimal setting. 

To mitigate the limitations of low-resolution symbolic input in MiniGrid, further refinements were applied to the object extraction pipeline. Rendered frames were first pre-processed using contrast equalisation and spatial upscaling to enhance visual saliency. Each segmented crop was then passed through CLIP \citep{ramesh2022hierarchical} to obtain a robust vision-language embedding. CLIP was chosen explicitly over conventional encoders such as ResNet \citep{he2016deep} or VGG \citep{simonyan2014very} because these standard architectures, trained on natural image statistics, catastrophically fail when applied to synthetic grid-based environments with flat colours and uniform textures. CLIP, by contrast, demonstrated the ability to align symbolic patterns with semantic concepts even in low-visual-complexity domains. This design choice was critical for preserving the object-centric nature of the SETLE framework and maintaining consistency between the original CREATE experiments and the MiniGrid adaptation.

Once the static objects were extracted and stored, SETLE followed its standard graph memory construction: State nodes were created at each timestep, Affordance nodes were linked to represent the agent’s chosen action at each transition, and the object nodes were consistently connected to all subsequent states in the episode.

We trained the SETLE encoder on MiniGrid episodes using the same procedure as described for CREATE. Figure~\ref{fig:minigrid_setle_loss} illustrates the impact of varying the triplet loss margin $\alpha$ in the hybrid loss function during training. As in CREATE, all models converge to similarly low loss values, with lower margins (e.g., 0.2 and 0.5) exhibiting slightly smoother convergence. However, convergence stability alone does not capture the quality of the learned embeddings. To better assess the impact of margin settings, we evaluated clustering performance using the same standard metrics: Silhouette Score, DBI and Dunn Index, computed separately for success and failure clusters. For example, with $\alpha = 0.2$, we observed a strong Silhouette Score of 0.69 for successful episodes and a low DBI of 0.45, indicating compact and well-separated clusters. By contrast, $\alpha = 1.2$ achieved a slightly lower silhouette (0.57) and a higher DBI (0.78), suggesting less distinct cluster boundaries. In contrast to CREATE, where higher margin values (e.g., 1.2 and 1.5) were needed to enforce separation between similar-looking episodes due to visual simplicity and frame-level redundancy, MiniGrid presented a reverse pattern. Despite its symbolic and low-resolution observations, MiniGrid's discrete and abstract representation space contains more structured and semantically distinct transitions. As a result, lower margin values (e.g., 0.2 and 0.5) were sufficient to stabilise training and encourage meaningful embedding separation. 

\begin{figure}[!ht]
    \centering
    \includegraphics[width=1\linewidth]{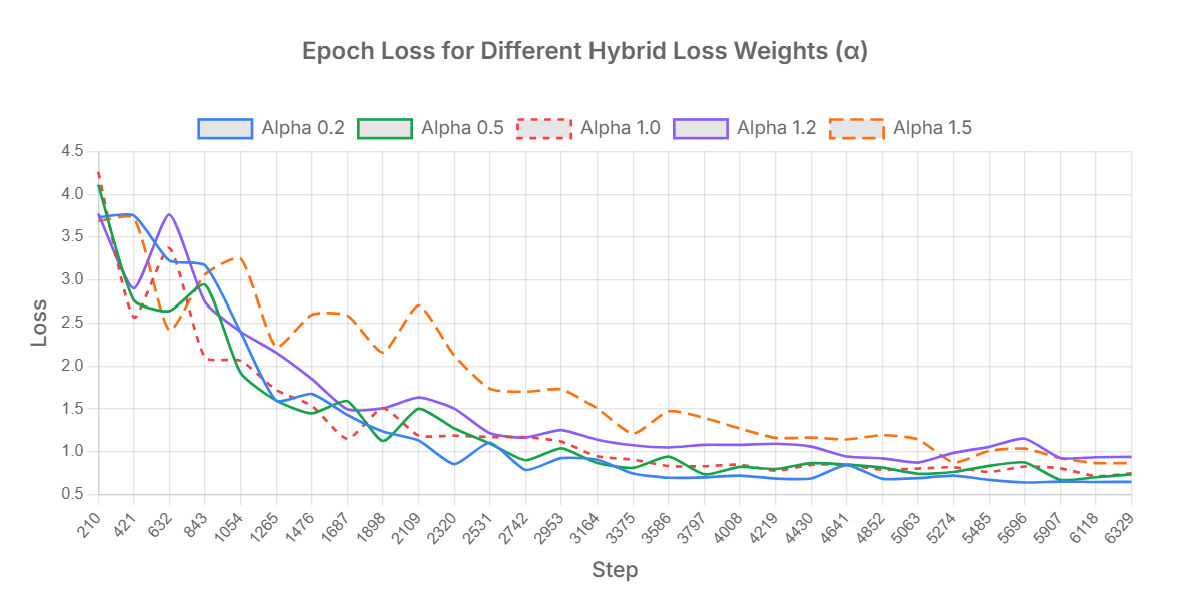}
    \caption[Training loss curves for hybrid loss weights]{Epoch-wise training loss curves for different values of the hybrid loss weight $\alpha$ during training with alpha cumulation.}
    \label{fig:minigrid_setle_loss}
\end{figure}

 We now turn to visual clustering analyses to further examine this effect. Using PCA-reduced embeddings and K-Means clustering, we visualise how different tasks and outcomes are organised in the latent space. These findings provide insights into the agent's ability to separate task semantics, represent outcome distinctions, and recognise structural task similarity.

 The results reveal several important trends. First, as shown in Fig. \ref{fig:minidiff}, tasks with structurally distinct layouts—such as "MultiRoom-N4-S5-v0", "UnlockPickup-v0" and "SimpleCrossing-S9N1", form clearly separated clusters, suggesting that the encoder successfully captures high-level differences in spatial configuration and affordance structure. When both successful and unsuccessful episodes are considered (Fig. \ref{fig:minidiffout}), we observe a consistent separation between outcomes, even within the same task. This indicates that SETLE encodes not only the task setting but also the behavioural trajectory and its result, capturing outcome-sensitive variations in agent interaction.

\begin{figure}[!ht]
    \centering
    \includegraphics[width=1\linewidth]{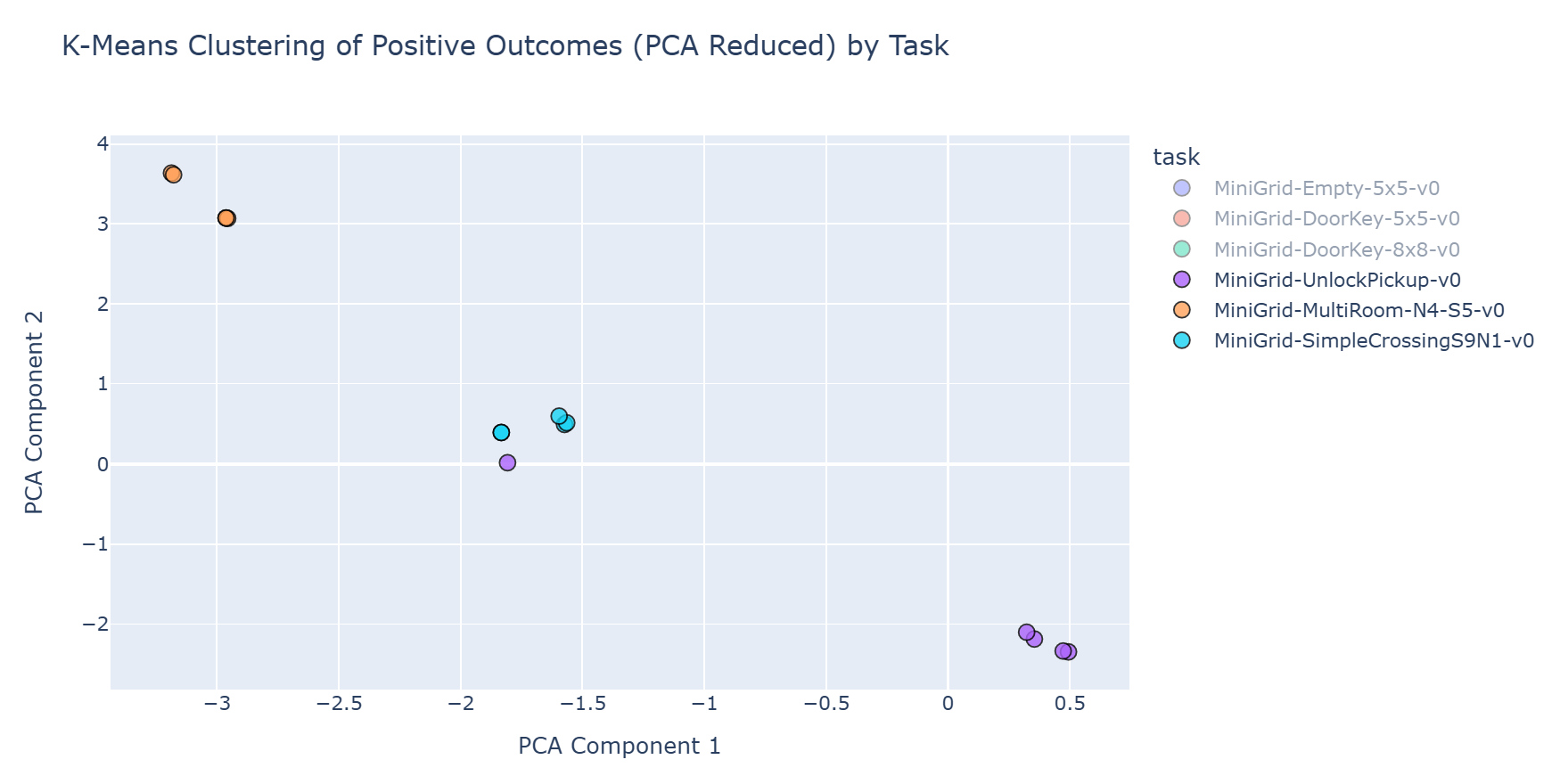}
    \caption[PCA of episode embeddings across tasks]{PCA-reduced embeddings of successful episodes from multiple MiniGrid tasks, clustered using K-Means. Tasks with structurally distinct configurations—such as MultiRoom-N4-S5-v0, SimpleCrossing-S9N1 and UnlockPickup-v0, form clearly separated clusters, indicating that the encoder captures meaningful variations in spatial layout and task structure.

}
    \label{fig:minidiff}
\end{figure}

\begin{figure}[!ht]
    \centering
    \includegraphics[width=1\linewidth]{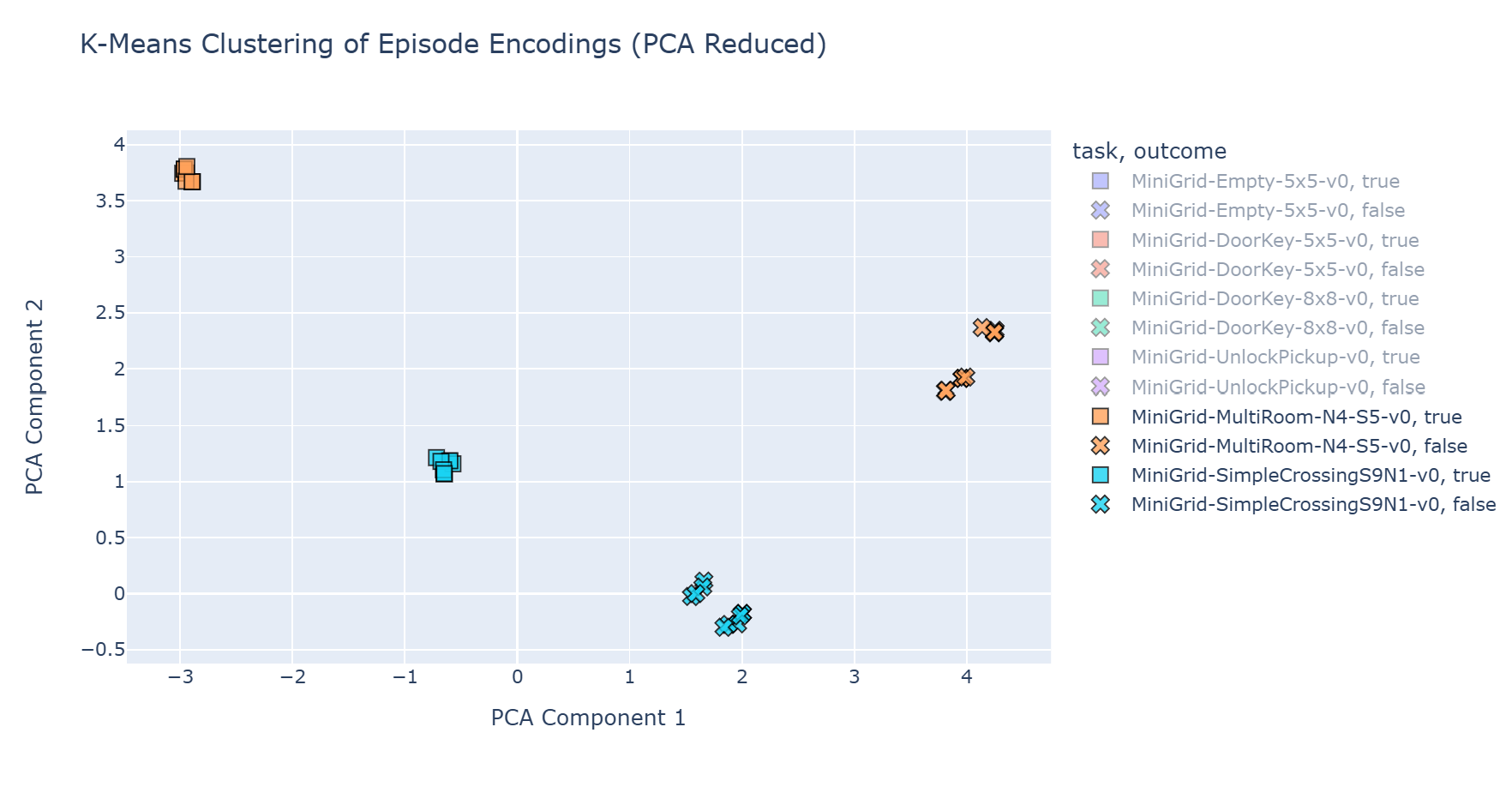}
    \caption[Clustering of episodes in complex Minigrid tasks]{K-Means clustering of PCA-reduced episode embeddings across tasks: MultiRoom-N4-S5-v0 and SimpleCrossing-S9N1, including both successful and unsuccessful outcomes. While task identity still influences clustering, clear separation between positive and negative outcomes emerges, highlighting the encoder’s sensitivity to behavioural results and trajectory success.}
    \label{fig:minidiffout}
\end{figure}

Interestingly, this separation is not universal: in simpler tasks such as "Empty-5x5-v0" and "DoorKey-5x5-v0", the embeddings overlap substantially, despite the fact that the latter includes a goal-directed key-door interaction. This overlap is further confirmed in Fig. \ref{fig:minisim}, which focuses exclusively on positive outcomes from these two tasks. Notably, "Empty-5x5-v0" can be interpreted as a simplified form of "DoorKey-5x5-v0", where the spatial layout and goal location remain constant, but the constraint introduced by the door is removed. The encoder appears to capture this equivalence, treating both tasks as functionally similar due to their shared goal structure and minimal planning horizon. Crucially, this similarity persists even though the encoder was trained to maximise discriminability between episodes, indicating that SETLE’s representations are not just outcome-aligned but also sensitive to deeper notions of task equivalence. 

\begin{figure}[!ht]
    \centering
    \includegraphics[width=1\linewidth]{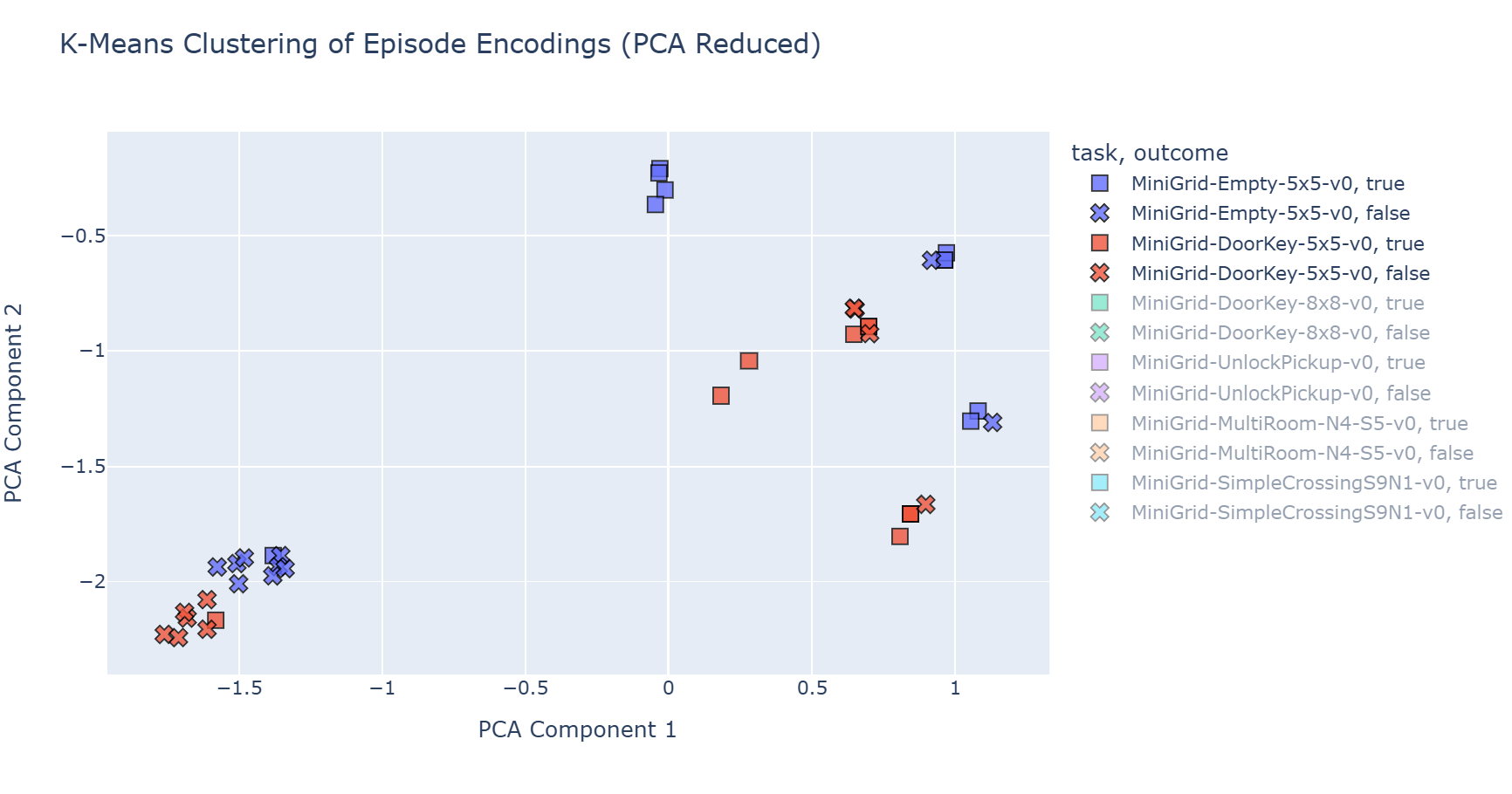}
    \caption[Clustering of episodes in Empty vs DoorKey tasks]{ K-Means clustering of PCA-reduced episode encodings for "Empty-5x5-v0" and "DoorKey-5x5-v0", including both successful and unsuccessful outcomes. While outcome-specific clusters are clearly distinguishable within each task, the two tasks themselves show substantial representational overlap—indicating that the encoder treats them as structurally similar despite the presence of an additional key-door interaction in the latter.}
    \label{fig:minisim}
\end{figure}

This alignment between simple tasks becomes even more apparent when extending the comparison to larger configurations. In particular, "DoorKey-8x8-v0", a more spatially demanding version of "DoorKey-5x5-v0", remains tightly clustered alongside the simpler "Empty-5x5-v0" and "DoorKey-5x5-v0" tasks when considering only positive outcomes (Fig. \ref{fig:minibigger}). This reinforces the hypothesis that SETLE encodes task representations not purely based on grid size or visual complexity, but rather on the underlying goal structure and interaction demands. Despite being trained to distinguish episodes through a contrastive objective, the encoder learns to cluster trajectories that share similar causal and functional properties. At the same time, outcome-specific variation is still well represented: as shown in Fig. \ref{fig:miniwithin}, positive and negative trajectories from the same task are consistently separated. This highlights SETLE's ability to encode both high-level structural similarity and fine-grained behavioural distinctions, making it well-suited for trajectory-level reasoning and generalisation.

\begin{figure}[h]
    \centering
    \includegraphics[width=1\linewidth]{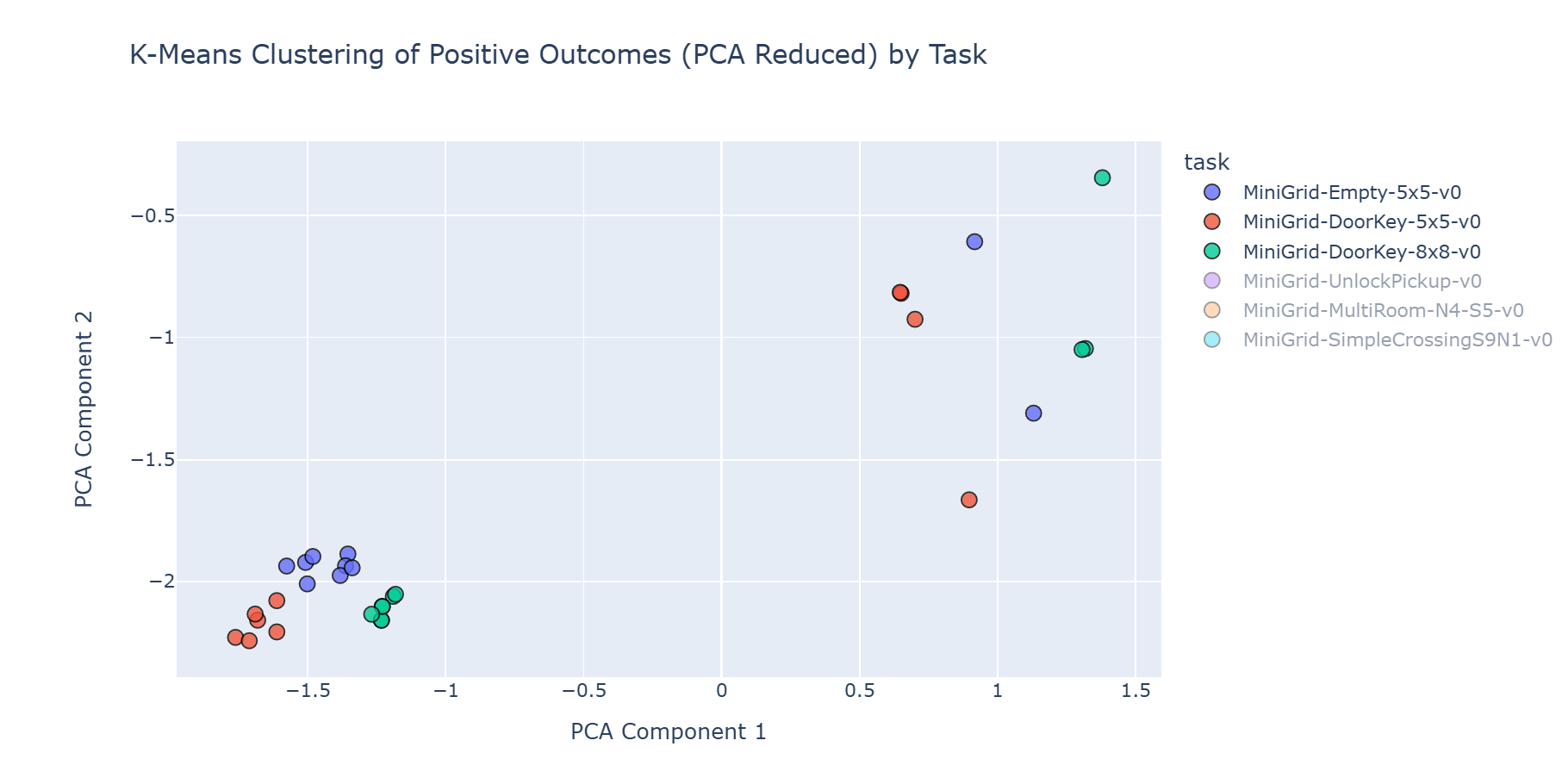}
    \caption[Cross-task clustering of successful outcomes]{ Clustering of positive outcomes across "Empty-5x5-v0", "DoorKey-5x5-v0", and "DoorKey-8x8-v0", visualised using PCA. The tight grouping across different grid sizes and levels of visual complexity suggests that SETLE encodes similarity based on functional structure and goal dynamics, rather than superficial spatial variation.}
    \label{fig:minibigger}
\end{figure}

\begin{figure}[h]
    \centering
    \includegraphics[width=1\linewidth]{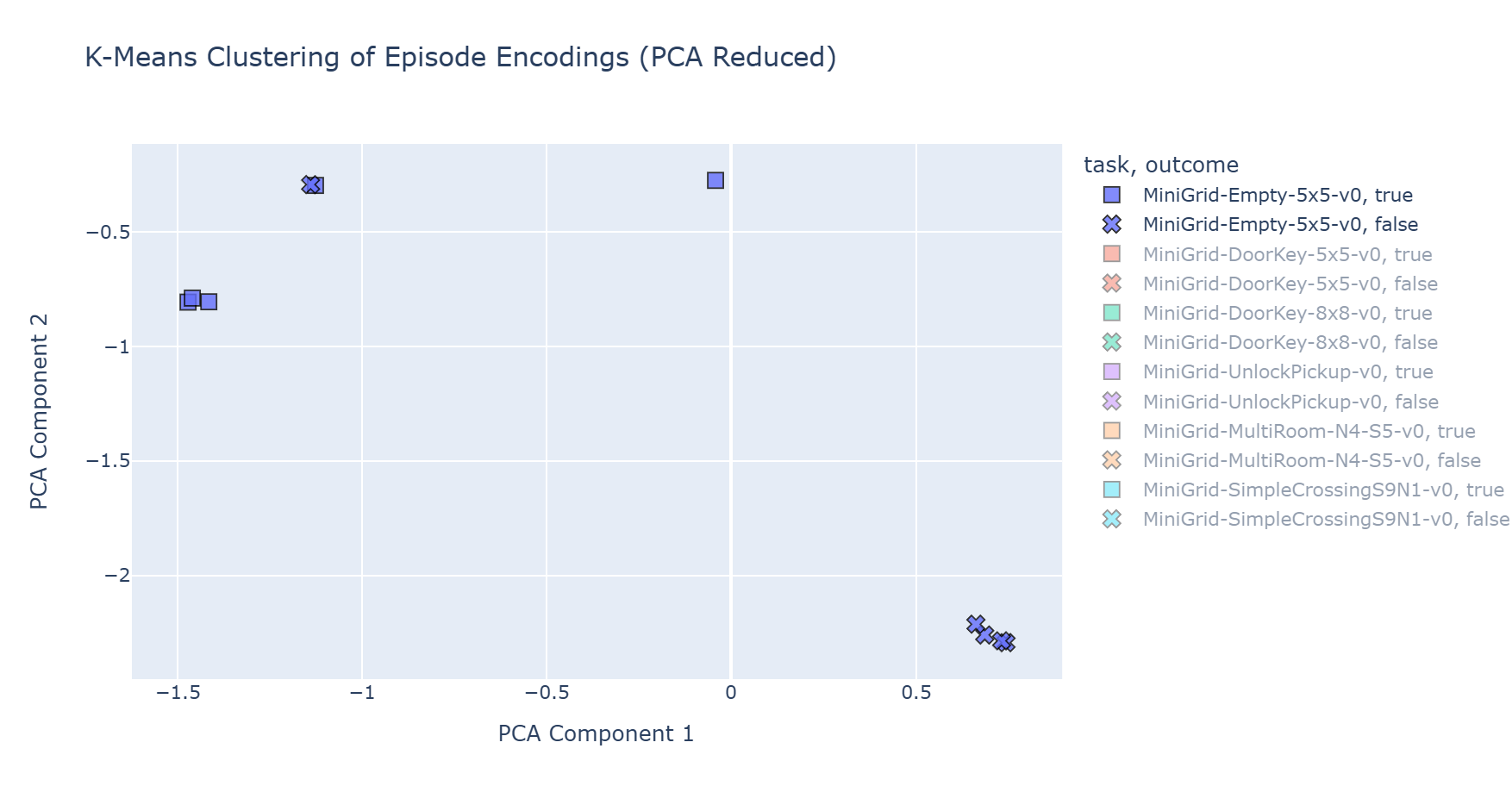}
    \caption[Episode clustering within Empty-5x5 task]{K-Means clustering of full episodes for "Empty-5x5-v0" reveals strong separation between successful and unsuccessful trajectories, even within a single task. This demonstrates that SETLE captures fine-grained behavioural differences and can distinguish between functionally meaningful outcomes.
}
    \label{fig:miniwithin}
\end{figure}

The clustering results in MiniGrid confirm that SETLE effectively captures both structural task similarity and outcome-based distinctions across a variety of environments. The model generalises across tasks with shared goals and low interaction complexity—such as Empty and DoorKey—while clearly separating those with more complex layouts or affordance demands. 

}

{
\subsection{Statistical Analysis and Error Reporting}
To rigorously evaluate our results, we performed a statistical analysis on the final epoch losses from 8 successful training runs. The data did not significantly deviate from a normal distribution, as confirmed by a Shapiro-Wilk test (N=8, p = 0.156), justifying the use of parametric tests. We grouped the runs by their margin value ($\alpha$) into three methodologically sound, discrete categories: Low ($\alpha \le 0.5$), Medium ($0.5 \le \alpha \le 1.0$), and High ($\alpha \ge 1.0$).

A one-way ANOVA was conducted to compare the effect of margin group on performance (N=8). The omnibus test did not reach conventional statistical significance (F(2, 5) = 2.34, p = 0.178). This p-value corresponds to approximately 82\% evidence against the null hypothesis, indicating moderate but sub-threshold support for group differences. Given the limited sample size (8 runs across 3 groups), this analysis is likely underpowered. A large partial $\eta^2$ (0.48) was observed and is reported descriptively; however, no inferential conclusions are drawn from effect size in the absence of statistical significance. These results are therefore exploratory and motivate further experiments with increased repetitions and formal power analysis.


To investigate this trend further, we conducted a planned comparison between the extreme groups. An independent samples t-test confirmed that the 'Low' margin group (N=3) produced a statistically significant lower mean loss (M=0.71, SD=0.05) compared to the 'High' margin group (N=3, M=0.85, SD=0.55), t(5) = -2.83, p = 0.037. This result provides strong evidence that lower margin values lead to more stable and effective training for the encoder in our experiments.
}
{
\section{Integrating SETLE into the Reinforcement Learning Loop}

To move from representation to application, this section details the integration of the SETLE framework into a reinforcement learning (RL) training loop. While earlier sections described the design and function of SETLE for encoding trajectories, this section demonstrates how these structured representations are used to guide an agent's learning process.

We describe how episodic graphs are encoded and enriched with relevant past knowledge via SETLE, and how this enriched state supports generalisation and adaptation during action selection. We also show how memory is updated incrementally, enabling the agent to form increasingly abstract conceptual structures over time. Specifically, we first describe the baseline reinforcement learning architecture for training in both CREATE and MiniGrid environments in subsection 5.1, and how it is integrated in SETLE in subsection 5.2. After specifying the training regimes and experimental designs we report results in subsection 5.3. This section thus completes the pipeline, from perceptual abstraction, through memory-guided reasoning, to adaptive behaviour within a functioning reinforcement learning agent.

\subsection{Overview of the Baseline RL Architecture}
To contextualise the performance of SETLE-enriched agents, we begin by outlining the baseline reinforcement learning (RL) setup used in our experiments. This architecture serves as the foundation upon which all enrichment mechanisms are evaluated.

The baseline agent is based on a Double Deep Q-Network (Double DQN) \citep{van2016deep}, selected for its stability in Q-value estimation and compatibility with mixed observation and action modalities. The agent uses two Q-heads: one for discrete action selection and one for continuous control, depending on the environment.

The baseline agent observes the raw state (for both environments), encodes it using a convolutional or linear encoder, and selects actions via an epsilon-greedy policy. The Q-network is updated using experience replay and periodic target network updates.

While simple, Double DQN provides a strong and interpretable baseline. Its lack of structured memory, graph-based reasoning, or conceptual abstraction makes it an ideal comparison point for assessing the contribution of the SETLE framework.

We evaluate across two distinct domains:

\textbf{CREATE Environment}: A visually rich, physics-based environment designed for object manipulation and tool use. Tasks include pushing balls into containers, navigating obstacles, and achieving complex multi-step goals. CREATE is particularly suited for testing structured planning, affordance learning, and behavioural reuse.

\textbf{MiniGrid Suite}: A minimalist 2D gridworld composed of symbolic entities (e.g., keys, doors, goals) and agent-centric observations.

\subsubsection{Training Procedure (CREATE Baseline)}

The baseline agent for the CREATE environment is trained using a Deep Q-Network (DQN)-style architecture adapted for a \textbf{hybrid action space}. At each timestep, the agent selects a tool from its inventory (discrete action) and predicts a spatial coordinate $(x, y)$ (continuous action) at which to apply the selected tool. Formally, each action is represented as:
\[
a_t = \langle o_i, (x, y) \rangle
\]
where $o_i$ is a tool from the current inventory, and $(x, y)$ denotes the interaction location in the environment.

To handle this mixed action structure, we adopt a \textbf{multi-headed Q-network} that outputs:
\begin{enumerate}
    \item A categorical Q-value vector over all possible tools (up to 939), masked by the current inventory.
    \item A continuous 2D vector representing the predicted coordinates $(x, y)$ for tool application.
\end{enumerate}
This follows the \textit{multi-headed architecture} approach to hybrid action modelling, allowing independent training of discrete and continuous components while enabling joint decision-making.

\paragraph{Action Selection Strategy.}
The agent follows a $\epsilon$-greedy policy. With probability $\epsilon$, it explores by randomly sampling a tool from the inventory and a random valid position. Otherwise, it exploits the learned Q-function by selecting the tool with the highest Q-value (subject to masking) and uses the network’s predicted position.

The exploration rate decays exponentially over time:
\begin{equation}
\epsilon_t = \epsilon_{\text{end}} + (\epsilon_{\text{start}} - \epsilon_{\text{end}}) \cdot e^{-t / \text{decay\_rate}}
\end{equation}
with parameters: $\epsilon_{\text{start}} = 1.0$, $\quad \epsilon_{\text{end}} = 0.1$, $\quad \text{decay\_rate} = 10^4$
\vspace{-1em}
\paragraph{Learning}
Both Q-heads are trained jointly using a combination of Temporal-Difference (TD) loss for the discrete tool selection and regression loss for coordinate prediction.

The discrete Q-head is initialised using \textit{Xavier uniform weights}, and positive bias initialisation is applied to encourage early exploration. All training metrics—Q-values, rewards, loss breakdown, and $\epsilon$ decay—are logged using Weights \& Biases (W\&B) and will be discussed in the Results subsection. The complete training process for the baseline agent is outlined in Algorithm 1.

\begin{algorithm}[H]
\caption{Hybrid Action DQN Training Loop}
\begin{algorithmic}[1]
\State Initialize Q-network $Q_\theta$ with:
\State \hspace{1em}1. Discrete head for tool selection
\State \hspace{1em}2. Continuous head for coordinate prediction
\State Initialize target network $Q_{\theta^-}$, experience buffer $\mathcal{D}$
\State Set $\epsilon \leftarrow \epsilon_{\text{start}}$
\For{each episode}
    \State Reset environment and tool inventory
    \For{each timestep $t$}
        \If{random $< \epsilon$}
            \State Sample tool $\hat{a}$ uniformly from inventory
            \State Sample $(\hat{x}, \hat{y})$ uniformly from valid space
        \Else
            \State $\hat{a} \leftarrow \arg\max Q_\theta(s_t, a)$ (masked by inventory)
            \State $(\hat{x}, \hat{y}) \leftarrow Q^{\text{coord}}_\theta(s_t, \hat{a})$
        \EndIf
        \State Execute action $(\hat{a}, \hat{x}, \hat{y})$, observe $r_t$, $s_{t+1}$
        \State Store $(s_t, a_t, r_t, s_{t+1})$ in buffer $\mathcal{D}$
        \State Sample mini-batch from $\mathcal{D}$ and update $Q_\theta$
        \State Update target network periodically: $Q_{\theta^-} \leftarrow Q_\theta$
        \State Decay $\epsilon$ using exponential schedule
    \EndFor
\EndFor
\end{algorithmic}
\end{algorithm}

\subsubsection{Training Procedure (MiniGrid Baseline)}

In contrast to the CREATE environment, where agents interact through a hybrid action space and inventory-based tool application, the MiniGrid setup features a \textbf{purely discrete action space}. The agent must learn to navigate, interact with objects, and solve tasks such as key-door unlocking and room traversal using a fixed set of primitive discrete actions: \texttt{move forward}, \texttt{turn left}, \texttt{turn right}, \texttt{pickup}, \texttt{drop}, and \texttt{toggle}.

To preserve consistency with the object-centric framing of the CREATE experiments while avoiding reliance on symbolic state input, the baseline MiniGrid agent receives only \textbf{pixel-level visual observations} via \texttt{env.render()}. It does not have access to internal simulator state or structured symbolic observations such as \texttt{obs["image"]}. This design choice aligns the setup with real-world perception constraints, where agents must extract actionable information from raw sensory input. Given MiniGrid’s discrete and minimalistic action space, a number of structural and computational adaptations were made to SETLE. Unlike in CREATE, where Action Representation nodes captured rich object-tool interactions, the MiniGrid graph structure was simplified. It included only State and Affordance nodes, connected by \texttt{influences} and \texttt{outcome} edges. This simplification allowed direct testing of the claim that the benefits of structured episodic data structures do not stem from predefined semantics but from the flexible object-centric trajectory modelling enabled by SETLE.

To reduce computational overhead, the visual perception module was adapted. The Segment Anything Model (SAM) and CLIP were used only once at the beginning of each episode, leveraging the fact that MiniGrid environments are static within each episode. Extracted objects (e.g., walls, keys, doors) remain unchanged during interaction, and are thus linked to every State node throughout the episode graph without needing re-inference. This drastically reduced GPU usage and inference time, while still ensuring that the episodic graph maintained its conceptual integrity.

For the reinforcement learning architecture, a standard Double DQN agent is used. The Q-network consists of a convolutional encoder followed by a fully connected Q-head predicting discrete action values. Exploration follows the same $\epsilon$-greedy schedule as described in Section 5.1.1.

The agent is trained using standard temporal-difference loss over minibatches sampled from an experience replay buffer, and the target network is periodically updated. No memory augmentation or enrichment is used in this baseline.

Training is conducted across six standard MiniGrid tasks:
\texttt{Empty 5x5}, \texttt{DoorKey 5x5}, \texttt{DoorKey 8x8}, \texttt{UnlockPickup}, \texttt{MultiRoom N4 S5}, and \texttt{SimpleCrossing S9N1}. These tasks collectively test basic navigation, symbolic object use, and multi-room spatial reasoning.

This baseline serves as the foundation for evaluating whether structured memory and enrichment mechanisms, introduced in subsequent sections, yield improvements in sample efficiency, stability, and generalisation across task variants.

\subsection{SETLE Integration in the Reinforcement Learning Loop}

To evaluate the impact of structured episodic memory on generalisation and learning efficiency, we integrate SETLE into the reinforcement learning (RL) training loop (see Algorithm \ref{alg:setle_match_enrich}). This allows the agent to supplement its current experience by matching recent trajectories against long-term memory (LTM), retrieving relevant structures, and enriching its working memory (WM) before selecting an action.

\begin{algorithm}[H]
\caption{SETLE Matching and Enrichment at Timestep $t$}
\label{alg:setle_match_enrich}
\begin{algorithmic}[1]
\Require Partial trajectory $G_{t-3:t}$, current $state\_id$, $episode\_id$, SETLE encoder, LTM
\Ensure Enriched working memory graph at time $t$
\State Encode partial trajectory: $z_{\text{query}} \leftarrow \texttt{encode\_episode}(G_{t-3:t})$
\State Retrieve LTM episode encodings $\{(z_i, \text{episode}_i)\}_{i=1}^N$
\State Compute similarity: $s_i = \cos(z_{\text{query}}, z_i)$
\State Select top-$K$ most similar episodes: $\texttt{top\_episodes} \leftarrow \texttt{argsort}(s)[:K]$
\For{each matched $\texttt{episode}_j$}
    \State Extract candidate nodes (objects, affordances, actions) not in $G_t$
    \State Apply relevance and attention filtering
\State Select top-$N$ nodes by attention score
\State Inject selected nodes and edges into WM graph at $state\_id$ with \texttt{source=setle}
\EndFor

\end{algorithmic}
\end{algorithm}

\paragraph{Functional Overview.}
Unlike standard RL agents that rely exclusively on immediate state observations, SETLE allows the agent to draw from structurally similar prior experiences. Rather than reusing entire episodes, the agent selectively retrieves and injects semantically aligned substructures, such as objects, affordances, and actions, into the current WM graph. This supplemented graph is then re-encoded and passed to the policy network, forming a cycle of perception, memory-guided reasoning, and decision-making.

To support temporal abstraction and contextual matching, SETLE encodes a rolling window of the last $k=4$ state graphs as a partial trajectory, denoted $G_{t-3:t}$. This representation captures both spatial and temporal dependencies, helping disambiguate states that may appear similar in isolation but differ in their broader interaction history. These richer representations improve the precision and utility of memory retrieval.

The core mechanism of SETLE follows a selective matching and supplement pattern. The agent compares the current partial trajectory $G_{t-3:t}$ to stored episodes in LTM, full encoded trajectories, and identifies structurally similar ones. From these, it extracts missing but relevant elements and injects them into the current WM graph. This interaction is formalised in Algorithm~\ref{alg:setle_match_enrich}.

To identify what knowledge is most relevant, SETLE employs a trainable attention mechanism that scores candidate nodes based on their contextual alignment with the current partial trajectory. This targeted prioritisation enables fine-grained control over memory reuse, ensuring that only structurally and semantically pertinent components are integrated into the working memory graph.

The selected nodes and edges form a \textit{supplement} to the current graph $G_t$, producing an enriched graph $G^*_t$ that includes potentially useful but previously absent information. This enriched graph is re-encoded using the SETLE GNN encoder and passed through an adapter layer before being input to the policy network.

The integration of SETLE modifies the standard RL loop as follows:
\begin{enumerate}
    \item Encode the last four states as a partial trajectory graph.
    \item Retrieve the top-$k$ most similar episodic embeddings from LTM.
    \item Expand the episodes to be able to select the needed concepts.
    \item Filter retrieved graph concepts by context and apply attention to rank and select the most relevant candidates.
    \item Inject selected nodes and edges to supplement the current WM graph.
    \item Re-encode the enriched graph and pass it to the policy head.
\end{enumerate}

\subsubsection{Attention-Based Retrieval and Enrichment}
To retrieve relevant knowledge from memory, SETLE employs an attention mechanism that scores candidate nodes from past episodes based on their contextual alignment with the agent's current partial trajectory. This mechanism applies attention over node embeddings using a vector projection architecture inspired by the Transformer model. Given an encoded partial trajectory (query) and candidate node embeddings (keys) from retrieved episodes, we compute a relevance score via:
\begin{equation}
\alpha_i = \frac{\exp \left( \langle W_q \cdot \text{query},\; W_k \cdot \text{candidate}_i \rangle / \tau \right)}{\sum_j \exp \left( \langle W_q \cdot \text{query},\; W_k \cdot \text{candidate}_j \rangle / \tau \right)}
\end{equation}

This enables the model to learn a flexible, task-specific notion of relevance, allowing for the selective injection of nodes (e.g., objects, affordances) into the agent's working memory. As formalized in Algorithm \ref{alg:setle_enrichment}, this process operates entirely at inference time and supports generalization by aligning latent concept representations across diverse task structures.

\begin{algorithm}[H]
\caption{SETLE Matching and Enrichment Loop at Timestep $t$}
\label{alg:setle_enrichment}
\begin{algorithmic}[1]
\Require Partial trajectory $G_{t-3:t}$, LTM, current working memory graph $G_t$
\Ensure Enriched working memory graph $G_t^*$

\State $z_{query} \leftarrow \texttt{encode\_episode}(G_{t-3:t})$
\State Retrieve top-K most similar episodes $\{(z_j, \text{episode}_j)\}$ from LTM using $z_{query}$
\State Initialize `candidates` list
\For{each matched $\texttt{episode}_j$}
    \State Extract nodes from $\texttt{episode}_j$ that are not in $G_t$ and add to `candidates`
\EndFor
\State Compute attention weights $\alpha$ for all `candidates` using the query $z_{query}$
\State Select top-N candidates based on attention scores
\State Inject selected nodes and their edges into $G_t$ to create enriched graph $G_t^*$
\State \textbf{return} $G_t^*$
\end{algorithmic}
\end{algorithm}

This attention-based retrieval enables adaptive memory reuse.

\subsection{Training Regimes and Experimental Design}

To evaluate the impact of SETLE-based memory enrichment on reinforcement learning performance, we design a controlled experimental framework that compares multiple agent configurations across diverse tasks. The goal is to assess whether structured memory reuse improves learning efficiency, generalisation, and policy stability.

\paragraph{Experimental Conditions}
We evaluate five distinct training configurations, each representing a different integration strategy for SETLE into the RL loop:

\begin{enumerate}
    \item \textbf{Baseline RL:} A standard Double DQN agent trained without any graph memory or enrichment. This serves as the control condition.
    
    \item \textbf{Action Selection Only:} SETLE enrichment is applied exclusively at inference time. The policy network receives graph-enhanced embeddings for action selection, but gradients do not propagate through the memory pathway.
    
    \item \textbf{SETLE for Action and Optimisation:} Graph-enriched representations are used for both policy inference and backpropagation. The encoder is updated end-to-end, allowing memory reuse to influence learning directly.
    
    \item \textbf{Adapter + Penalty:} A learnable adapter transforms SETLE embeddings before they are fed into the Q-network. Additionally, a penalty is applied when the same LTM episodes are repeatedly matched, encouraging more diverse memory access.
    
    \item \textbf{Adapter + Penalty + Soft Update:} This final configuration adds a target encoder updated via soft updates (Polyak averaging) to increase temporal stability and reduce volatility during enrichment.
\end{enumerate}

Each configuration is tested on five CREATE tasks and a suite of MiniGrid scenarios. Experiments are conducted across three independent runs with different PyTorch random seeds to ensure statistical robustness. All runs are logged using Weights \& Biases (W\&B), capturing reward trajectories, Q-value trends, enrichment frequency, and memory match statistics.

Randomness was handled using the default pseudo-random number generators provided by each library. Python’s built-in random module and NumPy’s global RNG rely on Mersenne Twister (MT19937), while PyTorch uses a Philox-based generator for CUDA operations. PyTorch seeds were explicitly set for each run, while Python and NumPy relied on their default generators. Each training run was executed independently, and no seed hashing or reuse was applied. We acknowledge that the number of repetitions is limited and that future work will incorporate explicit seeding across all libraries and increased repetitions to enable more rigorous statistical analysis.

Performance is assessed using a combination of behavioural and representational metrics:
\begin{itemize}
    \item \textbf{Success Rate:} Percentage of episodes in which the goal is successfully achieved.
    \item \textbf{Reward Frequency:} Average number of positive reward events per episode.
    \item \textbf{Q-Value Stability:} Variance of predicted Q-values over time, used as a proxy for value estimation reliability.
    \item \textbf{Embedding Quality:} Silhouette scores and clustering metrics are used to evaluate how well SETLE embeddings separate tasks or structure trajectories.
\end{itemize}

\subsubsection{Results in Physically Grounded Tasks (CREATE)}

\paragraph{Baseline vs Action Selection only strategy}

To evaluate the impact of SETLE-based enrichment applied only at action selection time, we compare the reward trajectories of four independent training runs on the "Create Level Buckets" task. Figure~\ref{fig:reward_buckets_compare} overlays the baseline and enriched configurations.

Baseline agents exhibit slow, consistent learning, with reward increasing modestly over time. In contrast, SETLE-enhanced agents demonstrate higher initial rewards—particularly in one run (Run 3), which peaks above 1.8 early in training. However, this reward advantage tends to decay over time, suggesting that initial enrichment may bootstrap early performance but lacks sustained refinement. Baseline runs are tightly clustered, with minimal variation in performance. The action selection only condition introduces higher variance: while one run considerably outperforms the baseline, others perform comparably. This variance reflects the dependency of enrichment efficacy on the relevance and timing of matched episodes from long-term memory.

The comparison reveals that even lightweight enrichment strategies, such as injecting prior knowledge only during policy inference, can substantially accelerate early learning. However, these benefits are inconsistent unless coupled with mechanisms for continual memory refinement or adaptation.

\begin{figure}[H]
    \centering
    \includegraphics[width=\textwidth]{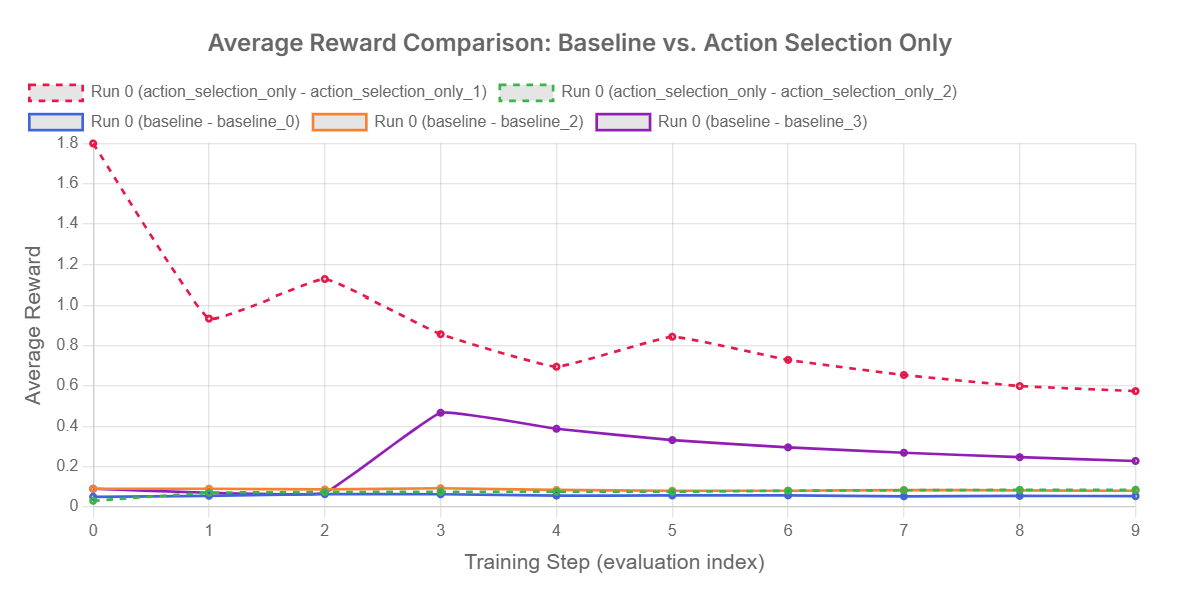}
    \caption[Reward curves for CreateLevel Buckets task]{Average reward over time on \texttt{Create Level Buckets} for baseline and SETLE-enhanced (action selection only) agents. Solid lines: baseline; dashed lines: enriched strategy.}
    \label{fig:reward_buckets_compare}
\end{figure}

\paragraph{Action selection only strategy vs Action Selection and Optimisation strategy}

The performance comparison of the inference vs inference and optimisation strategy Fig.\ref{fig:basket_optim_action} reveals a clear advantage of the full SETLE enrichment strategy that integrates both action selection and optimisation over the action-selection-only baseline. While both configurations benefit from retrieving relevant past experiences, applying enrichment only at inference time limits its long-term impact. In contrast, feeding enriched state representations into the training loss allows the agent to adjust its value function based on structurally informed information. This leads to more stable learning, faster convergence, and improved generalisation. The Basket task, which involves precise object placements and temporal dependencies, particularly benefits from this deeper integration. By using enrichment during both forward and backward passes, the agent reinforces not just immediate decisions, but also the internal representations that govern its learning dynamics.

\begin{figure}[ht]
    \centering
    \includegraphics[width=1\linewidth]{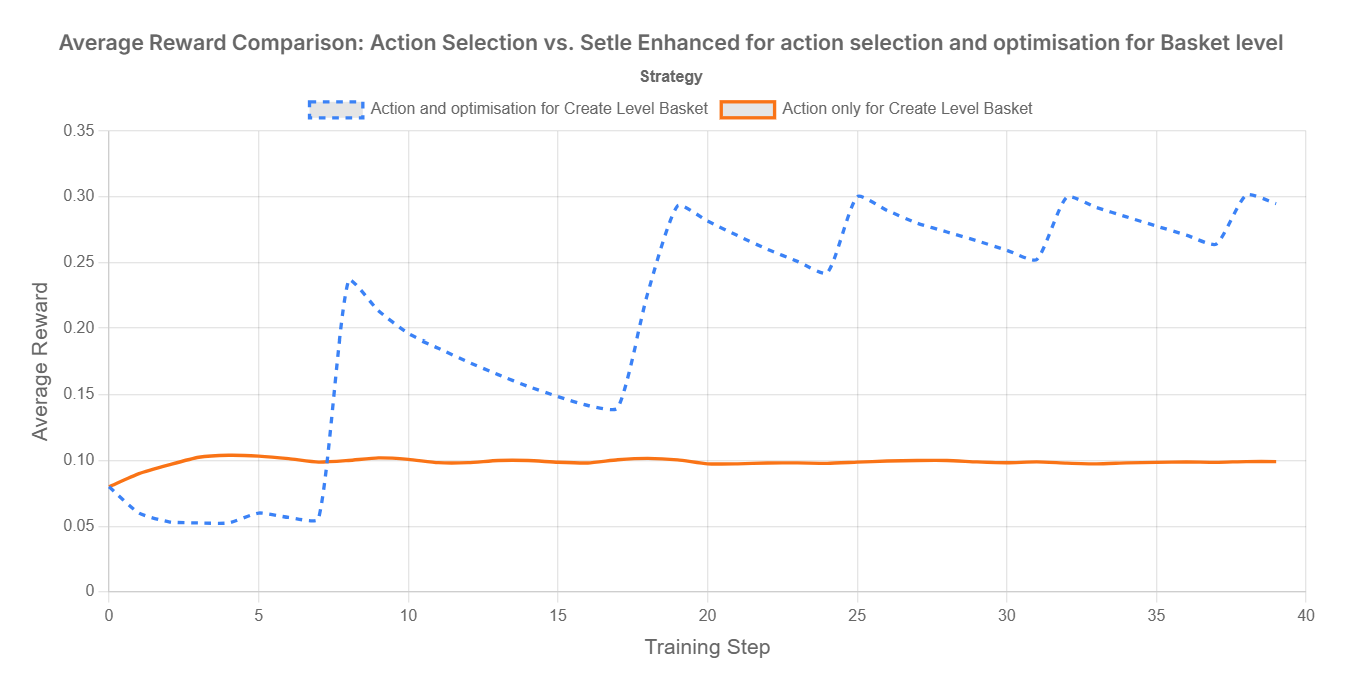}
    \caption[Reward comparison for SETLE optimisation strategies]{ Comparison of average reward over time for the CreateLevelBasket task using two SETLE configurations—Action Selection Only vs. Action + Optimisation.
The full optimisation strategy yields higher and more stable rewards, demonstrating the benefit of incorporating enriched memory representations during both inference and training}
    \label{fig:basket_optim_action}
\end{figure}
\vspace{-1em}
\paragraph{Further improvements to the RL loop: adapter and penalty}
To further stabilise and enhance the integration of memory into the reinforcement learning loop, we introduced two complementary mechanisms within the SETLE-enhanced agent: an adapter layer and a matching penalty. The adapter is a learnable projection module (implemented as a two-layer MLP) applied after graph-based enrichment. Its purpose is to map enriched state embeddings—composed of objects, affordances, and contextual relations retrieved from long-term memory—into a representation space better aligned with the Q-network's expectations. This addresses a critical issue in memory-based reinforcement learning: the potential distributional shift between learned embeddings and externally injected representations. Empirically, we found that the adapter substantially reduced the continuous loss (Fig. \ref{fig:setle_adapter_loss}), and also improved the Q-value stability (Fig. \ref{fig:setle_adapter_q}), indicating that the enriched information became more usable for the continuous action head.

\begin{figure}[H]
    \centering
    \includegraphics[width=1\linewidth]{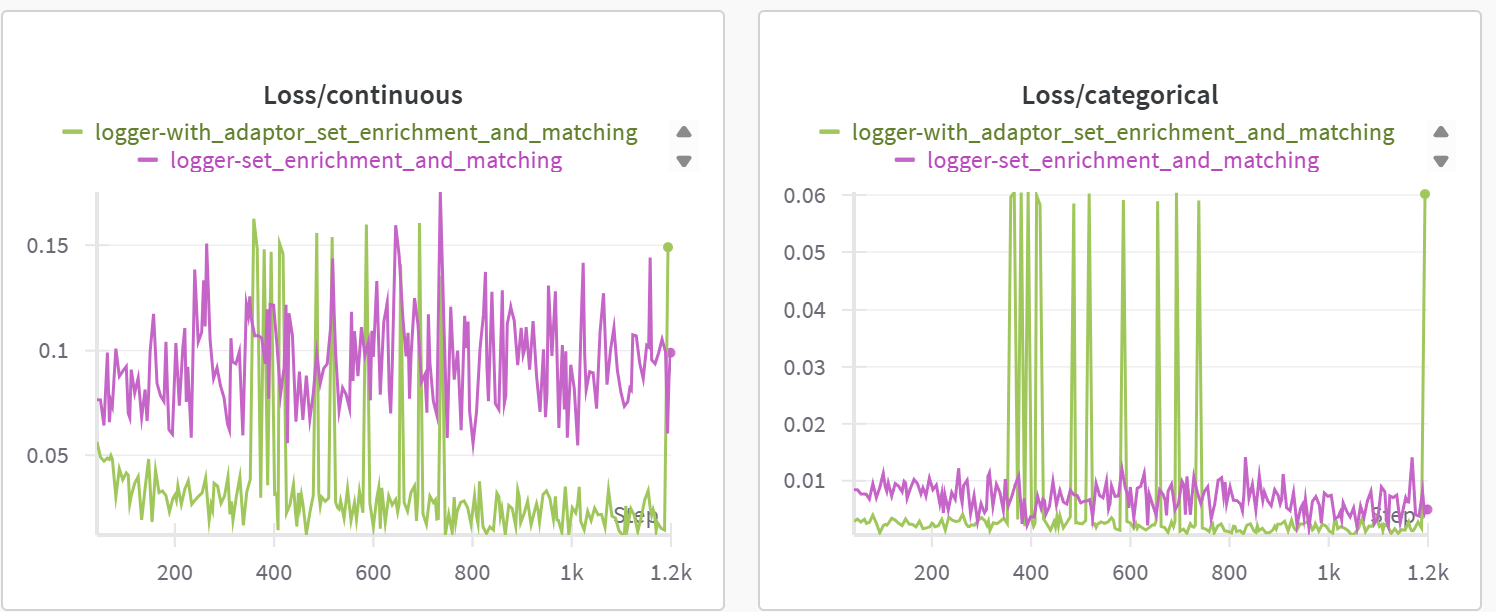}
    \caption[Training loss with and without adapter]{Comparison of training loss for the continuous (left) and categorical (right) heads under two strategies: standard SETLE enrichment and SETLE with an adapter. The adapter-based strategy achieves consistently lower loss in the continuous head and comparable or improved loss in the categorical head, indicating that the enriched embeddings are easier to optimise when mapped through a learnable projection layer.}
    \label{fig:setle_adapter_loss}
\end{figure}

\begin{figure}[H]
    \centering
    \includegraphics[width=1\linewidth]{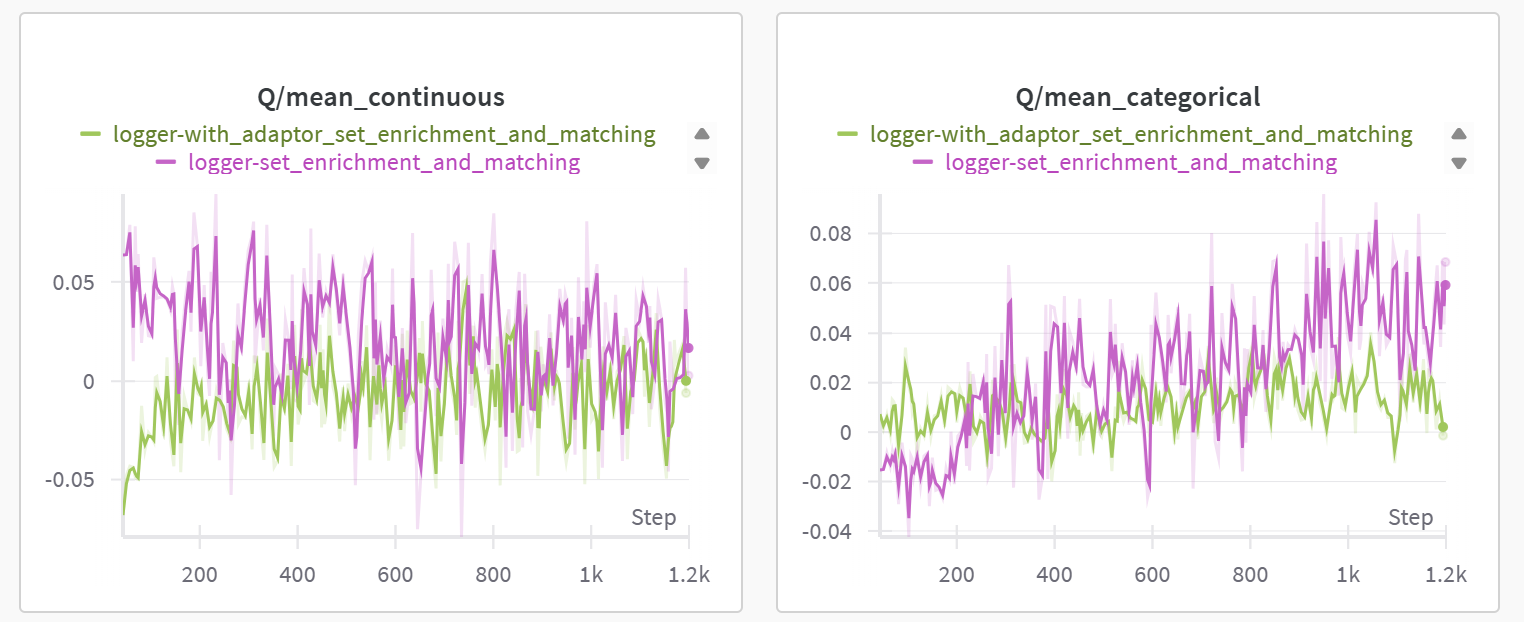}
    \caption[Q-value comparison with and without adapter layer]{Comparison of mean Q-values for the continuous (left) and categorical (right) action heads, with and without the adapter layer. The agent using the adapter (with\_adapter\_set\_enrichment\_and\_matching) exhibits smoother and more stable Q-value evolution in both heads. This suggests that the adapter helps reconcile the enriched embeddings with the policy network’s expectations, leading to more reliable value estimation.}
    \label{fig:setle_adapter_q}
\end{figure}
 The matching penalty, on the other hand, addresses a common failure mode in memory-based systems: repeated reliance on familiar or frequently matched trajectories. In earlier experiments, we observed that the agent would frequently retrieve the same high-scoring past episodes during enrichment, leading to repeated reinforcement of narrow behavioural patterns. To counteract this, we implemented a dynamic penalisation scheme that down-weights the similarity scores of frequently matched episodes. This forced the retrieval mechanism to consider a broader set of candidates, encouraging exploration and preventing overfitting to a small subset of trajectories. This penalty was particularly effective in later stages of training, where diversity in retrieved episodes was critical to maintain learning momentum and avoid premature convergence.

Together, these strategies aim to balance effective reuse of prior knowledge with the flexibility needed for robust adaptation across tasks.

To evaluate the downstream behavioural impact of the adapter and reuse penalty, we executed the full SETLE-enriched reinforcement learning loop across multiple tasks. Figure~\ref{fig:reward_push_adapter} focuses on the \texttt{Create Level Push} task, where agents must manipulate tools to move objects toward target zones, a task requiring spatial precision and multi-step affordance chaining.

Among the configurations tested, the adapter-based strategy consistently achieved higher average rewards over time, indicating improved sample efficiency and generalisation. This sustained performance suggests that the adapter module effectively aligns enriched graph representations with the policy network’s internal feature space, allowing the agent to better exploit retrieved knowledge without destabilising learning.

While all SETLE-augmented variants, including action selection only and action+optimisation, outperformed the early baseline, the adapter + penalty strategy was unique in maintaining reward gains after the initial peak. This was particularly notable given the challenge of delayed rewards in the Push task. The adapter appeared to dampen the mismatch between static memory embeddings and dynamic task features, while the penalty term discouraged overfitting to frequently matched episodes.

\begin{figure}[ht]
    \centering
    \includegraphics[width=1\linewidth]{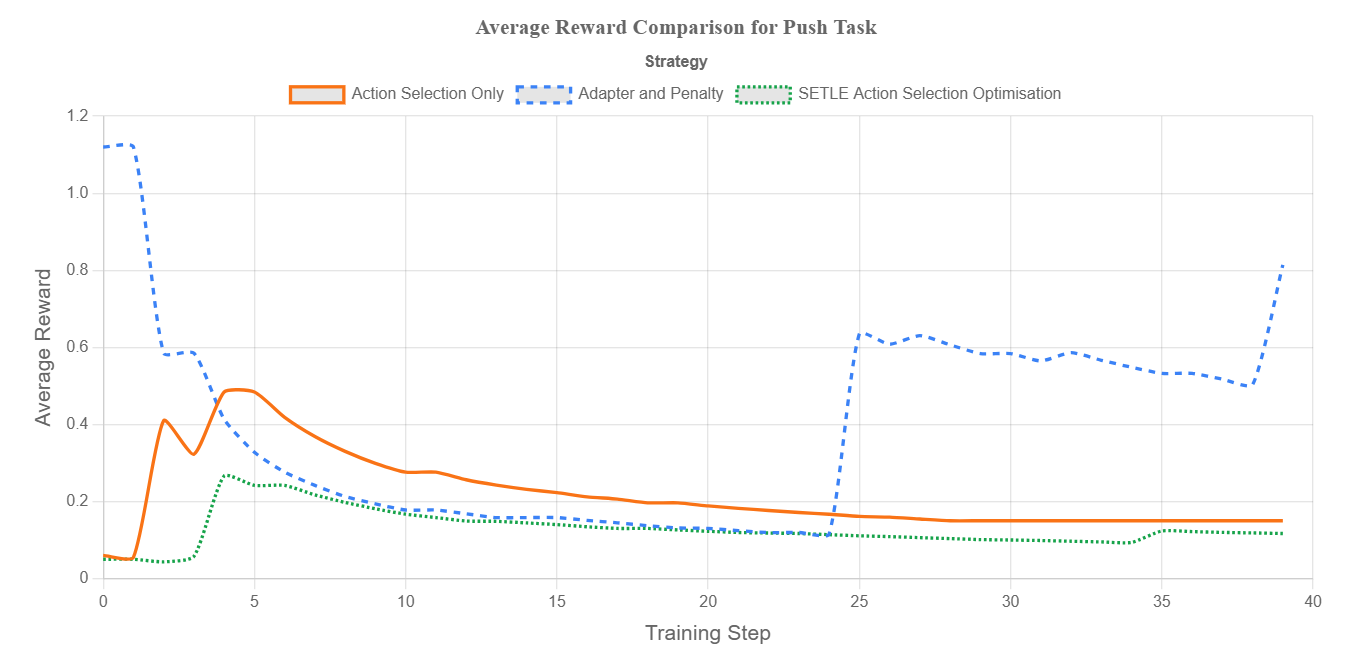}
    \caption[Average rewards across SETLE integration strategies]{ Average episode reward on CreateLevelPush across three SETLE integration strategies. The adapter-enhanced model (blue dashed line) consistently outperforms simpler strategies, demonstrating improved reward acquisition and training robustness.}
    \label{fig:reward_push_adapter}
\end{figure}

\paragraph{Further improvements to the RL loop: soft update}

While the adapter-enhanced SETLE model showed improved policy learning and reward acquisition, the Q-value and loss plots (Figures~\ref{fig:setle_adapter_q}, \ref{fig:setle_adapter_loss}) reveal training instabilities, particularly in the form of sharp spikes in both continuous and categorical loss. These fluctuations suggest that although enriched embeddings carry useful information, their integration introduces representational shifts that are difficult for the policy network to absorb consistently in a fully synchronous update scheme. To mitigate this, we introduce a soft target network update mechanism, which blends the parameters of the policy and target networks more gradually. By slowing the rate of change in Q-value targets, soft updates reduce oscillations and promote convergence when learning from enriched state representations that vary over time.

As shown in Figure~\ref{fig:spiky}, without soft updates, both continuous and categorical losses display frequent high-magnitude spikes. 
\begin{figure}[H]
    \centering
    \includegraphics[width=1\linewidth]{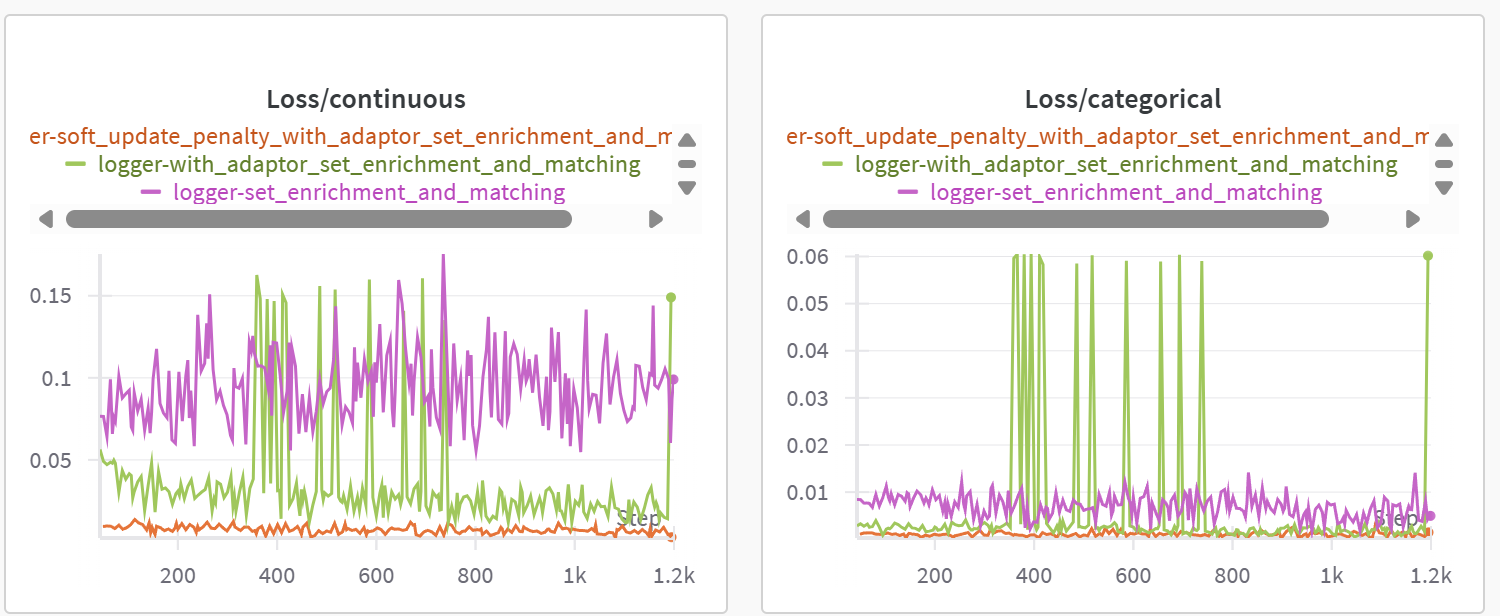}
    \caption[Loss curves with and without soft updates]{Loss curves for the adapter-enhanced SETLE agent with and without soft updates. Sharp spikes in both continuous and categorical loss indicate unstable value updates caused by abrupt changes in enriched state representations.}
    \label{fig:spiky}
\end{figure}
These indicate unstable value updates, especially when the enriched state representation changes abruptly between timesteps or episodes. By contrast, the introduction of the soft update mechanism leads to noticeably smoother training dynamics (Figure~\ref{fig:soft_loss}). 
All three loss components: total, continuous, and categorical, show reduced variance, lower peaks, and more consistent downward trends over time. Similarly, in the Q-value plots (Figure~\ref{fig:soft_q}), the mean Q-values remain more stable and within narrower ranges, especially for the continuous branch. This indicates that the soft update effectively damps abrupt value shifts caused by mismatch between new enriched information and the existing value function.

\begin{figure}[H]
    \centering
    \includegraphics[width=1\linewidth]{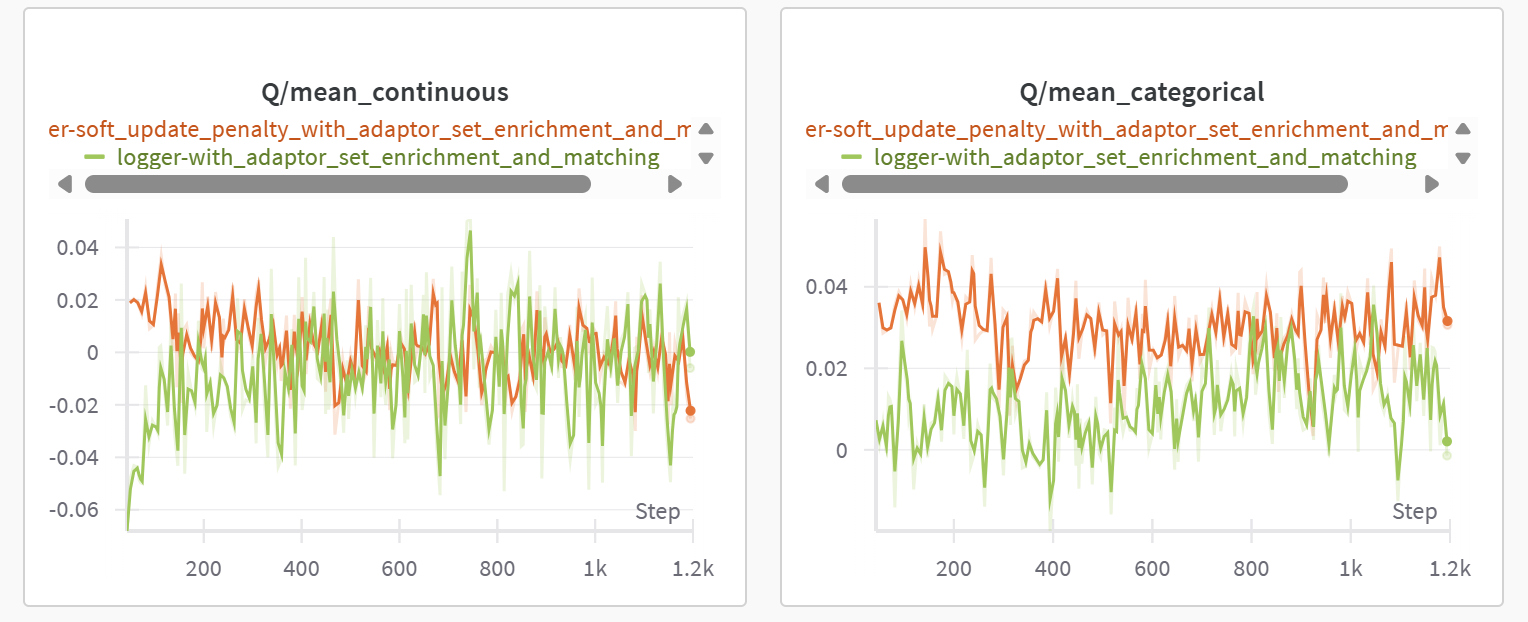}
    \caption[Q-value stability using soft update]{Q-value statistics for the soft update strategy. The mean Q-values remain bounded and relatively stable over time, indicating more consistent learning dynamics when enriched state information is gradually integrated.}
    \label{fig:soft_q}
\end{figure}

\begin{figure}[H]
    \centering
    \includegraphics[width=1\linewidth]{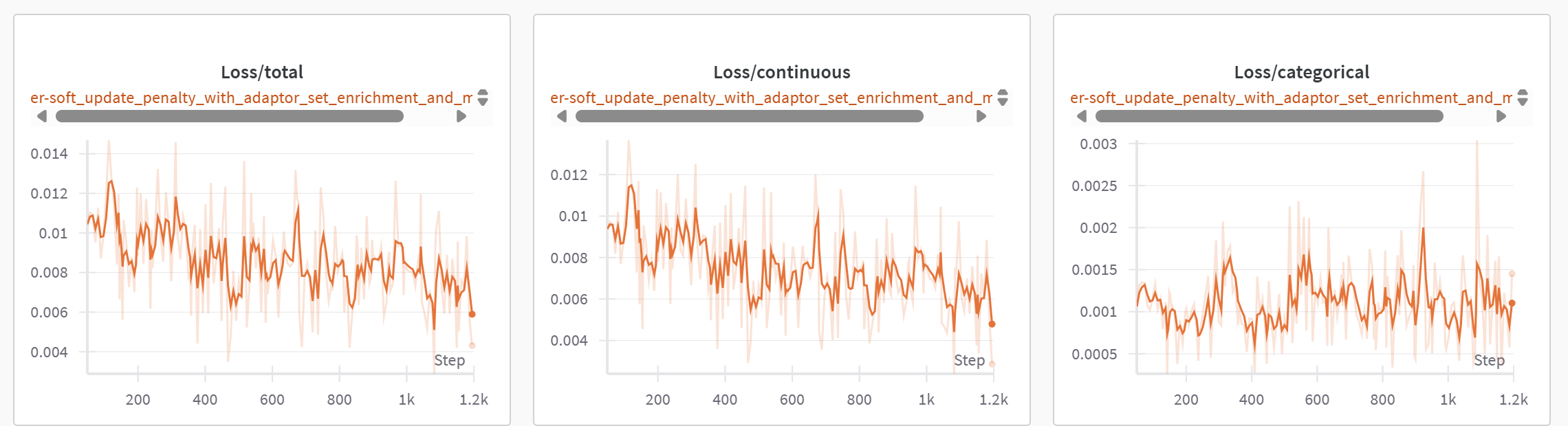}
    \caption[Stabilised training loss with soft updates]{Loss curves for the full strategy with adapter and soft target network updates. The soft update stabilises training by reducing oscillations and suppressing large loss spikes, enabling smoother convergence.}
    \label{fig:soft_loss}
\end{figure}

Together, these results support the conclusion that soft updates are critical for stabilising learning when using adaptive memory systems like SETLE.

\paragraph{Challenges of Generalisation in the CREATE Environment}

The CREATE environment presents a particularly challenging setting for structured reinforcement learning due to its requirements for high interactivity, sparse rewards, and multi-step problem-solving. Unlike symbolic environments such as MiniGrid, where object interactions and transitions are discretised and low-dimensional, CREATE demands fine-grained motor control (e.g., tool-object manipulation) and sequential reasoning over long temporal horizons (e.g., positioning a ball onto a ramp to launch it into a target).

In the preceding sections, we introduced several architectural components designed to address these challenges: SETLE-based memory enrichment, an adapter module for aligning latent spaces, a matching penalty to discourage repetitive retrieval, and a soft update mechanism to enhance stability during policy learning. While each of these mechanisms showed empirical benefits, their success depends heavily on the interplay between memory retrieval, enrichment quality, and policy plasticity. Not all runs achieve generalisation, as shown in Fig.~\ref{fig:not_always}.

\begin{figure}[ht]
\centering
\includegraphics[width=1\linewidth]{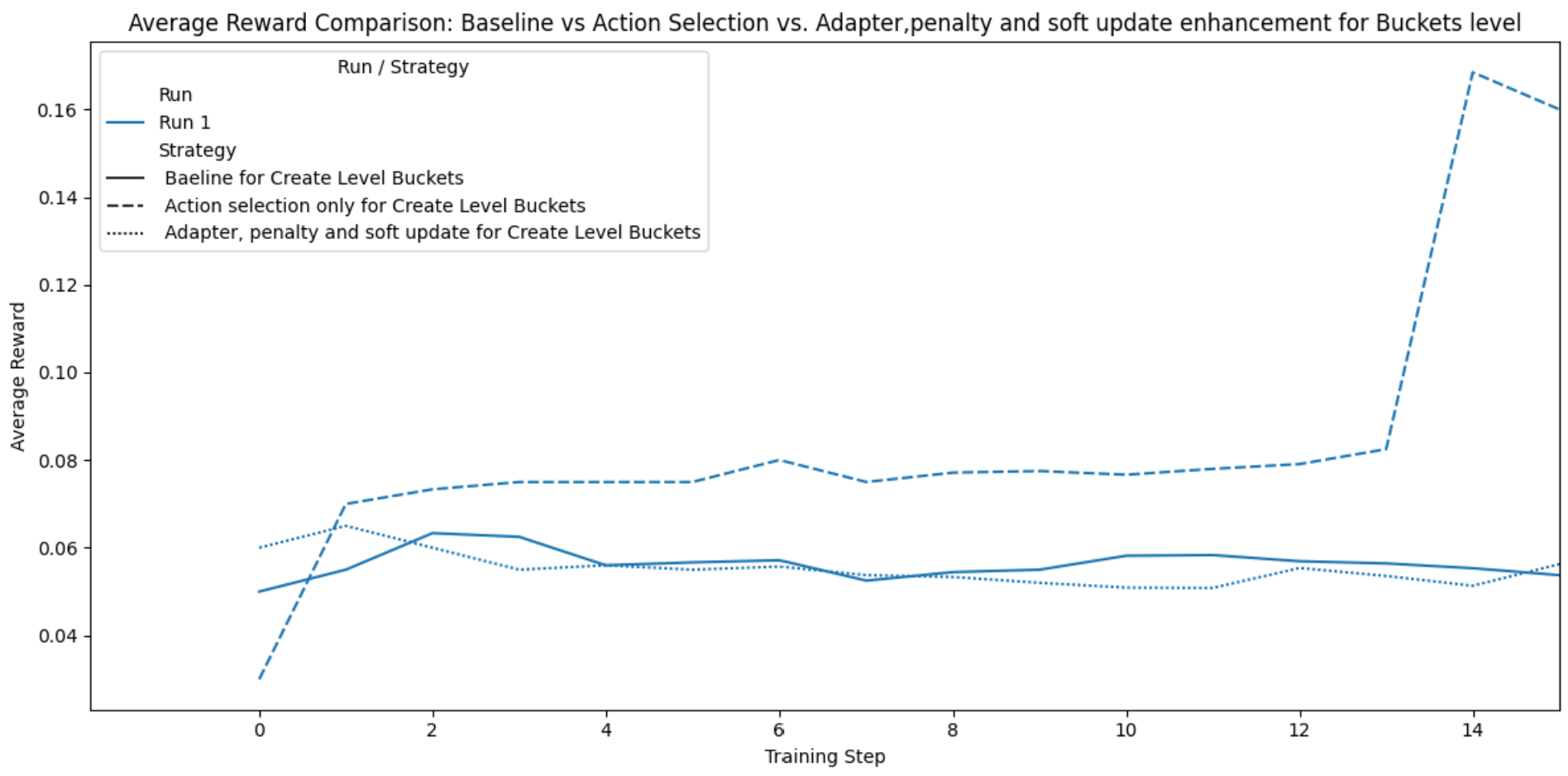}
\caption[Remaining failure cases in CREATE tasks]{Some CREATE tasks remain sub-optimal even with SETLE enrichment, highlighting the difficulty of achieving consistent generalisation across runs.}
\label{fig:not_always}
\end{figure}

Nevertheless, when memory matching and integration mechanisms align successfully, particularly under the full adapter and soft update strategy, the agent demonstrates substantial improvements in reward acquisition and trajectory quality, outperforming the baseline (Fig.~\ref{fig:superior}).

\begin{figure}[ht]
\centering
\includegraphics[width=1\linewidth]{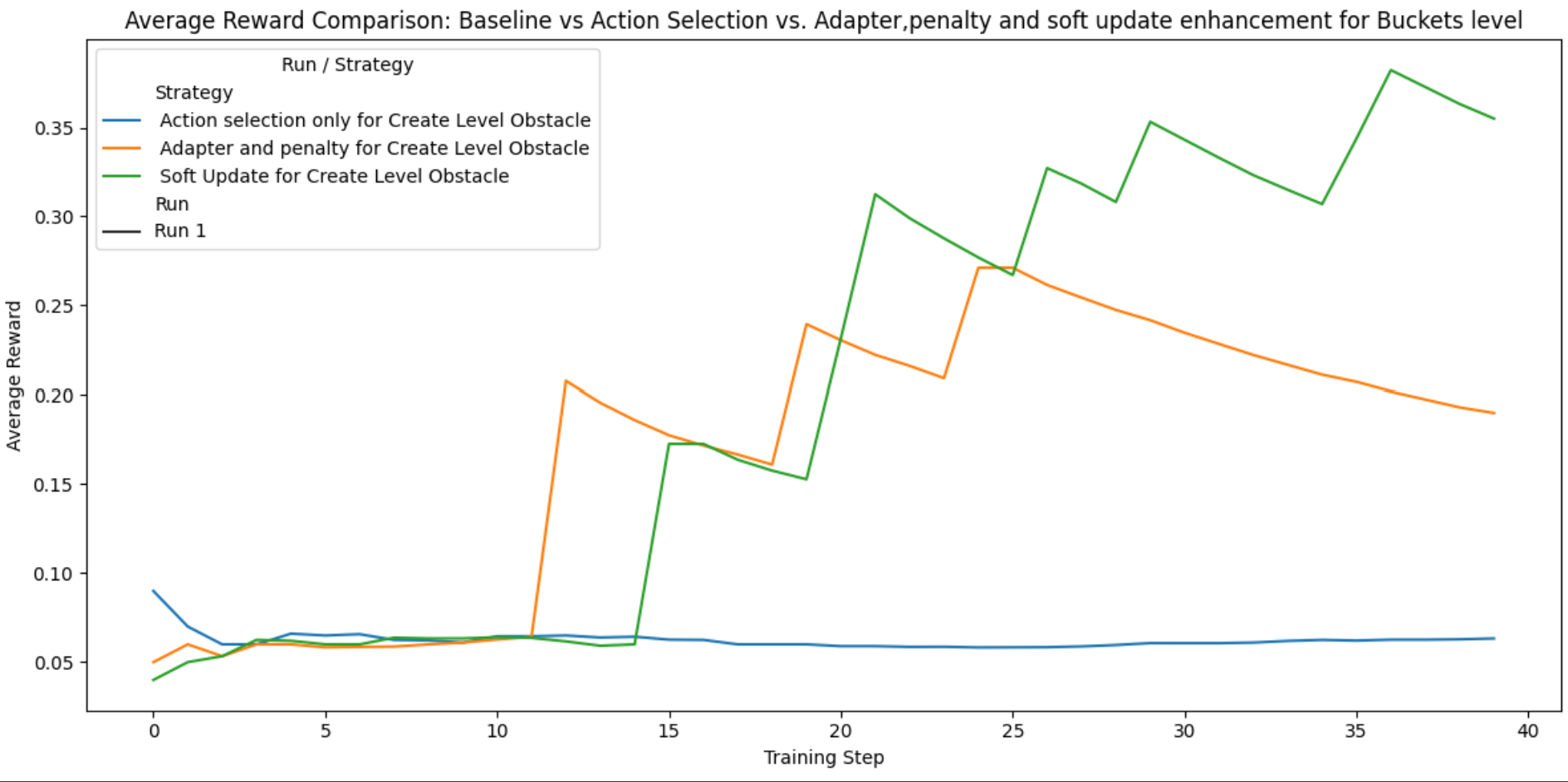}
\caption[Improved reward with full SETLE enrichment]{Performance comparison illustrating a case where full SETLE integration leads to dramatically improved reward outcomes.}
\label{fig:superior}
\end{figure}

Out of 42 independent runs across different tasks and configurations, only those using both the adapter and soft update enhancements managed to solve the environment at least once (Fig.~\ref{fig:only_success}). These results highlight not only the potential of structured trajectory enrichment, but also the fragility of the approach in complex domains. Importantly, these experiments were conducted under deliberately constrained training regimes: episode lengths were capped at 9 steps, and overall training time was limited. We deliberately constrained the training setup—limiting the number of episodes and capping each trajectory at 9 steps—to focus on evaluating whether structured memory and enrichment mechanisms can accelerate learning and enable generalisation under low-data, few-step conditions. This design tests the efficiency of SETLE-enhanced agents in challenging settings where standard reinforcement learning strategies would typically struggle to succeed without extensive trial-and-error or curriculum-based shaping.

\begin{figure}[ht]
\centering
\includegraphics[width=1\linewidth]{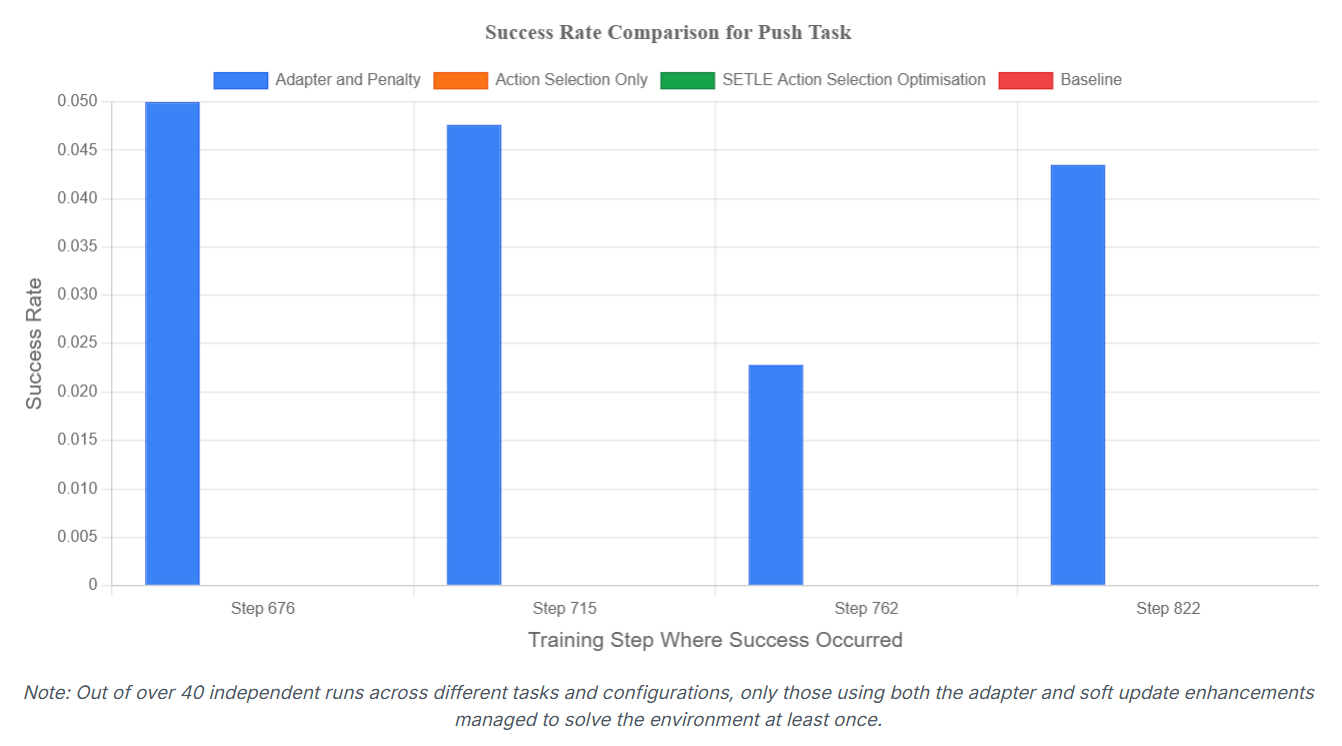}
\caption[Success rates in complex environments]{The agent's performance was tested across 42 independent runs. The chart visualizes a key finding from these experiments: across all 42 runs, the Baseline, Action Selection Only, and SETLE Optimisation strategies never succeeded once. Only the 'Adapter and Penalty' strategy achieved a non-zero success rate.}
\label{fig:only_success}
\end{figure}

\paragraph{Summary of key metrics across strategies}
\begin{figure}[ht]
    \centering
    \includegraphics[width=1\linewidth]{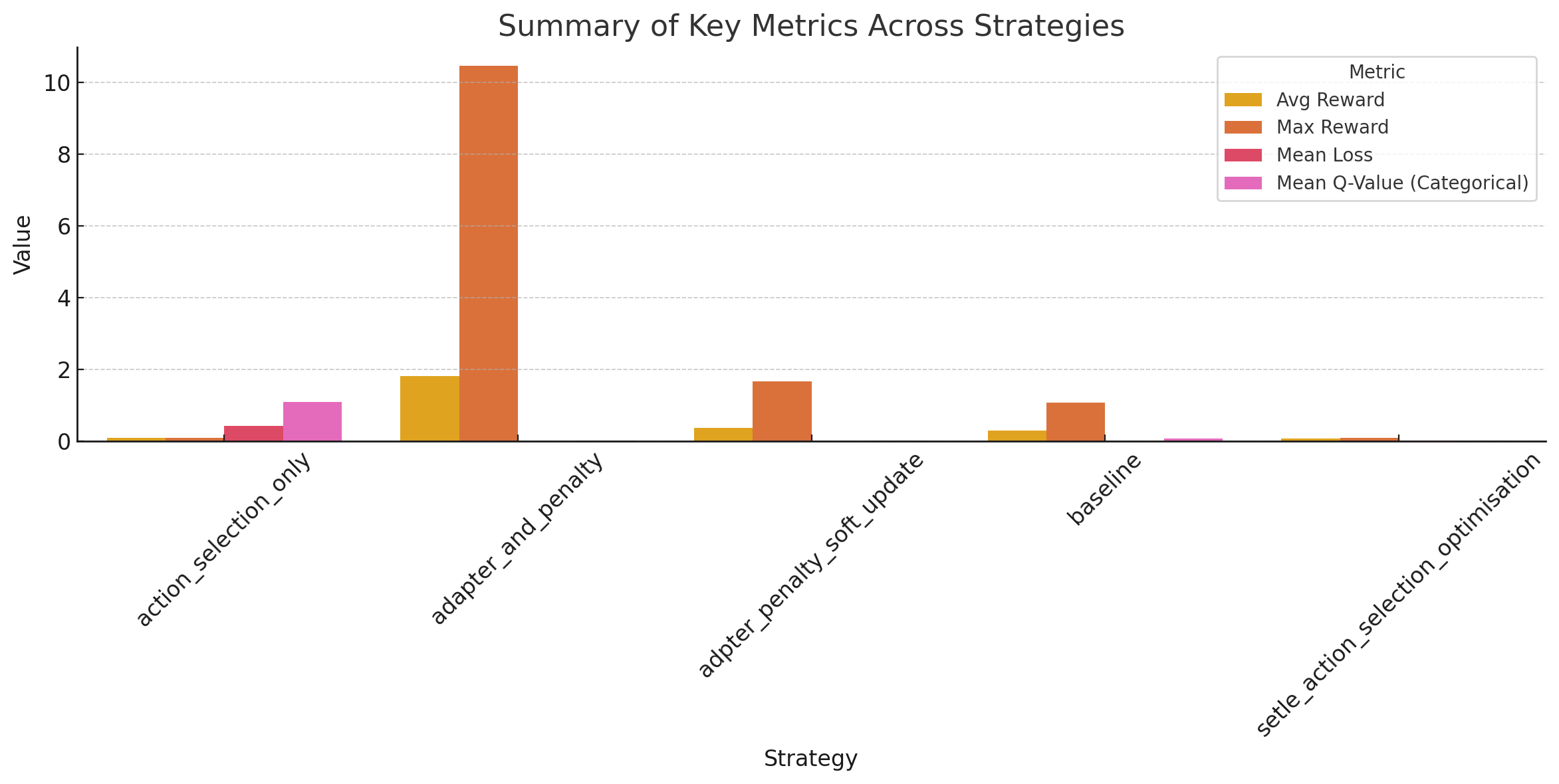}
    \caption[Reward, loss, and Q-value comparison across strategies]{The bar chart compares average reward, maximum reward, mean loss, and mean Q-value (categorical) across all strategy variants for 5 different runs for each of the task and strategy. The adapter and penalty strategy exhibits the highest average and peak rewards, while the action selection only strategy shows higher Q-values but also greater loss. Baseline and other non-optimised configurations remain lower across all axes, highlighting the benefits of integrated enrichment and adaptive mechanisms.}
    \label{fig:all_runs}
\end{figure}
The bar plot (Fig.\ref{fig:all_runs}) summarises key performance metrics across different reinforcement learning strategies enhanced with SETLE. Here are the main insights:
\begin{enumerate}
    \item Adapter and Penalty is the most effective strategy overall, achieving the highest average reward (around 1.8) and maximum reward ($>$ 10), with the lowest overall loss and modest Q-values. This supports earlier findings that combining knowledge integration (via SETLE) with adaptation and exploration encourages meaningful generalisation.
    \item Adapter + Penalty + Soft Update comes second in performance, with lower but still non-trivial rewards and stable losses, suggesting that soft updates help maintain training stability but may slow aggressive reward gains.
    \item Action Selection Only and SETLE for Action Selection and Optimisation yield the lowest average rewards, indicating that access to memory alone is not sufficient—adaptation layers and retrieval regulation are essential. While the Action and Optimisation strategy does outperform the baseline in several runs as seen previously, its performance is inconsistent across tasks. This suggests that optimisation over enriched states can lead to better outcomes, but without additional mechanisms like the adapter or penalty, the agent may struggle to stabilise learning or generalise across different task configurations.
    \item Baseline performs slightly better than random SETLE variants without adaptation but lacks the reward stability and peak performance of more structured approaches.
\end{enumerate}

Overall, these results confirm that structured enrichment must be accompanied by mechanisms for adaptation (adapter layers) and memory regulation (penalty, soft update) to produce consistent and transferable learning in environments like CREATE. Other logging, as well as how attention and tool reuse has been tracked, can be found in Appendix C.

\subsubsection{Results in Symbolic Reasoning Tasks (MiniGrid)}
To evaluate the behaviour of the SETLE-enhanced reinforcement learning agent in reward-sparse settings, we first conducted experiments in MiniGrid on the Empty-5x5 task. This environment is intentionally minimalist, containing no explicit goals or rewards, and thus provides no extrinsic learning signals during interaction. The aim of this experiment is not to assess task completion but to observe the \textbf{value dynamics and learning behaviour} of agents when operating under severe epistemic uncertainty and absence of reinforcement. Both the baseline and SETLE-enhanced agents were trained using Double DQN designed to reduce overestimation bias by decoupling action selection from evaluation.

Despite this bias correction, the baseline agent displayed a rapid and sustained increase in Q-values over time, even in the complete absence of reward (see Fig.~\ref{fig:q_empty5x5}).

\begin{figure}[H]
    \centering
    \includegraphics[width=1\linewidth]{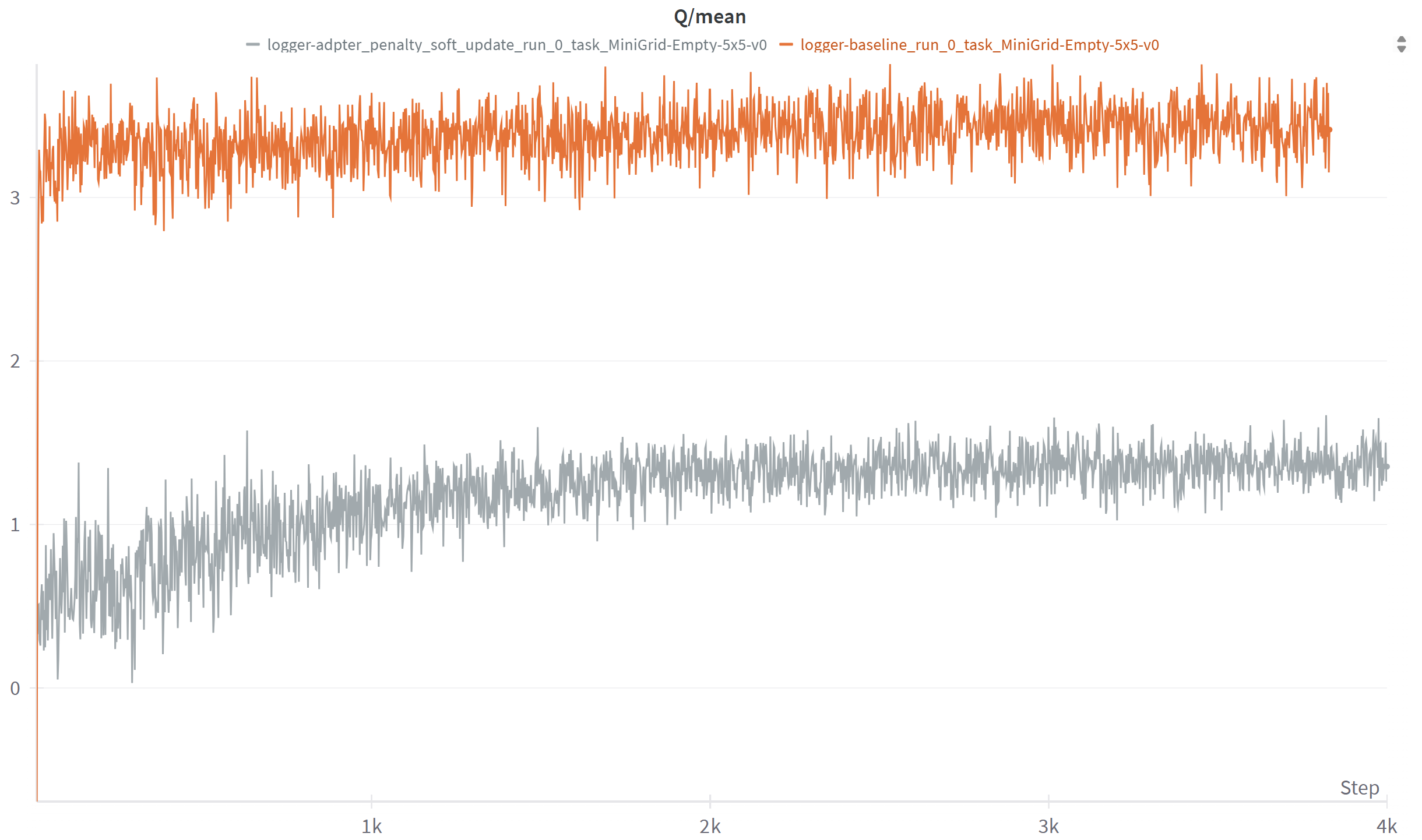}
\caption[Q-value behaviour in MiniGrid-Empty-5x5]{Comparison of Q-value evolution for Baseline and SETLE-enhanced agents in \texttt{MiniGrid-Empty-5x5}. The baseline agent exhibits sustained value inflation despite receiving no reward, with Q-values rising above 3.0. The SETLE-enhanced agent maintains more conservative Q-values around 1.0, reflecting structurally grounded value estimation under sparse feedback.}
    \label{fig:q_empty5x5}
\end{figure}

On average, baseline Q-values stabilized around 3.3 after a few hundred steps. This behaviour indicates a form of \textbf{residual value inflation}, wherein Q-values are recursively bootstrapped from their own overestimated targets. Since no extrinsic signal exists to anchor these predictions, the value function drifts into \textbf{ungrounded optimism}, assigning high expected returns to sequences of actions that are never positively reinforced. Such overconfidence, though syntactically correct in terms of Bellman updates, poses a threat to generalisation, especially when agents are transferred to environments where structural conditions change or delayed rewards appear.

\begin{figure}[ht]
    \centering
    \includegraphics[width=1\linewidth]{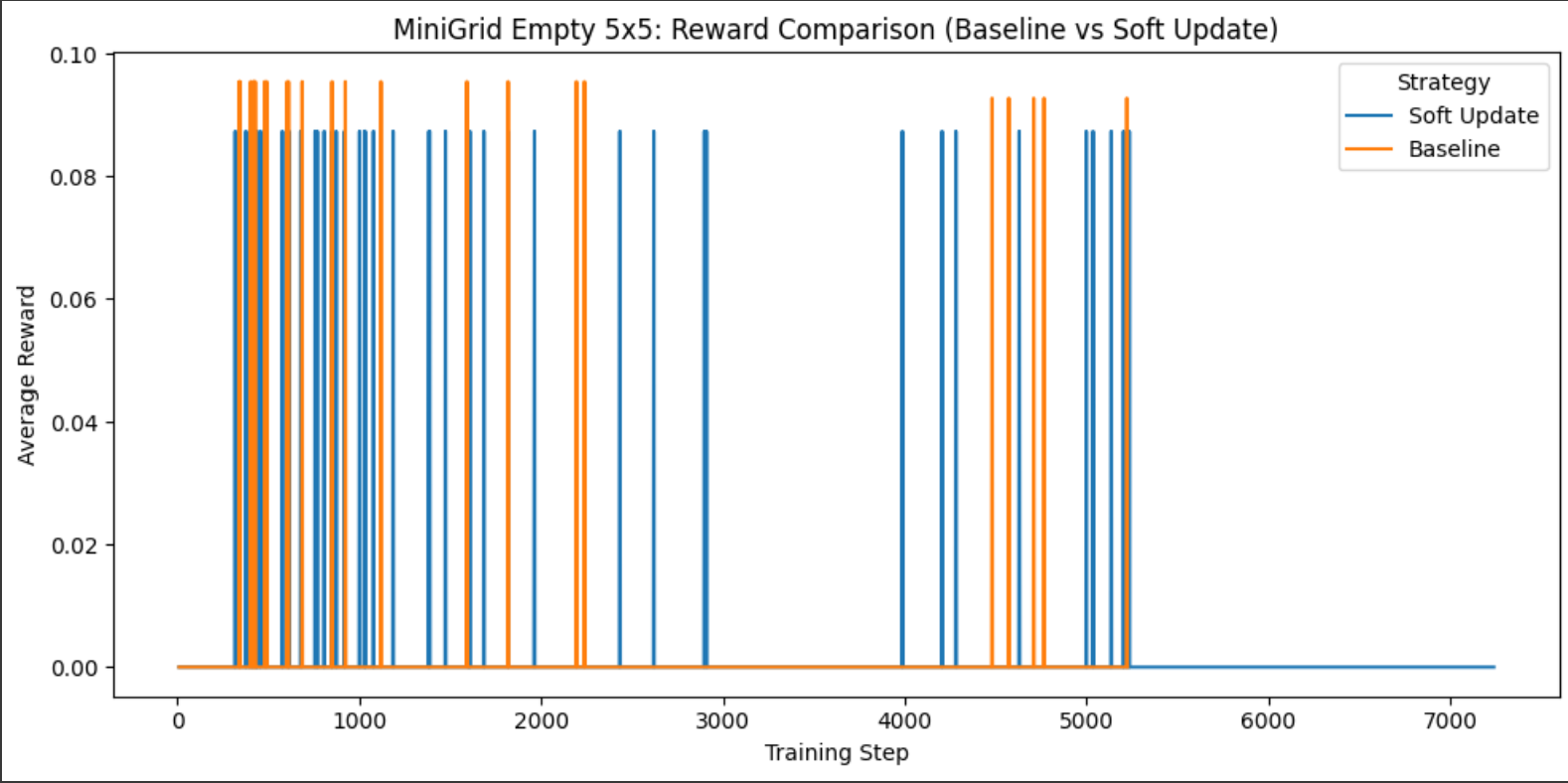}
\caption[Average reward per step in MiniGrid-Empty-5x5]{Average reward per step for Baseline (orange) and SETLE-enhanced agents (blue) in \texttt{MiniGrid-Empty-5x5}. Although overall rewards are low due to the sparse-reward nature of the environment, the SETLE-enhanced agent achieves nearly twice the reward frequency of the baseline, indicating improved exploration and task-relevant behaviour despite the absence of explicit supervision.}
    \label{fig:rew_log_mini}
\end{figure}

By contrast, the SETLE-enhanced agent maintained Q-values near 1.0 throughout training, less than one-third of the baseline level, indicating a more conservative and context-sensitive learning dynamic. At first glance, this may appear as an indication of stagnation or failure to learn. However, closer inspection of the reward logs (Fig.\ref{fig:rew_log_mini}) reveals that the SETLE-enhanced agent achieved \textbf{nearly double the reward frequency per step} compared to the baseline (0.0084 vs. 0.0048) (Fig.\ref{fig:mini_cum_rew}). Although absolute rewards remained low due to the task’s sparsity, the increased reward density and stabilised Q-values suggest that SETLE’s structured memory support fosters more \textbf{calibrated value estimation} and meaningful exploration.

\begin{figure}[H]
    \centering
    \includegraphics[width=1\linewidth]{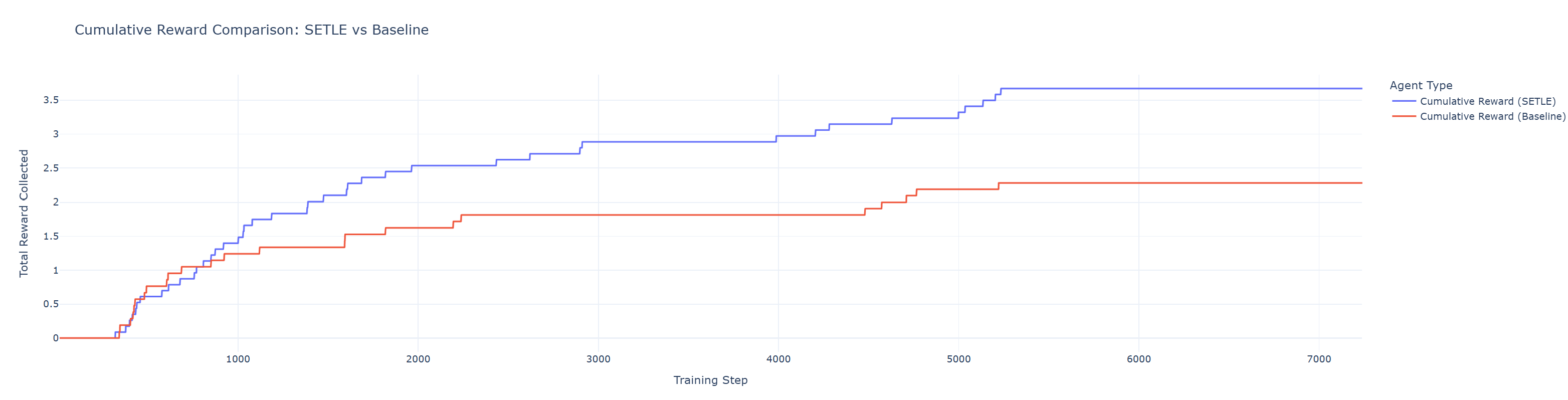}
\caption[Cumulative reward in MiniGrid-Empty-5x5]{Cumulative reward comparison between the baseline agent and the SETLE-enhanced agent in the \texttt{MiniGrid-Empty-5x5} environment. The SETLE-enhanced agent consistently collects reward more frequently and achieves a higher overall total, despite the sparse and minimal nature of the environment. This demonstrates the benefit of structured trajectory enrichment in guiding exploration and sustaining learning under low-signal conditions.}
    \label{fig:mini_cum_rew}
\end{figure}

Rather than hallucinating reward where none exists, SETLE retrieves structured prior experiences—such as object co-occurrences and contextual affordances, that allow the agent to infer when sequences are not worth propagating optimistic returns. In doing so, the agent suppresses speculative Q-value growth and aligns its internal value function with the actual utility of its behaviours. This form of value realism is increasingly recognised as a desirable property in deep RL. Recent work \citep{fujimoto2018addressing, kumar2020conservative} has highlighted the risks of overestimation in Q-learning and proposed conservative or uncertainty-aware alternatives to promote trustworthy generalisation.

In summary, while the baseline agent appears to "learn" faster in numerical terms, its Q-values are \textbf{ungrounded and misleading} in the context of a reward-sparse environment. The SETLE-enhanced agent, by contrast, produces Q-value trajectories that more accurately reflect the lack of positive reinforcement, and in doing so, demonstrates \textbf{stronger potential for generalisation} across structurally similar tasks. 

\paragraph{Behavioural Analysis in the Hardest Mini-Grid Task: SimpleCrossingS9N1}

In contrast to the \texttt{Empty-5x5} task, which served as a diagnostic environment for reward-free value estimation, the \texttt{SimpleCrossingS9N1} task represents the most complex environment in our Mini-Grid evaluation suite. It requires the agent to navigate through multiple rooms separated by walls and doors, some of which are partially obstructed, requiring careful planning and multi-step exploration. The sparse reward signal is delivered only upon successful navigation to a distant goal, making this task particularly challenging for unstructured exploration.

The results reveal divergence in agent behaviour across strategies. In terms of raw task success, the \textbf{adapter + penalty + soft update} strategy achieves the highest number of successful episodes (see Fig.\ref{fig:success_crossing_soft}), while the \textbf{adapter + penalty} strategy performs second, even though still low (Fig.\ref{fig:success_crossing_soft}). Both outperforming the baseline in terms of final performance, for which no success is obtained.

\begin{figure}[ht]
    \centering
    \includegraphics[width=1\linewidth]{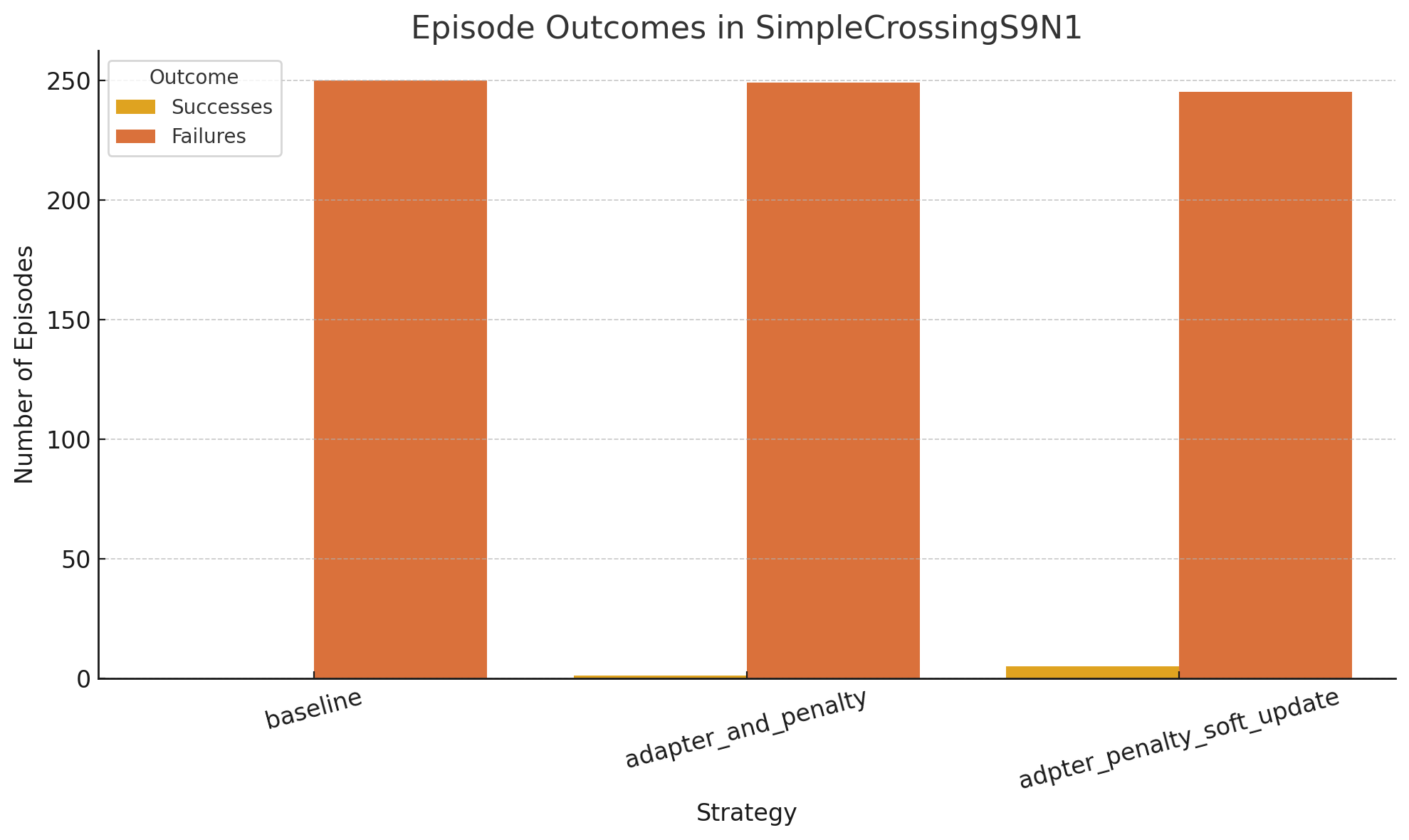}
\caption[Success/failure counts in SimpleCrossingS9N1]{Success and failure counts across strategies for the \texttt{SimpleCrossingS9N1} task. All agents were trained for the same number of episodes. The \textbf{adapter + penalty + soft update} strategy achieved the most successful completions. In contrast, the baseline agent recorded zero successes, underscoring the importance of structured memory for navigating long-horizon, sparse-reward environments.}
    \label{fig:success_crossing_soft}
\end{figure}

However, as shown in Fig.~\ref{fig:success_crossing_soft}, these success gains come at the cost of increased trial failures. The method although it comes with improvements it is yet to be satisfactory.

The Q-value trajectories in Fig.~\ref{fig:q_crossing} illustrate again the q-value overestimation. The baseline agent quickly settles into a narrow range of inflated Q-values (2.0–2.2), reflecting premature convergence to suboptimal policies. In contrast, SETLE-enhanced agents—particularly those using soft update and adapter integration—exhibit high-frequency but bounded Q-value fluctuations near 0.0. These fluctuations reflect richer but more cautious value estimation, as agents continuously reassess the utility of action paths in the absence of consistent external reinforcement.

\begin{figure}[ht]
    \centering
    \includegraphics[width=1\linewidth]{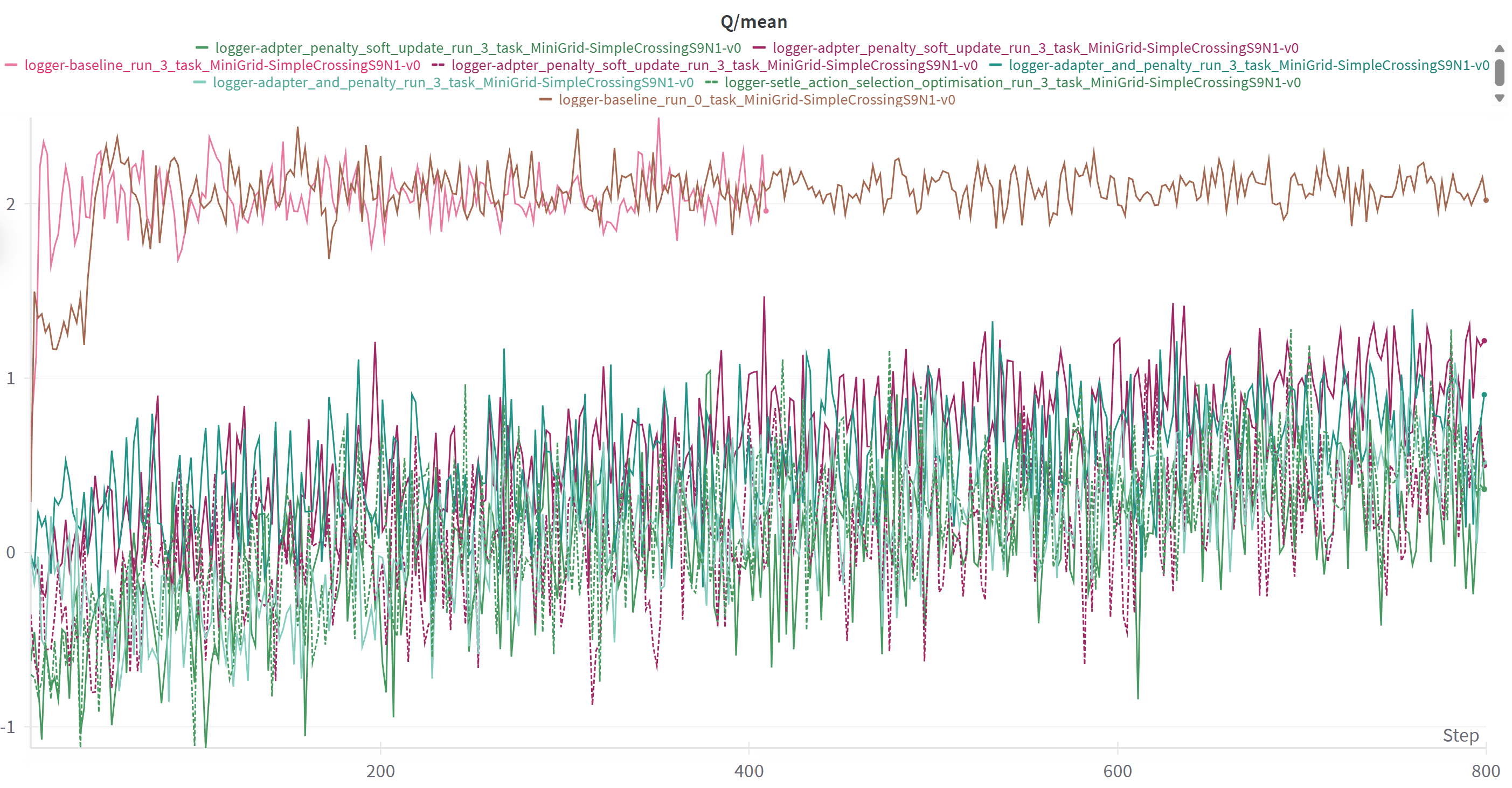}
\caption[Q-value evolution during SimpleCrossing training]{Evolution of average Q-values during training in \texttt{SimpleCrossingS9N1}. The baseline agent exhibits elevated and persistent Q-value inflation (around 2.0), which does not correlate with actual reward acquisition. SETLE-enhanced agents, especially those using adapters and soft updates, maintain lower, noisier Q-values—reflecting a more grounded and cautious value estimation process suitable for sparse-reward environments.}
    \label{fig:q_crossing}
\end{figure}

Overall, these findings underscore the strength of structured memory enrichment in complex symbolic environments. SETLE-enhanced strategies not only improve final success rates, but also exhibit more realistic and adaptive learning dynamics under sparse-reward conditions. In the hardest MiniGrid setting, this manifests as a trade-off between exploratory depth and training volatility—one that SETLE manages better than traditional baselines.

\subsubsection{Cross-task episode embedding comparison}
While traditional object-centric representations offer semantic interpretability, they often fall short in enabling cross-domain generalisation. Our analysis reveals this limitation clearly. The violin plot in Figure~\ref{fig:object_similarity_violin} shows the distribution of top-$k$ cosine similarities between MiniGrid objects (e.g., \textit{a golden key}, \textit{a red triangle}) and their closest counterparts in CREATE. Even for visually simple or conceptually similar categories like \textit{a red key}, the distribution remains narrow and shifted towards low similarity scores. This indicates that despite having similar names or roles, the embeddings of objects trained in distinct environments remain poorly aligned in practice. Such discrepancies stem from differences in rendering style, visual resolution, and context-specific usage—highlighting the fragility of object-level transfer across domains.

\begin{figure}[h]
    \centering
    \includegraphics[width=1\linewidth]{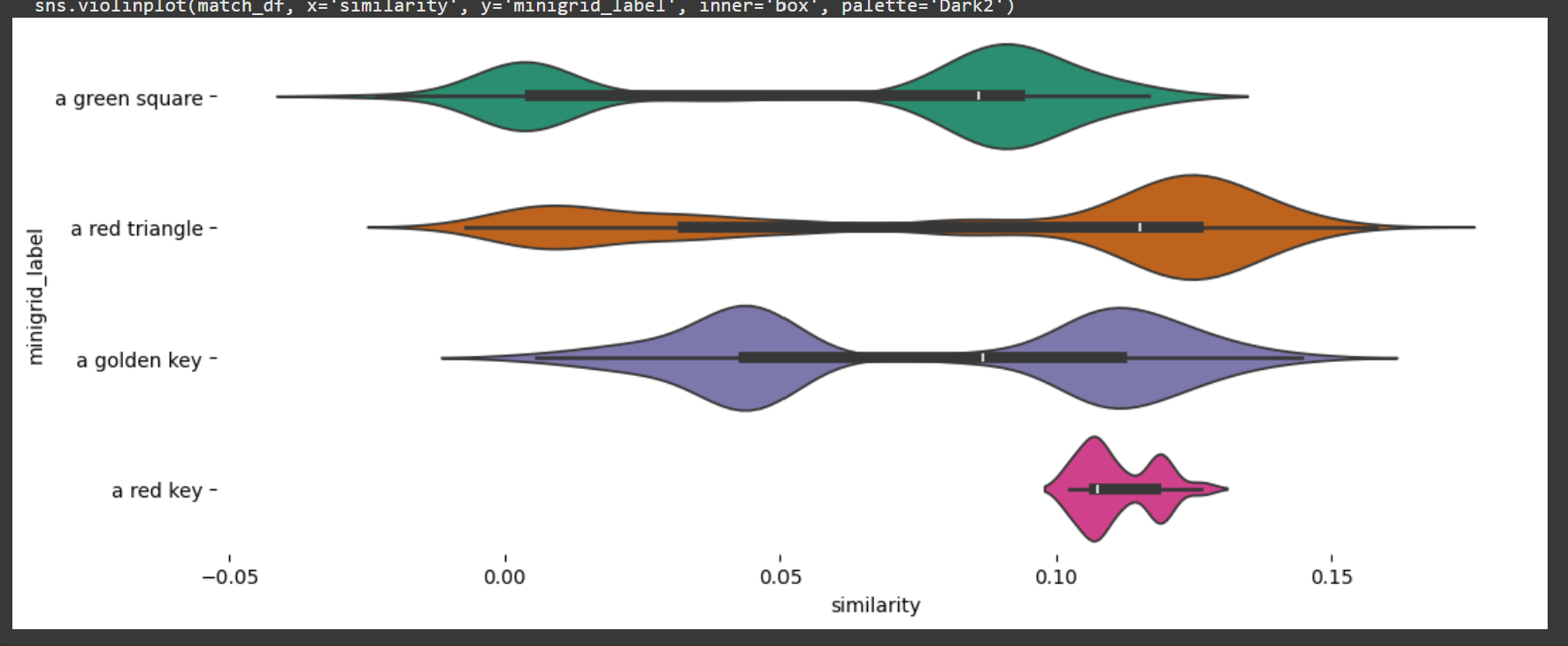}
\caption[Object embedding similarity across domains]{Distribution of top-$k$ cosine similarities for MiniGrid object embeddings matched to CREATE objects. Even semantically aligned objects exhibit narrow similarity distributions, indicating poor object-level alignment across domains. This supports the claim that functional generalisation must operate above raw perceptual encoding.}
    \label{fig:object_similarity_violin}
\end{figure}

In contrast, when comparing entire trajectory embeddings—capturing not just objects but also actions, affordances, and their interdependencies, a more encouraging picture emerges. As illustrated in the heatmap of Figure~\ref{fig:cosine_across_env}, several CREATE and MiniGrid task pairs show non-trivial cosine similarities at the episode level. For example, \texttt{CreateLevelPush-v0} and \texttt{CreateLevelBelt-v0} exhibit meaningful structural overlap with "MiniGrid-DoorKey-5x5 and \texttt{MiniGrid-Empty-5x5}, suggesting that the agent learns shared procedural structures such as navigation, obstacle manipulation, and goal seeking.

\begin{figure}[h!]
    \centering
    \includegraphics[width=1\linewidth]{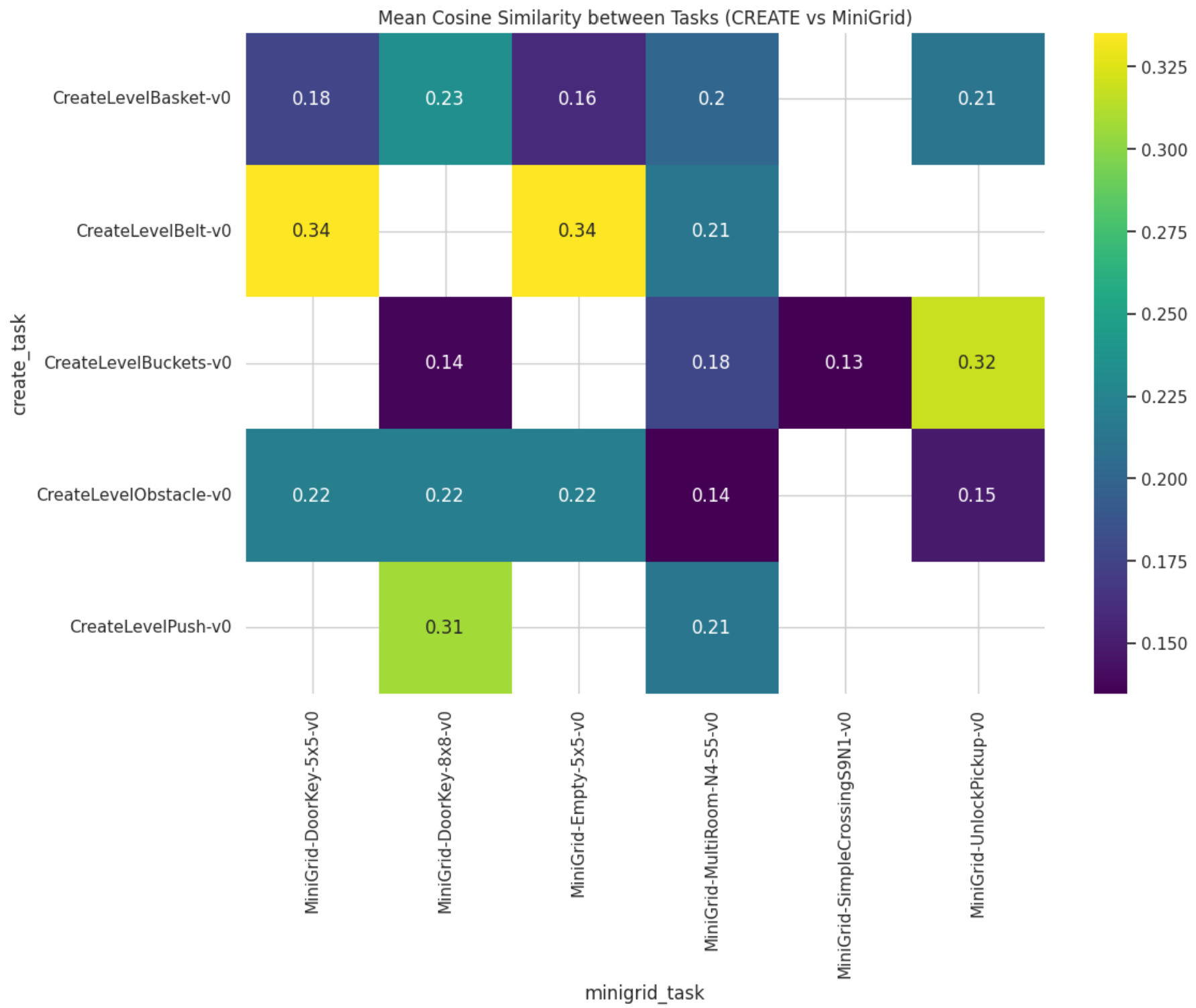}
\caption[Episode embedding similarity between task environments]{Mean cosine similarity between episode-level embeddings across CREATE and MiniGrid tasks. Higher values along diagonals suggest structural alignment between tasks like \texttt{CreateLevelPush} and \texttt{MiniGrid-DoorKey-5x5}, despite differences in appearance and control schemes. This highlights SETLE's ability to extract relational and sequential regularities shared across environments.}
    \label{fig:cosine_across_env}
\end{figure}

\begin{figure}[h!]
    \centering
    \includegraphics[width=1\linewidth]{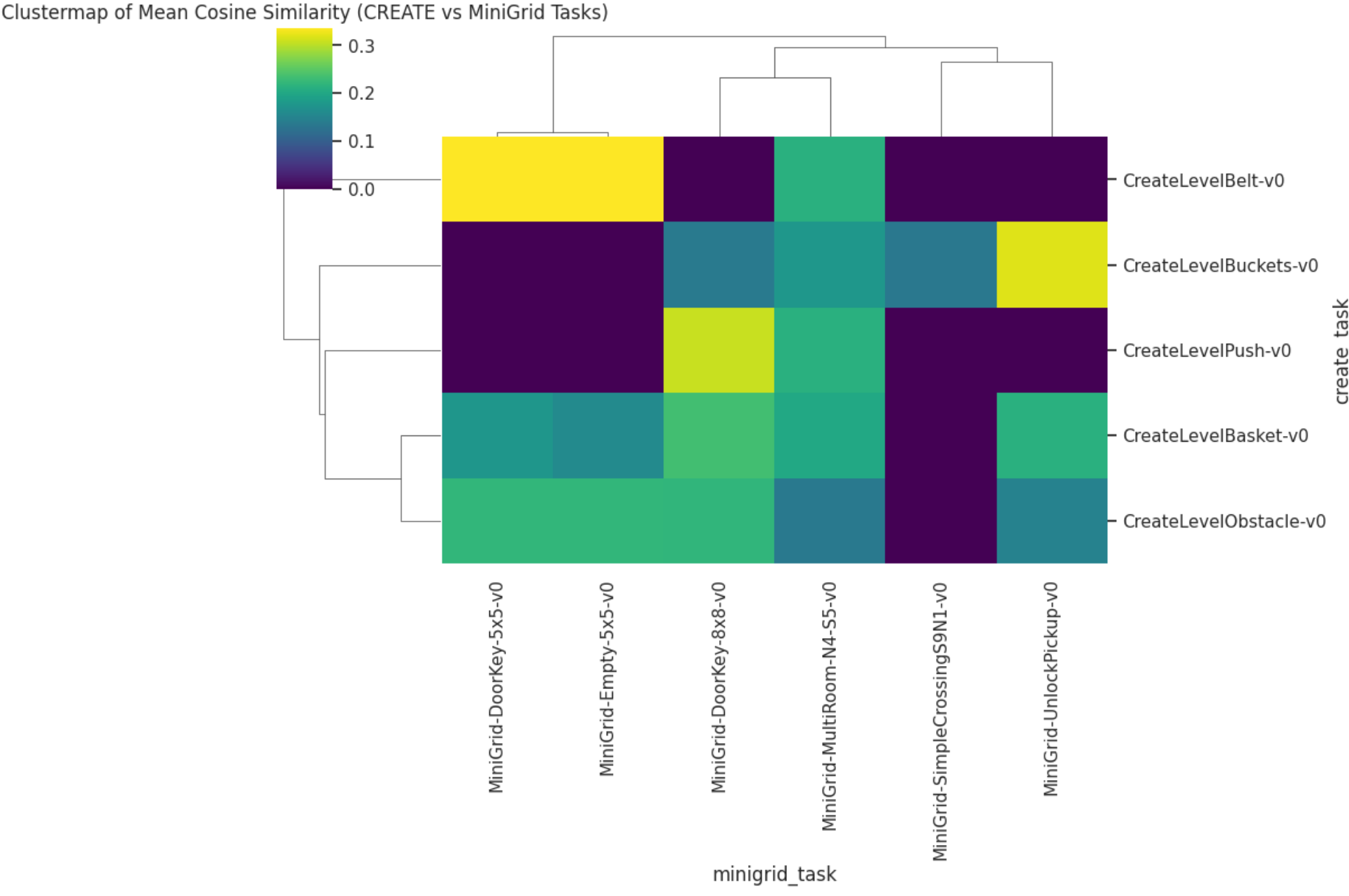}
\caption[Hierarchical clustering of task embeddings]{Hierarchical clustering of average episode embeddings from CREATE and MiniGrid tasks. Tasks from both environments are grouped based on shared interaction structure and relational dependencies, not domain origin.}
    \label{fig:clustermap_crossdomain}
\end{figure}

The clustermap in Figure~\ref{fig:clustermap_crossdomain} reinforces this trajectory-level alignment. Tasks from the two domains are not segregated by origin but grouped according to their functional and temporal similarities. For instance, "CreateLevelBuckets" and \texttt{MiniGrid-UnlockPickup} are clustered together, despite vast differences in visual and mechanical design. This pattern reflects the agent’s ability to learn and compare episodes based on high-level interaction graphs, rather than low-level appearance features.

Together, these results highlight the core advantage of SETLE's approach: rather than relying on static embeddings of isolated objects, it encodes episodes as structured graphs capturing action-effect relationships and contextual dependencies. This abstraction enables meaningful comparison across environments and supports the agent’s ability to retrieve and reuse experience even in domains with drastically different visual or symbolic structure. These findings strengthen our argument that better generalisation in reinforcement learning hinges on structured, relational memory—not only raw perceptual similarity.

}

\section{Discussion}
In this paper, we have introduced Structurally Enriched Trajectories (SETs), a structured extension of traditional trajectory representations that captures multi-level dependencies between states, interactions, and affordances. These SETs form the basis of SETLE, a computational framework that extends graph embedding approaches like HeCo to episodic Reinforcement Learning (RL) by structuring trajectories as hierarchical subgraphs. Our experimental results demonstrate that this method successfully constructs a task-focused latent space by clustering similar trajectories. {  For instance, the encoder achieved a high Silhouette Score of 0.90 for unsuccessful trajectories (with a margin of 1.5) and a low Davies-Bouldin Index of 0.15, indicating that the embeddings formed clear and well-separated clusters based on task outcome. These contributions have broader implications for representation learning, task generalisation, and lifelong AI systems.

The practical advantages of these structured representations were validated when integrated into a full reinforcement learning loop. In complex, physically-grounded tasks, the SETLE-enhanced agent outperformed the baseline; for example, the adapter and penalty strategy achieved a maximum reward over 10.0, while the baseline peaked below 1.0 (Fig. \ref{fig:all_runs}). Furthermore, in sparse-reward symbolic environments, the framework fostered more realistic value estimation. The SETLE agent's Q-values remained stable around 1.0, thereby avoiding the ungrounded value inflation to over 3.0 observed in the baseline agent (Fig. \ref{fig:q_empty5x5}).}

A key challenge in RL is to ensure that learnt representations transfer across tasks rather than remaining task-specific \citep{deramo2024,chen2024}. By incorporating hierarchical trajectory representations, we enable the recognition of shared patterns and support knowledge transfer. This aligns with findings that hierarchical structuring is an essential mechanism for generalisation \citep{sancho2024} and reflects principles from human cognition, where learning depends on recombining meaningful substructures from experience \citep{lake2017building}. Many AI models, however, struggle with compositional flexibility, limiting their adaptability. { The importance of SETLE's hierarchical structure was empirically validated; our ablation studies demonstrated that flattening the graph representation substantially degraded clustering quality, with the Dunn Index for success clusters dropping from 1.652 to 1.326 (Table 3). This shows that our framework advances a more effective, hierarchical approach that enables agents to retain and reuse structured knowledge.}

Although hierarchical structures have been central to RL research for a long time \citep{botvinick2009hierarchically}, most current approaches depend on predefined skill segmentation, which limits the ability to autonomously discover reusable structures. In contrast, by modelling the relational dependencies between trajectory components, our method aligns with broader trends in structured representation learning \citep{hafner2020dream, roads2024dimensions}, where explicitly modelling these relationships enhances transferability.

In addition to generalising tasks, this work also advances lifelong learning and adaptive AI. Effective open-ended learning depends on a structured memory system that enables agents to accumulate and refine knowledge over time \citep{finn2017model}. The hierarchical memory structure in SETLE lays the groundwork for such capabilities. This adaptability extends across domains; while object-level embeddings failed to align between the CREATE and MiniGrid environments, SETLE's trajectory-level representations revealed functional similarities, such as a mean cosine similarity of 0.34 between CreateLevelBelt-v0 and MiniGrid-Empty-5x5-v0 (Fig. \ref{fig:clustermap_crossdomain}), demonstrating that the framework learns transferable task abstractions. This structured approach also mitigates the issue of catastrophic forgetting.

This work contributes to the broader discussion on structured intelligence in AI \citep{mohan2024structuredRL, eckstein2020hierarchicalRL} by providing an effective mechanism to link low-level state-action execution with high-level conceptual reasoning. Achieving human-like adaptability requires moving beyond mere pattern recognition to extract, manipulate, and reuse structured representations across various domains \citep{schneider2006, hommel2000, wu2022}. While similar approaches in RL have been proposed, challenges still exist in designing frameworks that can generalise beyond task-specific representations \citep{mohan2024structuredRL}.

While our framework offers several advantages, several challenges remain. Although we have adapted the HeCo architecture as the foundation for SETLE's encoder, we have not incorporated modified versions of heterogeneous GNN baselines (e.g., HeCo, GTC) as direct comparison methods in our experimental evaluation. As discussed in Section 4.1, this decision was based on fundamental differences in problem formulation. Whereas SETLE addresses episodic, dynamic graphs representing trajectories, methods like HeCo and GTC are designed for static, node-centric graph tasks. This limitation reflects a broader challenge in the field: the existing divide between static graph mining and dynamic reinforcement learning. Most state-of-the-art heterogeneous GNNs, including HeCo \citep{wang2021selfsupervised} and the more recent GTC \citep{sun2025gtc}, are optimised for node-centric tasks on invariant graphs. Adapting these architectures to serve as standalone baselines for episodic, subgraph-centric trajectory encoding remains an open challenge. Without standardised evaluation protocols that bridge static representation learning and sequential decision-making, it remains difficult to quantify how much an agent's performance stems from the underlying graph architecture versus the way it structures its experience over time. While this limitation does not detract from the validity of the present work, which demonstrates SETLE's effectiveness in encoding structured trajectories and improving RL performance, developing such unified benchmarks is a necessary step toward understanding how structured representations can be decoupled from specific task dynamics to achieve more generalizable intelligence

{Another key limitation is scalability; future work should evaluate SETLE in more complex, real-world environments such as robotics simulations, which involve richer dynamics and higher-dimensional state representations. This, however, poses significant challenges, as the computational complexity of encoding such graphs can be prohibitive, and the efficiency of hierarchical representations in these tasks is an unresolved issue. Additionally, this work primarily focuses on representation learning. A key area for future research is to deepen the understanding of how these structured representations impact policy optimisation. Moving forward, exploring how SETLE can be extended toward dynamic adaptation, enabling agents to continuously refine their representations in response to evolving task demands, will be crucial for pushing reinforcement learning toward lifelong, generalisable intelligence.}

Future research may extend this framework toward dynamic adaptation, enabling agents to continuously refine their representations in response to evolving task demands, pushing reinforcement learning toward lifelong, generalisable intelligence.

\bibliography{bibliography.bib}

\end{document}